\documentclass[Afour,sageh,times]{sagej}

\usepackage{moreverb,url}
\usepackage{amsmath,amsfonts}
\usepackage{algorithmic}
\usepackage{array}
\usepackage[caption=false]{subfig}
\usepackage{textcomp}
\usepackage{stfloats}
\usepackage[colorlinks,bookmarksopen,bookmarksnumbered,citecolor=red,urlcolor=red]{hyperref}

\usepackage{url}
\usepackage{graphicx}
\usepackage{balance}
\usepackage{color}
\usepackage{ulem} 
\usepackage{siunitx}

\usepackage{algorithm}
\usepackage{stmaryrd}
\usepackage{tabularx}
\usepackage{graphbox} 
\usepackage{array}
\usepackage{tabularray}
\usepackage{enumitem}
\usepackage{stackengine}

\newcommand{\videoLink}[1]{https://youtu.be/bNVxTh0eixI?t=#1}
\newcommand{\videoUrl}{\texttt{\url{https://youtu.be/bNVxTh0eixI}}}

\newcommand\BibTeX{{\rmfamily B\kern-.05em \textsc{i\kern-.025em b}\kern-.08em
T\kern-.1667em\lower.7ex\hbox{E}\kern-.125emX}}

\def\volumeyear{2024}
\begin{document}

\runninghead{Corberes et al.}

\title{Perceptive Locomotion through Whole-Body MPC and Optimal Region Selection}

\author{Thomas Corb\`eres\affilnum{1}, Carlos Mastalli\affilnum{2}, Wolfgang Merkt\affilnum{3},  Jaehyun Shim\affilnum{1}, Ioannis Havoutis\affilnum{3}, Maurice Fallon\affilnum{3}, Nicolas Mansard\affilnum{4}, Thomas Flayols\affilnum{4}, Sethu Vijayakumar\affilnum{1}, Steve Tonneau \affilnum{1} }

\affiliation{\affilnum{1}School of Informatics, University of Edinburgh, UK\\
\affilnum{2}School of Eng. and Physical Sciences, Heriot-Watt University, UK\\
\affilnum{3}Oxford Robotics Institute, University of Oxford, UK\\
\affilnum{4} LAAS-CNRS, France}

\corrauth{Thomas Corb\`eres}
\email{t.corberes@sms.ed.ac.uk}

\email{t.corberes@sms.ed.ac.uk}


\begin{abstract}
This paper describes the use of the \LaTeXe\
\textsf{\journalclass} class file for setting papers to be
submitted to a \textit{SAGE Publications} journal.
The template can be downloaded \href{http://www.uk.sagepub.com/repository/binaries/SAGE LaTeX template.zip}{here}.
\end{abstract}


\begin{abstract}
Real-time synthesis of legged locomotion maneuvers in challenging industrial settings is still an open problem, requiring simultaneous determination of footsteps locations several steps ahead while generating whole-body motions close to the robot's limits. State estimation and perception errors impose the practical constraint of fast re-planning motions in a model predictive control (MPC) framework. We first observe that the computational limitation of perceptive locomotion pipelines lies in the combinatorics of contact surface selection. Re-planning contact locations on selected surfaces can be accomplished at MPC frequencies (50-100 Hz). Then, whole-body motion generation typically follows a reference trajectory for the robot base to facilitate convergence. We propose removing this constraint to robustly address unforeseen events such as contact slipping, by leveraging a state-of-the-art whole-body MPC (\textsc{Croccodyl}). Our contributions are integrated into a complete framework for perceptive locomotion, validated under diverse terrain conditions, and demonstrated in challenging trials that push the robot's actuation limits, as well as in the ICRA 2023 quadruped challenge simulation.
\end{abstract}


\maketitle

\section{Introduction}

Reliable and autonomous locomotion for legged robots in arbitrary environments is a longstanding challenge. 
The hardware maturity of quadruped robots~\cite{ANYmal-robot,unitree-robot,Spot-robot} motivates the development of a motion synthesis framework for applications including inspections in industrial areas~\cite{ARGOS}. Synthesising motions in this context requires handling the issues of \textit{contact decision} (where should the robot step?) and \textit{Whole-Body Model Predictive Control} (WB-MPC) of the robot (what motion creates the contact?). 

Each contact decision defines high-dimensional, non-linear geometric and dynamic constraints on the WB-MPC that prevent a trivial decoupling of the two issues: How to prove that a contact plan is valid without finding a feasible whole-body motion to achieve it? Even worse, the environments we consider comprise holes and gaps, introducing a \textit{combinatorics} problem: On which contact surface(s) should the robot step? 

\begingroup
\def\boxit#1#2{%
  \smash{\llap{\rlap{\hspace{4pt}\href{#1}{\strut\raisebox{\height}{\phantom{\rule{0.98\linewidth}{#2}}} } }~}}\ignorespaces}
\hypersetup{pdfborder=0 0 0}
\begin{figure}[t]
    \boxit{\videoLink{135}}{95pt}%
    \includegraphics[trim ={0px 16px 116px 00px}, clip,width =\linewidth, keepaspectratio = true]{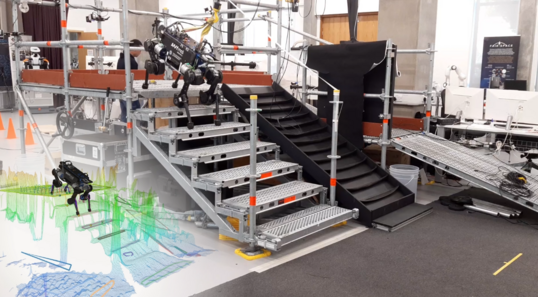}
    \vspace{-5mm}
    \caption{Industrial staircase descent with onboard perception. Video: 
    \videoUrl.}
    \label{fig::intro_walk_down}
\end{figure}
\endgroup

\subsection{State of the art}
The mathematical complexity of the legged locomotion problem in arbitrary environments is such that an undesired decoupling between contact planning and whole-body control is required. Typically, a contact plan describing the contact locations is first computed, assumed to be feasible, and provided as input to a WB-MPC framework to generate whole-body motions along it. As a result, the contact decision must be made using an approximated robot model, under the uncertainty that results from imperfect perception and state estimation. The complexity of the approximated model has, unsurprisingly, a strong correlation with the versatility and computational efficiency of the proposed approach.

\subsubsection{Offline contact planning with full kinematics}
Early approaches to contact planning are robust and complete, because they integrate a whole-body kinematic model in the planning phase, with quasi-static feasibility guarantees. These \textit{contact-before-motion} approaches~\cite{Bretl2006MotionPO, hauser2008motion, escande2009planning} mix graph-based search with contact-posture sampling to explicitly tackle the problem's combinatorics. Whole-body feasibility is explicitly checked before validating each new contact. The generality of these approaches is unmatched, but the associated computational cost is high (from minutes to hours), which prevents online re-planning. The computation time can be reduced to a few seconds by constraining the root's path~\cite{BouyarmaneGuide}, and then by approximating the robot with low-dimensional abstractions~\cite{Tonneau-Acyclic-planner,murookaastarreach}. These abstractions use hand-crafted heuristics for collision avoidance and geometry~\cite{7982757}, while dynamic feasibility is often asserted using centroidal dynamics (e.g.,~\cite{tonneau:2pac,Fernbach-CROC}). Such approximations proved unreliable for the most challenging scenarios (such as car egress motions~\cite{Tonneau-Acyclic-planner}), mostly because of the difficulty of approximating collision avoidance constraints. Still, most ``2.5D'' environments (which can be accurately represented with a heightmap) composed of quasi-flat contact surfaces (i.e., a friction cone containing a gravity vector) can be handled with such constraints~\cite{Tonneau-Acyclic-planner}. Unsurprisingly, these environments, including rubles, stepping stones and staircases, are the application targets for most contributions in the literature. 

\subsubsection{Online contact planning with reduced dynamics}
The combination of reduced dynamic models and simplified collision constraints makes optimisation-based techniques tractable. Optimal control is attractive as it allows us to find solutions robust to uncertainties through the minimisation of selected criteria. Sampling-based approaches, instead, only look at feasibility.
To model the combinatorics, the first approach is to relax the problem by modelling the discrete (boolean) variables that represent the contact decisions with continuous variables, resulting in a formulation that can be readily solved by off-the-shelf nonlinear programming (NLP) solvers~\cite{Mordatch2012DiscoveryOC,winkler18}. However, there is no guarantee that the system's dynamic constraints will be satisfied even with reduced models, with contacts potentially planned where no surfaces exist~\cite{tonneau:sl1m:9454381}. Alternatively, Linear Complementary Constraints (LCP) can be used to accurately model contact constraints, but they are notoriously difficult to handle by NLP solvers~\cite{Posa-TO-contact}. In both cases learning initial guesses can help the solver converge to a feasible solution~\cite{Melon-TOWR-Learning,Melon-towr-receiding}. The second approach is to explicitly handle combinatorics using Mixed-Integer Programming (MIP). MIP solvers tackle contact planning problems for bipeds~\cite{Deits2014FootstepPO} and quadrupeds~\cite{aceituno_cabezas-ral18}, provided that the underlying optimisation problem is convex, which results in a conservative approximation of the dynamics~\cite{Ponton_2021}. Monte Carlo Tree Search (MCTS) has recently been proposed as a promising alternative to MIP that could provide a relevant trade-off between exploration and exploitation~\cite{9812421}.  In this work, we choose MIP for planning contacts as it has experimentally led to the most effective results in terms of computation time and reliability~\cite{risbourg:hal-03594629}. 

\subsubsection{Perceptive locomotion with instantaneous decisions}
Whether the environment is fully known or not impacts on the validity of a method. Reactive perceptive pipelines exist on the LittleDog robot \cite{Zucker-Optimisation-LittleDog, Kolter-LittleDog, Kalakrishnan-LittleDog} and have inspired further works \cite{Fankhauser-Robust-rough-ETH}. However, they all require high-precision terrain pre-mapping and an external motion capture system. When the environment is not fully known, it is typically modelled as an elevation map by fusing depth sensor information within proprioceptive information~\cite{Fankhauser2018ProbabilisticTerrainMapping, Miki-ANYmal-wild-perception}.
Recent approaches propose to directly optimise the next contact position, the torso orientation and obstacle avoidance for the foot trajectory based on this input~\cite{Jenelten_2022_TAMOLS, Grandia-multi-layer-CBF, Grandia-perceptive-through-NMPC}. The approaches share similarities with the framework we propose in terms of the model's proposition. However, their main difference is that they focus on planning the immediate contact location and posture, which is why we argue that a preview window of several steps ahead is required for the scenarios we consider. As discussed in Section~\ref{sec::discussion}, we believe that combining these approaches is a promising research avenue.

\subsubsection{Whole-body predictive control relying on CoM motions}
Because of the nonlinearity induced by any changes to the contact plan, the WB-MPC rarely challenges the step locations, even though the approximations do not guarantee feasibility. The uncertainties resulting from state estimation and environment perception motivate frequent re-computation of the contact plan, which is not possible as their frequency is usually low (about \SI{5}{\hertz} in~\cite{tonneau:sl1m:9454381}). Furthermore, the WB-MPC is usually additionally constrained to track a reference trajectory for the Centre Of Mass (COM) or the base~\cite{carpentiermulticontact,crocoddyl20icra} to facilitate convergence, but we argue that this tracking is problematic when perturbations such as contact slipping occur.
Our conclusion is that the  use of reduced models for contact planning necessarily leads to errors in the WB-MPC that result in slipping contacts. In the current state of the art, reduced models appear necessary for satisfying real-time constraints. Mitigating this issue involves allowing to adapt a contact plan at a higher frequency. However, it involves writing a WB-MPC that robustly accommodates these errors and gives as much freedom as possible when following a contact plan.

\subsection{Contribution}
This paper extends our published conference paper~\cite{risbourg:hal-03594629}, where we propose a \textit{contact repositioning} module between the contact surface planner and the MPC to adapt to the robot's estimated state and perceived environment uncertainties. To achieve this, we decouple the contact surface selection from the control, but not the computation of the contact position on that surface, which we instead update synchronously with the MPC output and updated state of the robot, at \SI{50}{\hertz}. As such, the contact repositioning module is our primary contribution. This work is thus to be considered as an experimental and theoretical extension in the following manner.

We propose a novel, complete perceptive locomotion architecture comprised of the following features: terrain segmentation, real-time surface selection and footstep planning, free-collision foot-swing planning, and whole-body MPC which considers torque limits and generates local controllers. It relies on four technical contributions: 
\begin{enumerate}[label=(\roman*)]
    \item a theoretical contribution to the decoupling between contact planning and WB-MPC,
    \item an empirical demonstration of the added value of WB-MPC in perceptive locomotion,
    \item a convex segmentation approach using onboard terrain elevation maps, and
    \item exhaustive hardware trials on challenging terrains demonstrating increased capabilities for the ANYmal B robot.
\end{enumerate}

Finally, we extend the plane segmentation algorithm~\cite{Fallon2019PlaneSeg}, which decomposes potential contact surfaces into a sequence of convex surfaces needed by our contact planner. We use the Visvalingam–Whyatt and Tess2 algorithms to correctly handle overlapping surfaces and conservatively reduce the complexity of the scene, leading to a more versatile and robust environment construction. Our framework results in state-of-the-art locomotion capabilities under a wide range of conditions, robust to strong perturbations including sliding contacts and missed steps during stair climbing.

The reader should note that we have incorporated a feedback WB-MPC~\cite{Agile-Maneuvers-mastalli, Inv-Dyn-MPC-mastalli} into our framework, as opposed to the more conventional approach in our previous work, which involved MPC with reduced dynamics followed by a WBC running at a higher frequency. Constraining only the end-effector trajectories, instead of the CoM motions, allows the WB-MPC to freely accommodate substantial perturbations and perception errors, and to maximise the robot's capabilities by optimising its posture. 
While a subset of our experimental results are shared by both papers, their contributions are orthogonal.

\section{Architecture overview}

\label{sec::overview}

An overview of the locomotion pipeline is presented in Fig.~\ref{fig::archi_overview}. Walkable surfaces are described via convex planes extracted from the terrain elevation map at \SI{1}{\hertz}. Given the current state of the robot (position / orientation of the base, position of active contacts) and a desired velocity (joystick input), a mixed-integer program is used to select the convex surfaces on which the next 6 or 8 steps will occur, depending on the gait used. A new surface plan is computed at the beginning of each new phase, which corresponds to approximately 3-5\,Hz in our experiments. Given the next contact surfaces, the trajectory of each end-effector is updated at \SI{50}{\hertz}, before each iteration of the whole-body MPC. The control policy is then sent to the robot with a Riccati gain controller. State estimation is performed onboard at \SI{400}{\hertz} by fusing inertial sensors from IMU and odometry provided by the ANYbotics software \cite{Bloesch-RSS-12}. Finally, a LIDAR is used to correct the drift of these measurements by analysing fixed points in the environment.  
 
\begingroup
\begin{figure*}[t]
\centering
\stackinset{l}{0pt}{t}{0pt}{\hyperref[sec::perception]{\makebox(195,35){}}}{%
\stackinset{l}{128pt}{b}{0pt}{\hyperref[sec::surface_selection]{\makebox(67,35){}}}{%
\stackinset{l}{246pt}{b}{0pt}{\hyperref[sec:foot_trajectory_generation]{\makebox(67,35){}}}{%
\stackinset{l}{246pt}{t}{0pt}{\hyperref[sec::motion_generation]{\makebox(163,35){}}}{%
\includegraphics[width=0.98\textwidth]{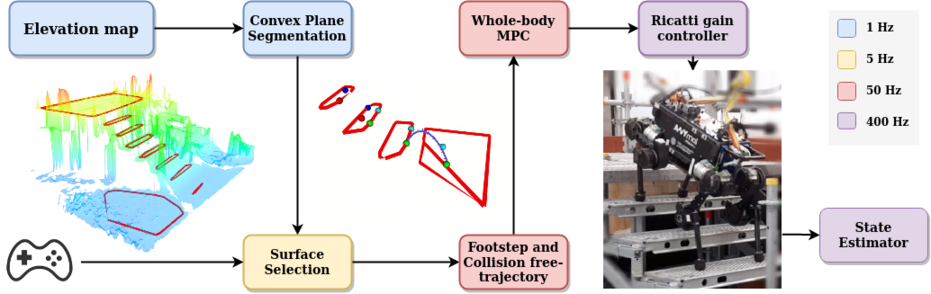}}}}}
\caption{Overview of our perceptive locomotion pipeline. Around 1Hz, the perceptive elements \textit{Elevation map} and \textit{Convex Plane Segmentation} extract convex planes from the surrounding environment (Section \ref{sec::perception}). The \textit{Surface Selection} block, running around 5Hz (depending on the gait chosen), chooses between them the next surfaces of contact (Section \ref{sec::surface_selection}). At the frequency of the MPC, \SI{50}{\hertz}, \textbf{Footstep and Collision free-trajectory} elements generate the curve for each moving foot (Section \ref{sec:foot_trajectory_generation}). Finally, the \textit{Whole-Body MPC} and \textbf{Riccati gain controller} synthesise the motion (Section \ref{sec::motion_generation}).}
\label{fig::archi_overview}
\end{figure*}
\endgroup

\section{Definitions and notations}
\label{sec:def}

In line with the notations used in our previous work \cite{risbourg:hal-03594629}, the \textit{robot state} is formally described by the:
\begin{itemize}
    \item Centre Of Mass (COM) position, velocity and acceleration $\mathbf{c}$, $\mathbf{\dot{c}}$ and $\mathbf{\ddot{c}}$, each in $\mathbb{R}^3$; 
    \item base transformation matrix in the world frame;
    \item 3D position of each  end-effector in the world frame;
    \item gait, i.e., the list of effectors currently in contact, as well as the contacts to be activated and deactivated over the planning horizon. 
\end{itemize}

The \textit{horizon} $n$ is defined as the number of future contact creations considered. In the case of the trotting gait, a horizon $n=6$ describes three steps, as at each step two contacts are created simultaneously.  

At the \textit{Surface Selection} planning stage, motion is decomposed into \textit{contact phases}. Each contact phase is associated with a number of feet in contact and one or more contacts are broken/created at each phase. In the case of a trotting gait with a horizon $n=6$, it corresponds to three contact phases since two feet move at the same time. For a walking gait, the horizon $n=8$ contains 8 contact phases since only one foot moves at a time.

The \textit{environment} is the union of $m+1$ disjoint quasi-flat\footnote{ie such that its associated friction cone contains the gravity vector.} contact surfaces $\mathcal{S} = \bigcup_{i=0}^{m} \mathcal{S}^i$. Each set $\mathcal{S}^i$ is a polygon embedded in a 3D plane, i.e.,
\begin{equation}\label{eq:surf}
\begin{aligned}
    \mathcal{S}^i := \{\mathbf{p} \in \mathbb{R}^3 |  \mathbf{S}^i \mathbf{p} \leq \mathbf{s}^i \}  \,,
\end{aligned}
\end{equation}
 where $\mathbf{S}^i \in \mathbb{R}^{h\times3}$ and $\mathbf{s}^i \in \mathbb{R}^{h}$ are respectively a constant matrix and a vector defining the $h$ half-space bounding surface.

The \textit{contact plan} is described as a list of contact surfaces $\mathcal{S}^j_k \in \mathcal{S}, 1 \leq j \leq l$ with $l$ being the total number of end-effectors and $k$ being the $k$-th contact phase. 
\section{Perception}
\label{sec::perception}

This section reviews the \textit{Elevation map} and \textit{Convex Plane Segmentation} components of the architecture (Fig.~\ref{fig::archi_overview}).

\subsection{Sensors}

Regarding the proprioceptive sensors, the state is estimated by fusing leg odometry and the Xsens MTi-100 IMU \cite{Bloesch-RSS-12}. Additionally, a rotating Hokuyo UTM-30LX-EW lidar sensor is placed at the back of the robot to correct the drift of the state estimation with an iterative closest point (ICP) method \cite{Hutter-anymalB}. A single depth camera Intel RealSense D435, mounted at the front of the robot, 
extracts height information from the surrounding area into a point cloud. High-accuracy presets are used on the camera to prioritize the accuracy of the point cloud over speed. This increases the quality of the elevation map and no filters are needed to post-process the point clouds. However, this preset degrades the frequency in our setup to a range between 2 and \SI{5}{\hertz} for point cloud collection.
\subsection{Contact surfaces extraction}
The probabilistic and local mapping method in \cite{Fankhauser2018ProbabilisticTerrainMapping} converts point cloud data into an elevation map locally around the robot's pose. Including proprioceptive localisation from kinematic and inertial measurements produces an estimate of the surroundings as a 2.5D heightmap. Potential surfaces are then extracted with the Plane-Seg algorithm \cite{Fallon2019PlaneSeg} by clustering planar points with similar normals. The planes are extracted without memory and therefore the surfaces can change suddenly with each height map update. The quality and consistency of the surfaces then depend entirely on the quality of the height map, as developed in Sec.~\ref{sec::discussion}.

\subsection{Refinement and margin of safety}
Post-processing of Plane-Seg planes is essential to ensure safe footstep decisions. We have added it for three reasons. First, the complexity of the contact surfaces (i.e., the number of points) has a significant impact on the computation time of the surface selection algorithm. We propose conservatively approximate surfaces with more than 8 vertices with an 8-vertex polygon. Filtering is also done to remove unreachable surfaces. 
For example, the plane extraction method could return overlapping surfaces when considering a staircase. Here, the ground surface is often detected below the steps and planning a footstep inside it will obviously result in failures.
Finally, a safety margin is applied, around \SI{4}{\centi\meter} on each surface, to avoid putting feet on the edges of the surfaces. This can be harmful in the event of estimation errors. This margin is also useful to avoid knee collisions of the knees with the environment, which are not explicitly accounted for.

\subsubsection{Vertices number reduction}The number of vertices is first reduced using the conservative Visvalingam–Whyatt line simplification algorithm \cite{Visvalingam1993LineGB}. It eliminates progressively the points from a line that forms the smallest area with its closest neighbours as described in pseudo-code \ref{alg::Visvalingam}. No hyperparameters other than the final number of points (which is 8) are needed. Fig.~\ref{fig::reduction_pts} shows an example of this reduction.

\begin{algorithm}
\caption{Pseudo-code for Visvalingam–Whyatt line simplification}
\label{alg::Visvalingam}
\begin{algorithmic}[1]
\STATE List of $n$ vertices : $L =   [ P_0 \dots P_n]$
\WHILE{ len($L$) $>$ $n_{\textrm{max}}$}
    \FOR{$i \in [0,len($L$)-2]$}
        \STATE Compute area of $[ P_i, P_{i+1}, P_{i+2} ] $
    \ENDFOR
    \STATE Remove $P_{i+1}$ corresponding to the smallest area.
\ENDWHILE
\RETURN $L$
\end{algorithmic}
\end{algorithm}


\subsubsection{Safety margins} 
\begin{figure}[ht]
     \centering
    \subfloat[]{%
      \includegraphics[width=0.45\linewidth]{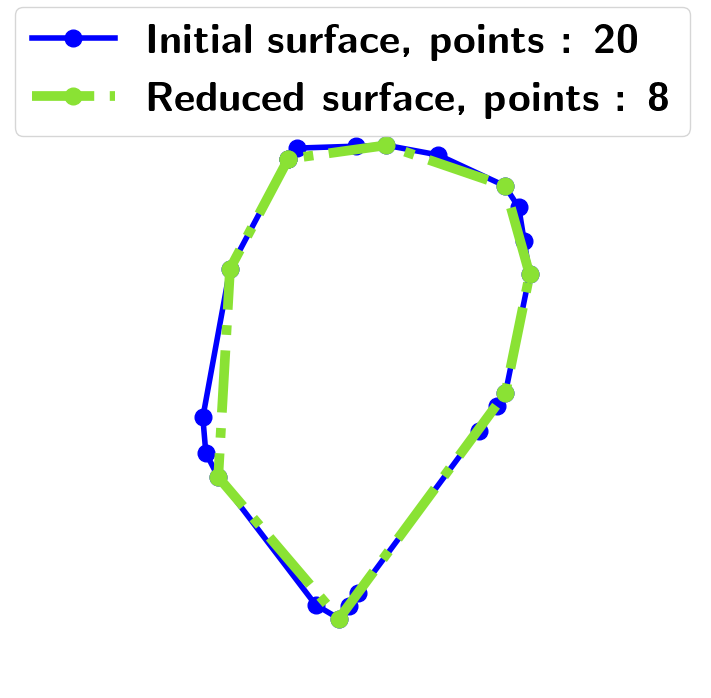}
      \label{fig::reduction_pts}}
    \subfloat[]{%
      \includegraphics[width=0.45\linewidth]{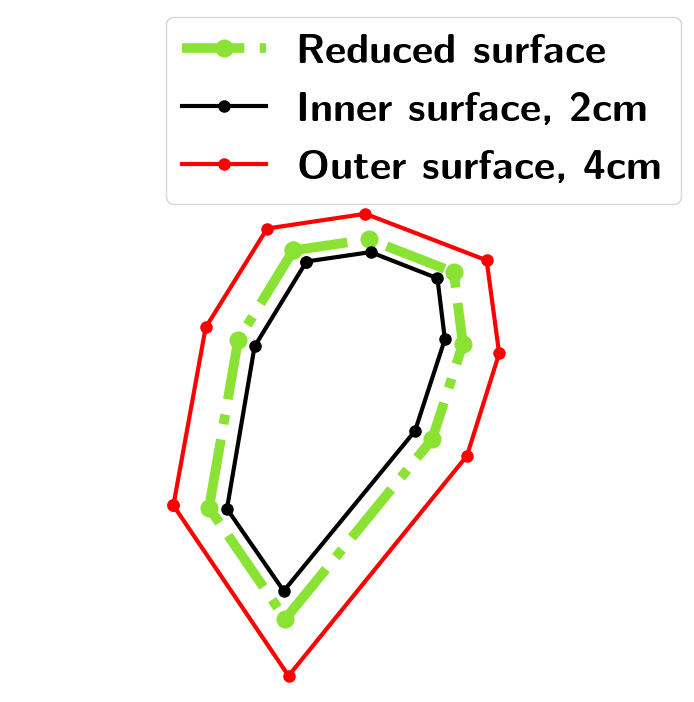}
      \label{fig::inner_outer}}

    \caption{Reduction to an 8-vertex polygon using the Visvalingam–Whyatt algorithm on an initial 20-vertex polygon on the left \ref{fig::reduction_pts}. Applying inner and outer margins to an 8-vertex polygon \ref{fig::inner_outer}.}
    \label{fig:margin}
\end{figure}

On this updated contour, an inner and an outer margin are computed parallel to each edge as shown in Fig.~\ref{fig::inner_outer}. The inner margin allows the robot to step into a safe area. The outer margin artificially increases the size of the obstacle to prevent the end-effector from getting too close of the obstacle. This is done to avoid collisions while computing swing-foot trajectories. These two margins are used to avoid stepping on the corner of an obstacle due to state estimation errors and for collision avoidance with the body. 

\subsubsection{Convex decomposition}
\begin{figure}[ht]
     \centering
    \subfloat[]{%
      \includegraphics[width=0.45\linewidth]{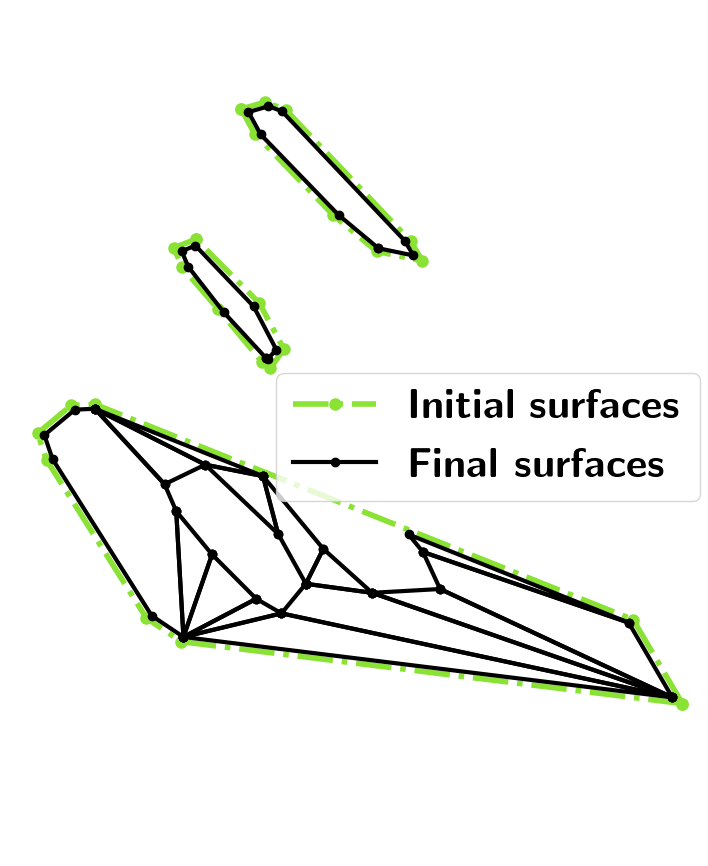}
      \label{fig::3D_decompo}}
    \subfloat[]{%
      \includegraphics[width=0.45\linewidth]{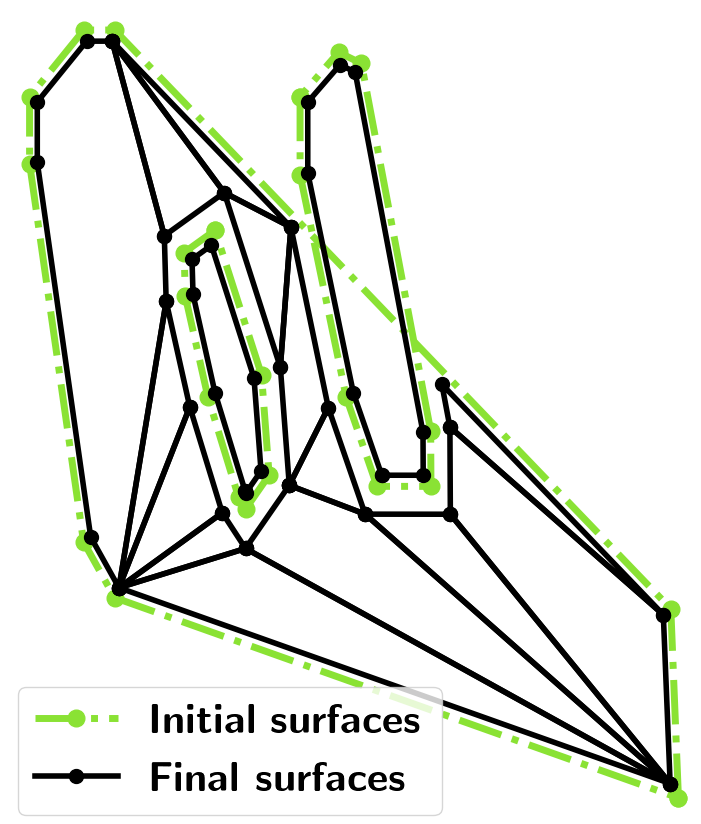}
      \label{fig::2D_decompo}}

    \caption{Example of surface processing and convex decomposition. These two figures represent the same 3D scene with 2 air overlapping the ground surface. On the right Fig.~\ref{fig::2D_decompo}, the scene is viewed from a top perspective.}
    \label{fig:processing_surfaces}
\end{figure}

Starting from the lowest surface, the overlapping surfaces are removed and a convex decomposition is performed on the resulting contour using the Tess2 algorithm \cite{DecompoSoftware} as shown in Fig.~\ref{fig::3D_decompo}. Smaller areas under 0.03 $\textrm{m}^{2}$ are deleted. Algorithm \ref{alg::surface_decompo} gives the pseudo-code of this post-processing. In addition, a rectangle is added below the robot's position, at the estimated height of the feet. This is to ensure that there is always a surface under the robot's feet if the elevation map has not been built. This surface is treated differently in the process to avoid overlapping it with real obstacles but has been removed from the pseudo-code for clarity.

\begin{algorithm}
\caption{Pseudo-code for surface processing}
\label{alg::surface_decompo}
\begin{algorithmic}[1]
\FORALL {surfaces}
\STATE   Reduce nb of points with Visvalingam's algorithm.
\STATE   Compute inner and outer contour.
\ENDFOR
\STATE $L_i \leftarrow $ List of surfaces in ascending order of height.
\STATE $L_f \leftarrow $ Empty list.
\WHILE{$L_i$ is not empty}
    \STATE Get first surface $Sf$ and remove it from $L_i$.
    \STATE $L_o \leftarrow$ Empty list.
    \FORALL{surface in $L_i$}
        \IF{surface.outer intersect with $S_f$.inner}
            \STATE Increment $L_o$ with outer contour.
        \ENDIF
    \ENDFOR
    \IF{$L_o$ is not empty}
        \STATE Convex Decomposition between $S_f$.inner and $L_o$.
        \STATE Remove small areas.
        \STATE Add remaining surfaces in $L_f$.
    \ELSE
        \STATE Add $S_f$.inner in $L_f$
    \ENDIF
\ENDWHILE
\RETURN $L_f$
\end{algorithmic}
\end{algorithm}




\section{Surfaces selection}
\label{sec::surface_selection}

Given the current state of the robot (active contacts and position, COM location), the environment given as a union of non-intersecting surfaces, as well as a desired target velocity for the robot and a desired gait, our 
\textit{Surface Selection} (Fig.~\ref{fig::archi_overview} - yellow) algorithm computes a feasible contact plan, composed of $n$ contact surfaces that the robot should step on for the planning horizon (Sec.~\ref{sec:def}). We set $n=8$ for a walking gait and $n=6$ for a trotting gait in our experiments. The surface selection algorithm is executed between 3 and \SI{5}{\hertz}. This means that the contact plan is updated before each new step is made by the robot. The frequency depends on the gait and each optimisation starts at the beginning of each step. 

The algorithm is implemented as a Mixed-Integer Program (MIP) \cite{Deits2015EfficientMP}. We use the SL1M formulation of this algorithm~\cite{tonneau:sl1m:9454381}, with adaptations to better match the desired robot behaviour. These adaptations are in \cite{risbourg:hal-03594629} and are described here for completeness. In this previous work, the number of contacts optimised was set to 4 for the Solo robot \cite{grimminger2020open}. Here the ANYmal's robot dynamics are slower and the timing between each new contact is longer\footnote{The duration of a step was set to \SI{160}{\milli\second} on Solo whereas it is set to \SI{600}{\milli\second} for a walking gait and \SI{300}{\milli\second} for a trotting gait on ANYmal.}. This gives more time for computing the contact plan, allowing us to increase the planning horizon. We refer the reader to \cite{risbourg:hal-03594629} for empirical justifications for these choices. 

We first describe how the potential contact surfaces are pre-filtered to improve the algorithm's computational performance without loss of generality. We then provide the mathematical formulation of the Surface Selection algorithm. We conclude this section with the details of the cost function used in this optimisation problem.
\subsection{Pre-selection}
\label{sec::pre_selection}
\begin{figure*}[t]
    \subfloat[]{%
    \includegraphics[clip, trim=0 62px 0 0, width=0.325\textwidth]{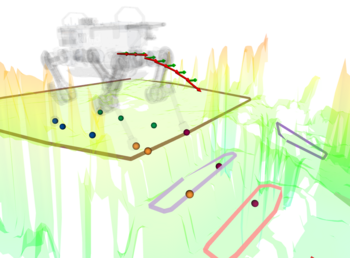}
    \label{fig::configs}}
    \subfloat[]{%
    \includegraphics[clip, trim=65px 0 0 0, width=0.325\textwidth]{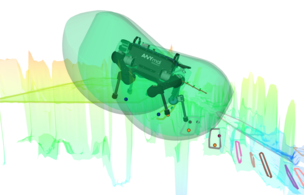}
    \label{fig::roms}}
    \subfloat[]{%
    \includegraphics[clip, trim=402px 0 0 70, width=0.325\textwidth]{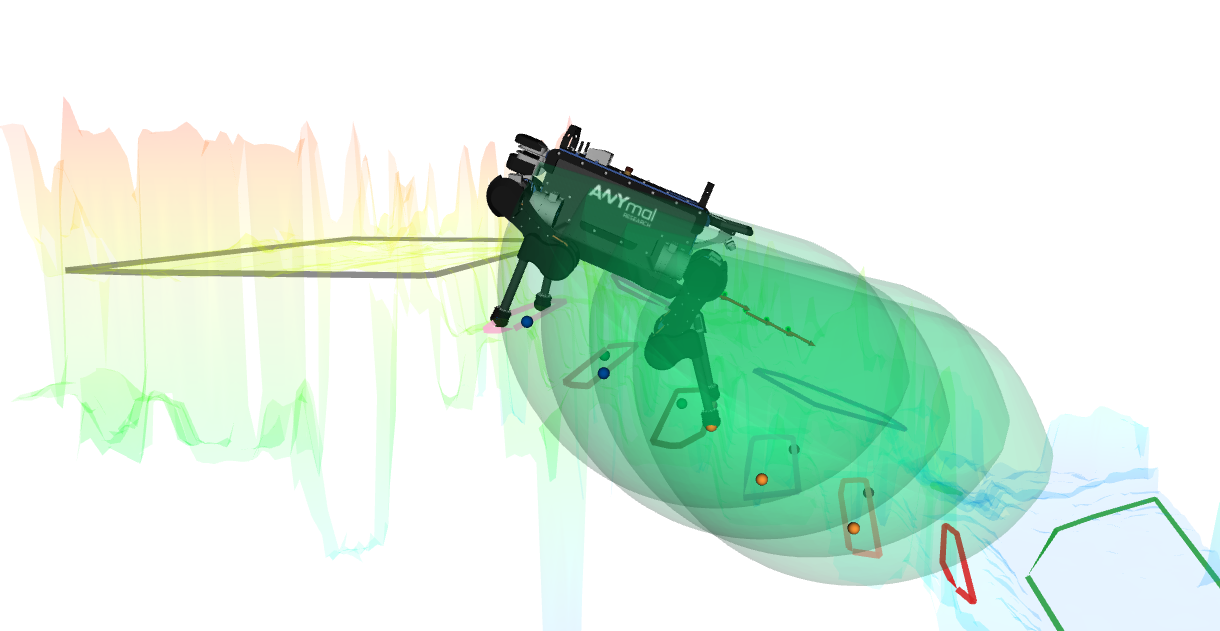}
    \label{fig::roms_FR}}
    \caption{(a) CoM extrapolation along the horizon. (b) ROMs of the 4 effectors for the current state. (c) ROMs along the horizon for the front right foot.}
    \label{fig:preselection}
\end{figure*}
We first pre-filter the number of contact surfaces using the robot's range of motion (ROM) to reduce the combinatorics. 
To do so, a CoM trajectory is extrapolated from the joystick velocity command (Fig.~\ref{fig::configs}) over the contact phases. The position of each state and the yaw angle (orientation around the z-axis) are integrated from the linear and angular (yaw only) desired velocity. The roll (orientation around the x-axis) and pitch (orientation around the y-axis) angles of the guide are computed as the average slope of the terrain around the robot position, given by solving (Fig.~\ref{fig::configs})
\begin{subequations}
    \label{eq:fiiting_plane}
     \begin{align*}
        \min_{\mathbf{n}} & \quad \| \mathbf{A}\mathbf{n} - \mathbf{B} \|^2 \\
        \textrm{with} \quad \forall \ i,j \in N_x \times N_y, \quad 
        &  \mathbf{A}[i + j N_x, \ :\ ] = [x_i,y_j,1.], \\
        & \mathbf{B}[i + j N_x] = \textrm{elevation($x_i$,$y_j$)},\\
    \end{align*}
\end{subequations}
where $x_i$ and $y_j$ are respectively the positions on the x and y axis around the robot's position with a user-defined resolution of $(N_x,N_y) \in \mathbb{N^+}^2$. The elevation function gives the terrain height at the position $(x_i,y_j)$ and can be obtained directly from the heightmap or by evaluating the convex plane corresponding to this 2D position. $\textbf{n} = [a,b,c] \in \mathbb{R}^3$ is the vector optimised to form the plane of equation $ax + by - z + c = 0$. The experiments use $10\times10$ as resolution. For each extrapolated state $\textbf{c}^*_j$ (8 states for a walking gait as $n = 8$  and 3 for a trotting gait as $n = 6$ but two feet contacts are created at each phase), the 6D configuration is given by:
\begin{equation}
    \label{eq:state_config}
    \textbf{c}^*_j = 
    \begin{bmatrix}
    x_0 + \int^{t_j}_0 (v_x^* \cos(\dot{\psi}^*t) + v_y^*\sin(\dot{\psi}^* t)) dt \\ 
    y_0 - \int^{t_j}_0 (v_x^* \sin(\dot{\psi}^*t) - v_y^*\cos(\dot{\psi}^* t)) dt \\ 
      ax_j + by_j + c + h_{\mathrm{ref}} \\ 
      \arctan(b) \\ 
      -\arctan(a)\\
      \psi_0 + \psi^* t_j
    \end{bmatrix},
\end{equation}
where the state is described using a 3D position $[ x_j, y_j, z_j ]$ and 3 Euler angles. The current extrapolated positions $x_j$ and $y_j$ are used to compute the configuration height. $v_x^*$, $v_y^*$ are the x-y linear velocities and $\dot{\psi}^*$ the yaw angular velocity coming from the joystick input. $h_{\mathrm{ref}} = 0.48$ is the robot's nominal height. $t_j$ is the step duration, manually defined for each gait. 

For each $\textbf{c}^*_j$, the ROM of each moving leg is intersected with the surfaces. The ROM is approximated by a convex set and represented by the green surfaces in Fig.~\ref{fig::roms}. Only the surfaces that intersect this ROM are selected. The distance between two convex sets is computed efficiently with the GJK algorithm \cite{GJK_algo} from the \textsc{Pinocchio} and \textsc{HPP-FCL} libraries \cite{carpentier2019pinocchio}. In our experiments, this  usually reduces the number of potential surfaces from 20 to 3 for each moving foot on average. This significantly reduces combinatorics (from about $20^n$ to $3^n$ possible combinations).

\subsection{Surface Selection algorithm as a MIP}
The surface selection module computes a contact plan for the robot that satisfies linearized kinematic constraints~\cite{tonneau:2pac,Wieber2006}. In the following, we recall the mathematical formulation of the problem as a mixed-integer program (MIP). This MIP outputs the contact surfaces selected. It also computes the 3D locations of footsteps as a by-product. However, they are discarded as the target footsteps will be adapted at a higher frequency by the footstep planner.

\begin{figure}
    \centering
    \includegraphics[width = \linewidth]{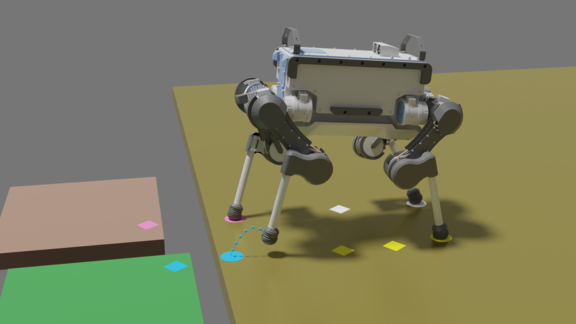}
    \vspace{-5mm}
    \caption{Environment $\mathcal{S}$ with 3 contact surfaces for a walking gait (1 foot moving at a time). Circles: initial position; squares: next steps locations.}
    \label{fig:surface-env}
\end{figure} 

\subsubsection{Contact constraint representation}
The perception pipeline provides a set of $m+1 \in \mathbb{N}^+$ disjoint quasi-flat surfaces, which define the environment $\mathcal{S}$, and the motion is decomposed into $n$ contact phases (Sec.~\ref{sec:def}), as shown in Fig.~\ref{fig:surface-env}.
To constrain a point $\mathbf{p} \in \mathbb{R}^3$ to be in contact, we must write:   
\begin{equation}
            \exists \ i,  \mathbf{S}^i \mathbf{p} \leq \mathbf{s}^i \Leftrightarrow \\
            \mathbf{S}^0 \mathbf{p} \leq \mathbf{s}^0  \vee \dots \vee \mathbf{S}^{m} \mathbf{p} \leq \mathbf{s}^{m}.
            \label{eq:or}
\end{equation}
The \textit{or} constraint is classically expressed as an integer constraint using the Big-M formulation \cite{bigm} as follows. We introduce a vector of binary variables $\mathbf{a} = [a_0, \dots, a_{m}] \in \{0,1\}^{m+1}$ and a sufficiently large constant  $M \in \mathbb{R}^{+}, M >> 0$. (\ref{eq:or}) is equivalently rewritten as:
\setlength{\belowdisplayskip}{0pt} \setlength{\belowdisplayshortskip}{0pt}
\setlength{\abovedisplayskip}{0pt} \setlength{\abovedisplayshortskip}{0pt}
\begin{equation}
            \forall i,  \mathbf{S}^i \mathbf{p} \leq \mathbf{s}^i + M (1-a_i) ;
            \sum_{i = 0}^{m} a_i = 1.
            \label{eq:orint}
\end{equation}
Under this formulation if  $a_i = 1$, then  $\mathbf{p}$ belongs to  $\mathcal{S}^{i}$ (and is thus in contact). Instead, if $a_i = 0$ for a sufficiently large M then the constraint $\mathbf{S}^i \mathbf{p} \leq \mathbf{s}^i + M (1-a_i)$ will be satisfied for any value of $\mathbf{p}$, in other words, the constraint is inactive. $\sum_{i = 0}^{n} a_i = 1$ implies that $ \exists i,  a_i = 1$. The obtained behaviour is thus the desired one: if (\ref{eq:orint}) is true then $\mathbf{p}$ is in contact with a surface.
\setlength{\belowdisplayskip}{\baselineskip} \setlength{\belowdisplayshortskip}{\baselineskip}
\setlength{\abovedisplayskip}{\baselineskip} \setlength{\abovedisplayshortskip}{\baselineskip}

\subsubsection{Problem formulation}
The final MIP problem is obtained by combining contact and surface constraints as follows. For simplicity, and without loss of generality, we assume that the candidate contact surfaces are the same for each step.
\begin{equation}
\label{eq:mixed_integer_feasibility}
\begin{aligned}
    \textbf{find} \quad & \mathbf{P}, \mathbf{A} = [\mathbf{a}_0,\cdots,\mathbf{a}_m] \in \{0,1\}^{(m+1)\times n}&  \\
    \textbf{min} \quad & l(\mathbf{P}) \\
    \textrm{s.t.} \quad & \mathbf{K} \mathbf{P} \leq \mathbf{k}, \\ 
    \quad &\mathbf{P} \in \mathcal{C}, \\
    & \forall j \in \{0, \dots, n-1\}: \\
    & \quad \forall i,  \mathbf{S}^i \mathbf{p}_j \leq \mathbf{s}^i + M (1-a_i^j); \sum_{i = 0}^{m} a_i^j = 1,
\end{aligned}
\end{equation}
where $\mathbf{P} = [{\mathbf{p}_0} \ \dots \ \mathbf{p}_n ] \in \mathbb{R}^{3\times n}$ is the vector comprising the variable $n$ next foot positions (6 or 8 in our experiments); $\mathbf{a}^j = [a_0^j, \dots, a_{m}^j]$ is the vector of binary variables associated with the $j$-th optimised foot position; $l(\mathbf{P})$ is an objective function; $\mathcal{C}$ is a set of user-defined convex constraints (in our case, constraints on initial contact positions); $\mathbf{K}$ and $\mathbf{k}$ are constant matrix and vector representing the linearised kinematic constraints on the position of each effector with respect to the others~\cite{Wieber2006,tonneau:2pac,winkler18}.

\subsubsection{Cost function details}
As mentioned in \cite{risbourg:hal-03594629}, two quadratic costs are used in the problem. The first attempts at regularising the footstep locations using Raibert's heuristic. Since the optimisation is triggered at the beginning of each step (between 3 and \SI{5}{\hertz}), and not at each time step (\SI{50}{\hertz}), an approximation of the Raibert heuristic is applied. The idea is to interpolate the base position given the desired velocity and place the foot accordingly to the estimated hip position. The second term penalises the distance between the hip and the foot location. This aims at penalising solutions close to reaching the robot's kinematic limits.
\begin{equation}
    l(\mathbf{P}) = \sum_j^n w_1 \| \mathbf{p}_j - \mathbf{p}_{j}^* \|^2_{x,y} + w_2  \| \mathbf{p}_j - \mathbf{p}_{\textrm{hip},j}^* \|^2 ,
\end{equation}
where $\mathbf{p}_{_j}^*$ is the extrapolated foot position taking into account the linear and angular reference velocity at the corresponding contact phase.  The weights $w_1$ and $w_2$ are set to 1 and 0.5.  This penalisation only considers the $\ell^2$ norm on the x- and y-axis. Similarly, $\mathbf{p}_{\textrm{hip},j}^*$ is the extrapolated hip position from the desired velocity. The computation of the reference foot location is detailed in  Sec.~\ref{sec:foot_trajectory_generation}.  


\section{Foot trajectory generation}
\label{sec:foot_trajectory_generation}

Our \textit{Footstep and Collision free-trajectory} (Fig.~\ref{fig::archi_overview} - red) algorithms compute the end-effector trajectory as a Bezier curve given the current robot state (COM position/orientation), the next moving foot within the horizon of the MPC, the contact plan obtained by the Surface Planner, the timings of contact depending on the gait chosen, as well as a desired target velocity for the robot. 
At the MPC frequency, set to \SI{50}{\hertz}, a first quadratic program (QP) computes the next foot location and a second QP calculates its trajectory during the flight. Their formulation differs from our previous work \cite{risbourg:hal-03594629} as the base velocity is not optimised here. The total computation time for the foot trajectory is negligeable in the pipeline, as evidenced by Table~\ref{tab::info-surface-planning}.


\subsection{Foot position optimisation}

We use an alternative to Raibert's heuristic~\cite{raibert1986legged} as it is commonly done on quadruped robots~\cite{WBIC,Cheetah3-convex,leziart:hal-03052451}. In our case, we are simply interested in coherent foot locations according to the base position, reference velocity and gait period. The measured velocity is not part of this heuristic, making it a feed-forward strategy. Instead, our MPC plans the CoM trajectory (Sec.~\ref{sec::motion_generation}). A target 2D position of the foot $\mathbf{p^*}$ is computed as follows:
\begin{equation}
    \mathbf{p^*} = \mathbf{p}_{\textrm{hip}} + 
    \frac{T_s}{2} \mathbf{v}_{ref} +
    \sqrt{\frac{h}{g}} \mathbf{v}_{ref} \times \mathbf{\omega}_{ref},
\end{equation}
where $T_s$ is the stance phase time extracted from the phase-based gait pattern, $\mathbf{v}_{\textrm{base}}$, and $\mathbf{v}_{\textrm{ref}}$ are respectively the current base velocity and the reference velocity commanded by the user. Finally, $h$, $g$ and $\mathbf{\omega}_{\textrm{ref}}$ are respectively the nominal height of the robot, the gravity constant and the reference angular velocity around the z-axis. The estimated hip position at contact $\mathbf{p}_{\textrm{hip}}$ accounts for the accumulated delay in the estimation of the base position, since the moving average is used to reject oscillations of the same period as gait, as discussed in section \ref{sec::base_position_ref_vel}. The final position $\mathbf{p}$ is the closest to $\mathbf{p}^*$ that lies on the selected surface $\mathcal{S}$ under the locally-approximated kinematic constraints $\mathcal{K}$:
\begin{subequations}
 \begin{align}
    \min_{\mathbf{p}} &   \quad \| \mathbf{p} - \mathbf{p^*} \|^2 \\
    \quad \textrm{s.t.} & \quad  S  \mathbf{p} \leq s, \\
                             & \quad  \mathbf{p} \in \mathcal{K}.
\end{align}
\end{subequations}
The difference with our MIP optimisation lies in the \SI{50}{\hertz} computation frequency. As a result, the footstep is constantly updated relative the base position.

\subsection{Collision free trajectory}
\label{sec::collision-free-traj}

\begin{figure}[h]
    \centering
    \includegraphics[width = 0.6\linewidth, keepaspectratio = true]{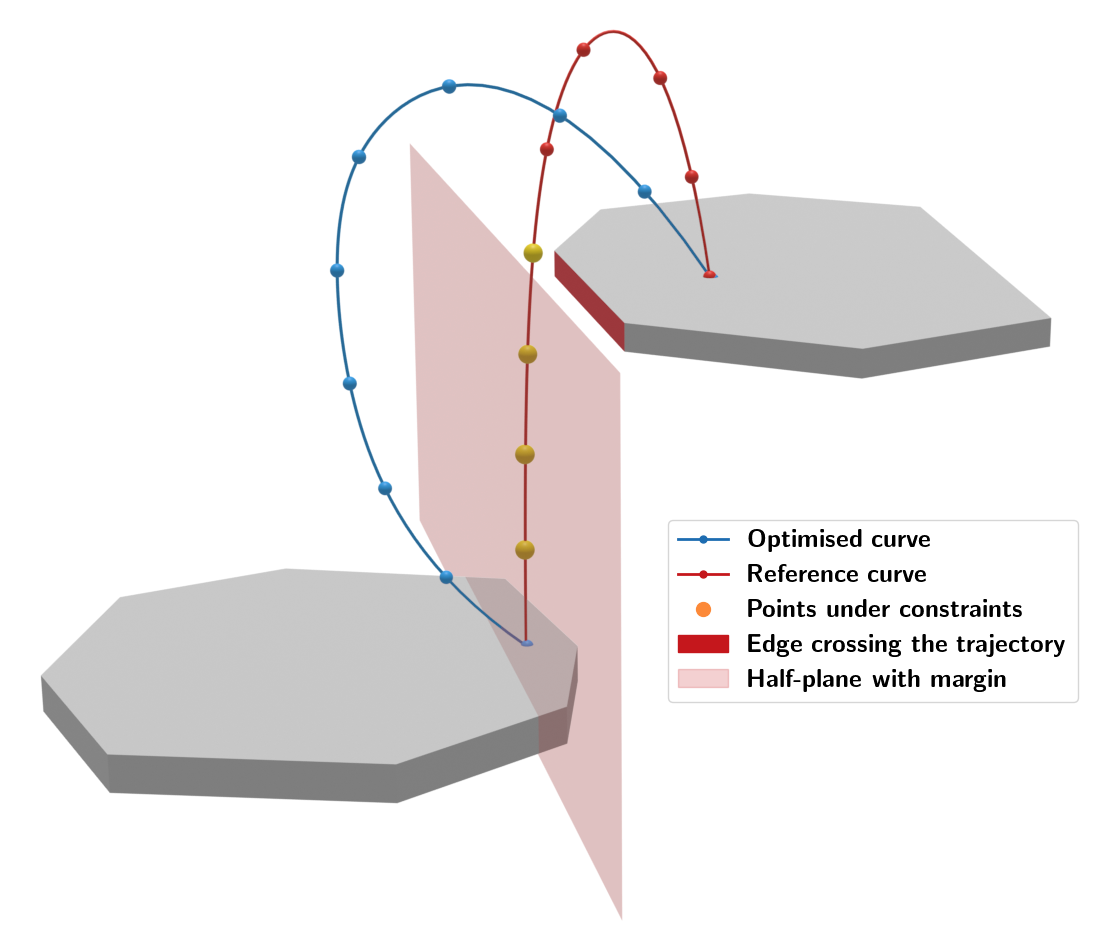}
    \caption{Adaptation of the end-effector trajectory to climb a step. The active constraint corresponds to the half-plane crossed by the reference trajectory -red curve-. Both curves are discretised to perform the optimisation and the orange points highlight the ones under constraints.}
    \label{fig::bezier-curve}
\end{figure}


A collision-free trajectory  $\mathbf{p}(t): \mathbb{R} \rightarrow \mathbb{R}^3$ connects the current foot location to the optimised location.  $\mathbf{p}(t)$ aims at following a reference trajectory while avoiding obstacles and keeping computational time low as it is parameterized using a Bezier curve of degree d on the Bernstein basis $(B_i)_{i \leq d}$:
\begin{equation}
\mathbf{p}(t) = \sum^d_{i=0}  B_i^d(\frac{t}{T}) \mathbf{P}_i,
\end{equation} 
where $\mathbf{P} = \begin{bmatrix} P_0 & \dots & P_d \end{bmatrix}^T$ are the $d+1$ control points and $T$ is the total time of the trajectory, empirically set to 600 ms.
\newline
\subsubsection{Reference trajectory}
The trajectory $\mathbf{p}_{ref}(t): \mathbb{R} \rightarrow \mathbb{R}^3$ is composed of a degree 6 polynomial curve of degree 6 on the z-axis and a degree 5 one on the x and y axes. All of which have with the first 3 control points fixed to ensure the continuity in position, velocity and acceleration of the curve from the current state. End velocity and acceleration are set to 0 to avoid slippage at the end position. Additionally, the height at $\frac{T}{2}$ is fixed on the z-axis, constraining all degrees of freedom.


\subsubsection{Collision avoidance} 
The trajectory  $\mathbf{p}(t_k)$ is a 3D curve of degree $d=7$ that follows $\mathbf{p}_{ref}(t)$ while avoiding collisions.
To do so, we enforce collision avoidance along the foot trajectory. This is achievable iteratively~\cite{campana} by solving a sequence of QPs and adding constraints where collisions occur. For computational efficiency, we empirically identify the points likely to be in a collision and add collision constraints to them (Fig.~\ref{fig::bezier-curve} - Yellow dots). 
These $n_c+1$ points $\mathbf{p}(t_k) = \mathbf{A}_k \mathbf{P}, \forall k \in [0,\dots,n_c]$ are linearly defined by Bezier's control points, with $\mathbf{A}_k \in \mathbb{R}^{3 \times (d+1)}$.
 We choose the active constraint to be the half-space traversed by the reference curve (Fig.~\ref{fig::bezier-curve} - pink half-space).
 We thus write $ \forall k, \mathbf{S}^i_m \mathbf{p}(t_k) \geq \mathbf{s}^i_m$, where $\mathbf{S}^i_m \in \mathbb{R}^3$ and $\mathbf{s}^i_m \in \mathbb{R}$ define what constitutes the current half-space. 
 All the collected constraints are then stacked into a single matrix and vector $\mathbf{G}$ and $\mathbf{h}$, leading to the QP:
\begin{subequations}
 \begin{align}
    \min_{\mathbf{p}} & \quad \sum_{k=0}^{ n_c} \| \mathbf{p}(t_k) - \mathbf{p}_{\textrm{ref}}(t_k) \|^2 
    \\
    \quad \textrm{s.t.} 
    & \quad  \mathbf{p}(0) = \mathbf{p}_{\textrm{ref}}(0), 
     \quad \quad  \mathbf{p}(T) = \mathbf{p}_{\textrm{ref}}(T),
    \\
    & \quad  \Dot{\mathbf{p}}(0) = \Dot{\mathbf{p}}_{\textrm{ref}}(0),    \quad \quad  \Dot{\mathbf{p}}(T) = \mathbf{0}_{\mathbb{R}^3},
    \\
    & \quad  \Ddot{\mathbf{p}}(0) = \Ddot{\mathbf{p}}_{\textrm{ref}}(0),
     \quad \quad \Ddot{\mathbf{p}}(T) = \mathbf{0}_{\mathbb{R}^3},
    \\
    & \quad  \mathbf{G} \mathbf{P} \leq \mathbf{h}.
\end{align}
\end{subequations}
This approach can be expanded to avoid local obstacles along the trajectory by imposing constraints that depend on the environment's heightmap. Currently, our trajectory avoidance is based only on obstacles perceived as walkable surfaces, which has been found to work effectively in many scenarios. Nevertheless, if an obstacle is along the trajectory but not detected as such, it will not be avoided. This highlights the importance of including height-map information when available.

\section{Motion Generation}
\label{sec::motion_generation}

Given the current state of the robot (6D position and velocity), the next contact sequence  and the end-effector trajectories, our \textit{Whole-Body MPC} (WB-MPC) and \textit{Riccati gain controller} elements (Fig.~\ref{fig::archi_overview} - red and violet) generates the motion of the legged robot. The WB-MPC optimises a trajectory at \SI{50}{\hertz} and the Riccati gain controller applies it at \SI{400}{\hertz}. 

\subsection{Optimal control formulation}

The legged robot generates motions through a model predictive controller that relies on the robot's full-body dynamics. The formulation is based on two previous papers \cite{Agile-Maneuvers-mastalli, Inv-Dyn-MPC-mastalli}. We solve the optimal control (OC) problems using Crocoddyl's advanced solvers~\cite{crocoddyl20icra}. Our pipeline relies on two different OC formulations based on forward and inverse dynamics. Both yield similar performances, demonstrating our method's generality. Both formulations can be written as \cite{Agile-Maneuvers-mastalli, Inv-Dyn-MPC-mastalli}:
\begin{subequations} \label{eq:ocp_general}
 \begin{align}
\min_{ \{\mathbf{q}, \mathbf{v}\}, 
       \{\boldsymbol{\tau}\} \mathrm{or} \{\mathbf{\Dot{v}},\boldsymbol{\lambda}\}}  
& \displaystyle \ \ell_N(\mathbf{x}_N) + \displaystyle \sum_{k=0}^{N-1} \int_{t_k}^{t_k+1} \ell_k(\mathbf{q}_k,\mathbf{v}_k, \boldsymbol{\lambda}_k, \boldsymbol{\tau}_k) \ dt  \\
\nonumber 
\textrm{s.t.} \quad \quad  
\hspace{-5mm} \quad \mathbf{q}_{k+1} & = \ \mathbf{q}_k \ \oplus \ \int_{t_k}^{t_{k+1}} \mathbf{v}_{k+1} \ dt, \\
\nonumber 
 \hspace{-5mm} \quad \mathbf{v}_{k+1} & = \ \mathbf{v}_k \ + \ \int_{t_k}^{t_{k+1}} \mathbf{\Dot{v}}_k \ dt, \\
\nonumber 
\hspace{-5mm} \quad 
\begin{bmatrix} \Dot{\mathbf{v}_k} \\ 
- \boldsymbol{\lambda}_k \end{bmatrix} & = \
\begin{bmatrix} 
\mathbf{M}_k & \mathbf{J}_c^\top \\ 
\mathbf{J}_c &
\end{bmatrix}^{-1}
\begin{bmatrix} 
\boldsymbol{\tau}_b \\ 
-\mathbf{a}_c,
\end{bmatrix}
\ \textrm{(forward dyn.)} \\ \nonumber
\textrm{or} & \\
\nonumber
& \hspace{-5mm} \textrm{ID}(\mathbf{q}_k,\mathbf{v}_k,\mathbf{\Dot{v}}_k,\boldsymbol{\lambda}_k) = \mathbf{0},  
\quad \textrm{(inverse dyn.)} \\ \nonumber
\end{align}
\end{subequations}
\noindent
where the robot state $\mathbf{x} = (\mathbf{q}, \mathbf{v})$ contains the generalized position and velocity vectors. More precisely, $\mathbf{q} \in \mathbb{SE}(3) \times \mathbb{R}^{n_j}$ with the generalized velocity lying in the tangent space $\mathbf{v} \in \mathfrak{se}(3) \times \mathbb{R}^{n_j}$ where $n_j$ is the number of articulated joints and $\oplus$ denotes the integration operator in $\mathbb{SE}(3)$ inspired by \cite{blanco2010se3} and used in \cite{crocoddyl20icra}.

For the forward-dynamics formulation, the input command is $\mathbf{u} = \{ \boldsymbol{\tau} \}$ corresponds to the joint torques. $k$, $\mathbf{M}_k$ and $\mathbf{J}_c$ corresponds to the discrete-time $t_k$, the generalised mass matrix and the contact Jacobian, respectively. $\boldsymbol{\tau}_b$ includes the joint torque command, Coriolis and gravitation terms and $\mathbf{a}_c$ is the desired acceleration which results from the rigid contact constraint (contact point velocity is null) and includes the Baumgarte stabilisation term \cite{Baumgarte}. The impulse dynamics described in more detail in \cite{Agile-Maneuvers-mastalli} which allow velocity changes at impact have also been omitted from the formulation. This contact dynamics formulation comes from the application of the Gauss principle of least constraint \cite{Wieber2006}, as described in  detail in~\cite{DDP_contact_dynamic-Budhiraja}. By handling this constraint in the backward pass, the decision variables can be condensed with forward dynamics \cite{Agile-Maneuvers-mastalli,DDP_contact_dynamic-Budhiraja,crocoddyl20icra} and contact forces can be removed. This drastically reduces the number of decision variables.

Regarding the inverse-dynamics formulation, where $\mathbf{u} = \{ \mathbf{\Dot{v}},\boldsymbol{\lambda} \}$, the contact forces and the acceleration become decision variables of the problem \cite{Inv-Dyn-MPC-mastalli, Erez2012TrajectoryOF}. We can perform efficient factorizations by applying nullspace parametrization as proposed in \cite{Inv-Dyn-MPC-mastalli}. Furthermore, an alternative inverse dynamic equation that allows us to remove joint torques from the control vector is possible. Both strategies reduce the size of the problem needed to enable MPC applications.



\begingroup
\renewcommand{\arraystretch}{1.5} 

\begin{table}[h]
\centering
\caption{Details of the costs used in the OCP}
\label{tab::costs_detail}
\begin{tabular}{l c c}
    \multicolumn{1}{c}{\textbf{Name}} & \textbf{Formulation} & \textbf{Cost}\\
    \hline\hline
     State bound & $\mathbf{x}_{\textrm{min}} \leq \mathbf{x} \leq  \mathbf{x}_{\textrm{max}}$ & $10^3$ \\
     
     Base orientation reg. &  $\log (\mathbf{p}_{\textrm{base}}^{-1})$ & $[0_{\mathbb{R}^3} \ 10^2 \ 10^2 \ 0 ]$ \\
     
     Base velocity reg. &  $\| \mathbf{\Dot{p}_{base}}\|$ & $10$ \\
     
     Joint position reg. & $\| \mathbf{q_a}_k -  \mathbf{q_a}_{\textrm{ref}} \|$ & $0.01$ \\ 
     
     Joint velocity reg. & $\| \mathbf{\Dot{q}}_{\mathbf{a}k} \|$ & $1$ \\
     
     Torque reg. & $\| \mathbf{u}_k \|$ & $0.1$ \\
     
     Force reg. & $\| \boldsymbol{\lambda}_{\mathcal{C}_k} \|$ & $1$ \\
     
     Friction cone & $C \boldsymbol{\lambda}_{\mathcal{C}_k} \leq c$ & $10$ \\
  
    Feet position track. & $\log (\mathbf{p}^{-1}_{\mathcal{G}_k} \mathbf{p}^{}_{\textrm{ref},\mathcal{G}_k} )$ & $10^6$\\
    
    Feet velocity track. & $\mathbf{\Dot{p}}^{}_{\textrm{ref},\mathcal{G}_k} - \mathbf{\Dot{p}}^{}_{\mathcal{G}_k}$ & $10^4$\\
\end{tabular}
\end{table}
Regularisation terms included in the cost function $\ell$ are summarized in Table~\ref{tab::costs_detail}. These different formulations are transcribed into quadratic terms, where $\mathbf{q_a}, \mathbf{\Dot{q}_a} \in \mathbb{R}^{12}$ are respectively the actuated joint positions and velocities. $\mathbf{p}_{\textrm{base}} \in \mathbb{SE}(3)$ describes the base pose in the world frame. $\mathbf{p}^{}_{\mathcal{G}_k} \ominus \mathbf{p}^{}_{\textrm{ref},\mathcal{G}_k} = \mathbf{p}^{-1}_{\mathcal{G}_k} \mathbf{p}^{}_{\textrm{ref},\mathcal{G}_k}$ represents the pose error in $\mathbb{SE}(3)$ of the feet relative to the reference position. It is penalised by considering it in its tangent space \cite{blanco2010se3} with the $\log$ function that maps $\mathbb{SE}(3)$ to $\mathfrak{se}(3)$. Feet position and velocity tracking error are expressed in the robot's inertial frame. The base position is not penalised whereas its rotation is penalised around the x- and y-axis only, corresponding to the roll and pitch angles. $\mathbf{C}$ and $\mathbf{c}$ are the matrix and vector of the linearised friction cone. $\mathcal{C}_k$ represents the set of feet in contact with the ground and inversely $\mathcal{G}_k$ is the set of feet in the swing phase for the node k. Costs are set to zero if the corresponding feet are in the contact phase. 

\subsection{Riccati-gain controller}
OC formulations with reduced-order dynamics require an instantaneous whole-body controller for tracking computed forces and maintaining robot balance~\cite{Mastalli-motion-planning, leziart:hal-03052451, WBIC}. Such instantaneous controllers may compete with the MPC policy and not necessarily generate the motion predicted by them \cite{improved_controller_leziart}. At a higher cost in computing time, which has a non-negligible influence, many benefits can appear when including whole-body dynamics in the OC problem, e.g., imposing the joint effort limits. Another benefit is to derive local feedback controllers from optimisation principles. This control policy looks like this:
\begin{subequations}
 \begin{align}
    \boldsymbol{\tau}_d & = \boldsymbol{\tau}_{ff} + \mathbf{K} (\mathbf{x} \ominus \mathbf{x^*}),\\
           & = - \mathbf{Q_{uu}}^{-1} \mathbf{Q_{u}} -\mathbf{Q_{uu}}^{-1} \mathbf{Q_{ux}} (\mathbf{x} \ominus \mathbf{x^*}),
\end{align}
\end{subequations}
where $\ominus$ denotes the $\mathbb{SE}(3)$ error. $\mathbf{Q_{uu}}$, $\mathbf{Q_{u}}$ and $\mathbf{Q_{ux}}$ are respectively the partial derivatives of the value function for the joint-effort command and the state \cite{Agile-Maneuvers-mastalli}. While the MPC runs every \SI{0.02}{\second} (\SI{50}{\hertz}), the discretisation is set to \SI{0.01}{\second}. Instead, the low-level Riccati-gain controller runs at \SI{400}{\hertz}, and an interpolation step is necessary. For this, the feedback term is computed by using an interpolation of the reference state $\mathbf{x^*}$ by integrated the dynamic with the contact model dynamic \ref{eq:ocp_general}. The optimal effort command is then provided at \SI{400}{\hertz} in addition to a joint impedance controller based on the joint command and velocity obtained from $\mathbf{x^*}$.
\section{Results}
\label{sec::results}

\begingroup
\newcommand{\widthFig}{0.195\textwidth}
\setlength{\tabcolsep}{0.25pt} 
\renewcommand{\arraystretch}{1.5} 
\setlength\arrayrulewidth{1.pt}
\newlength{\tempdima}
\newcommand{\rowname}[1]{\rotatebox{0}{\makebox[\tempdima][c]{(#1)}}}

\makeatletter
\newcommand{\newtag}[2]{#1\def\@currentlabel{Fig#1}\label{#2}}
\makeatother

\newcommand{\subFigA}[1]{\raisebox{-.5\height}{\includegraphics[clip,trim=100px 16px 83px 66px,width=\widthFig]{{#1}}}}
\newcommand{\subFigB}[1]{\raisebox{-.5\height}{\includegraphics[clip,trim=106px 16px 83px 83px,width=\widthFig]{{#1}}}}
\newcommand{\subFigC}[1]{\raisebox{-.5\height}{\includegraphics[clip,trim=140px 50px 83px 50px,width=\widthFig]{{#1}}}}
\newcommand{\subFigD}[1]{\raisebox{-.5\height}{\includegraphics[clip,trim=100px 16px 83px 66px,width=\widthFig]{{#1}}}}
\newcommand{\subFigE}[1]{\raisebox{-.5\height}{\includegraphics[clip,trim=400px 250px 600px 200px,width=\widthFig]{{#1}}}}
\newcommand{\subFigF}[1]{\raisebox{-.5\height}{\includegraphics[clip,trim=420px 300px 570px 90px,width=\widthFig]{{#1}}}}
\newcommand{\subFigG}[1]{\raisebox{-.5\height}{\includegraphics[clip,trim=100px 16px 83px 16px,width=\widthFig]{{#1}}}}
\newcommand{\subFigH}[1]{\raisebox{-.5\height}{\includegraphics[clip,trim=300px 100px 250px 150px,width=\widthFig]{{#1}}}}

\def\boxit#1#2{%
  \smash{\llap{\rlap{\hspace{4pt}\href{#1}{\strut\raisebox{-.5\height}{\phantom{\rule{0.98\textwidth}{#2}}} } }~}}\ignorespaces}
\hypersetup{pdfborder=0 0 0}

\begin{figure*}
\centering
\begin{tabular*}{\textwidth}{c@{\hskip 7pt} c c c c c}
\rowname{a}&
\boxit{\videoLink{210}}{62pt}%
\subFigA{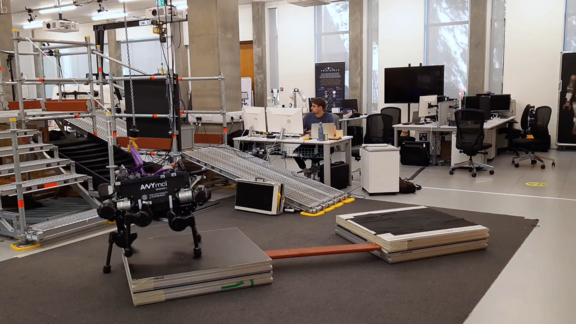}&
\subFigA{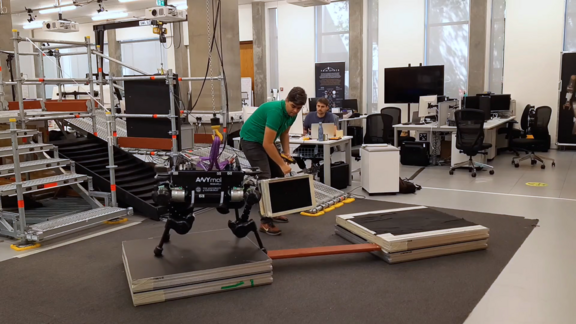}&
\subFigA{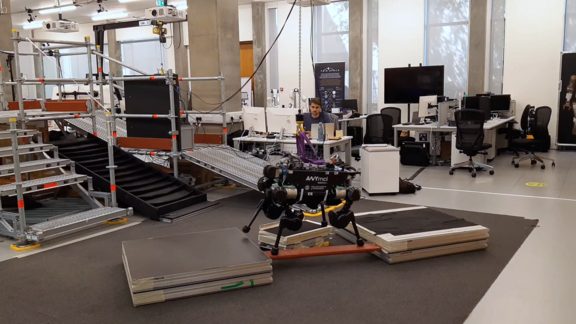}&
\subFigA{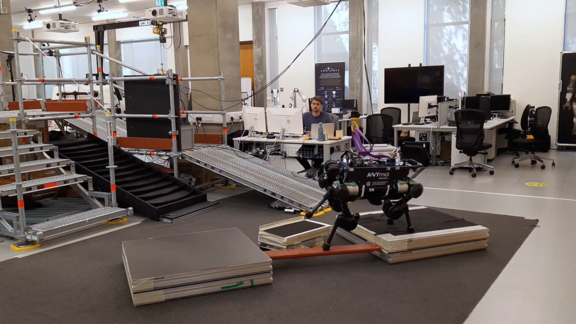}&
\subFigA{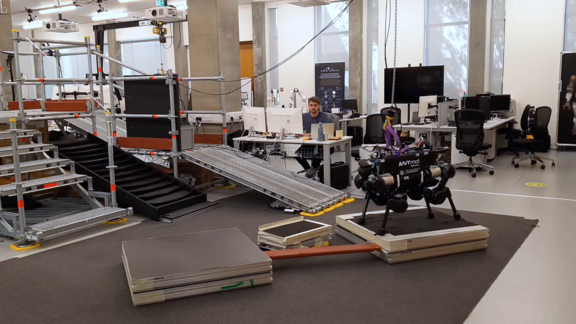}\\
\\[-2.5ex]
\cline{2-6}
\\[-2.2ex]
\rowname{b}&
\boxit{\videoLink{15}}{60pt}%
\subFigB{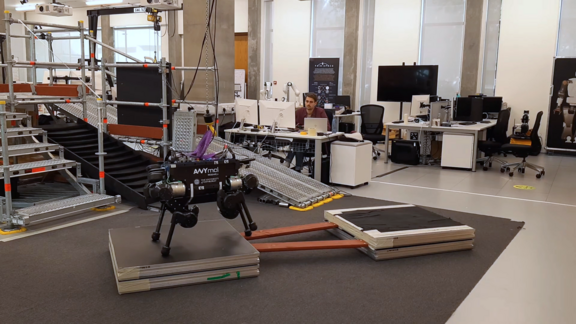}&
\subFigB{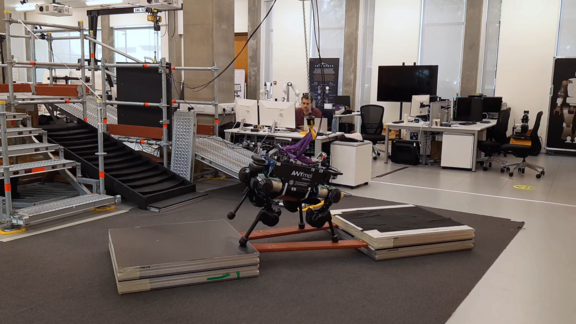}&
\subFigB{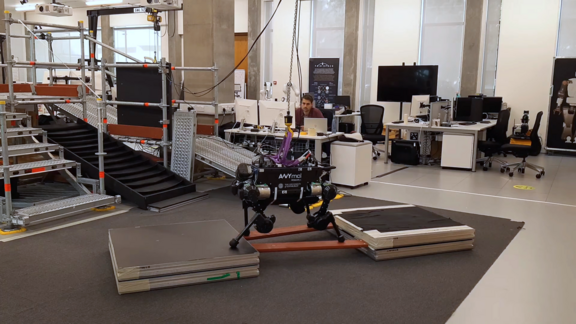}&
\subFigB{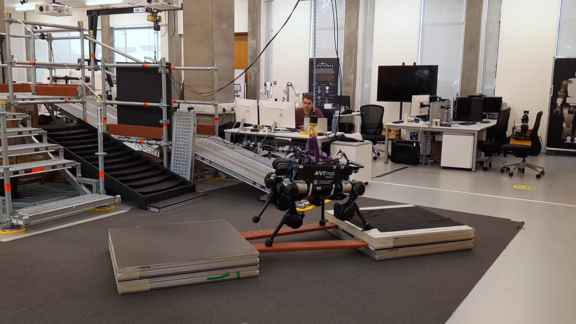}&
\subFigB{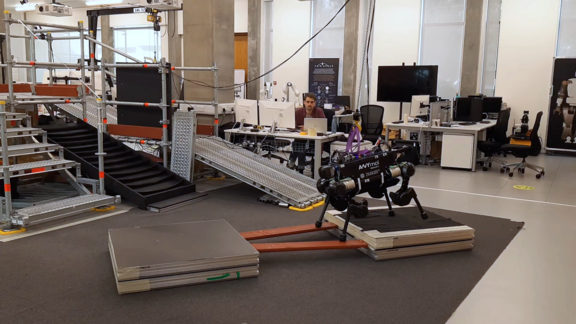}\\
\\[-2.5ex]
\cline{2-6}
\\[-2.2ex]
\rowname{c}&
\boxit{\videoLink{50}}{70pt}%
\subFigC{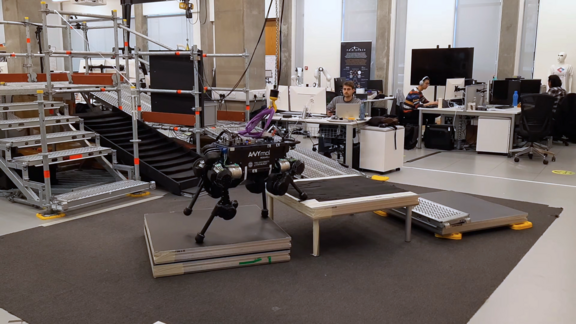} &
\subFigC{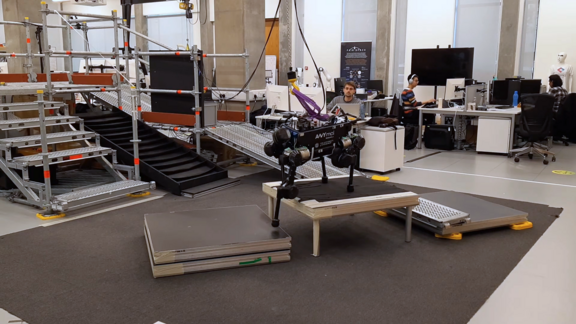} &
\subFigC{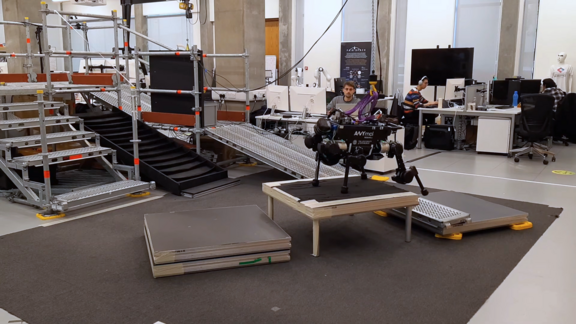} &
\subFigC{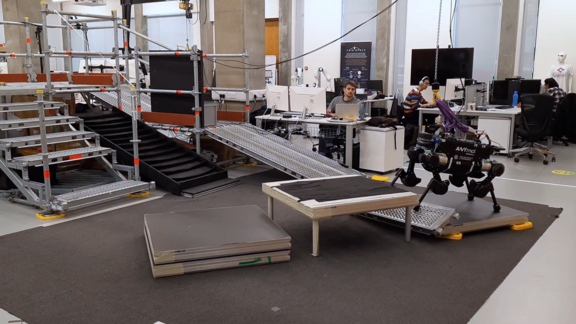} &
\subFigC{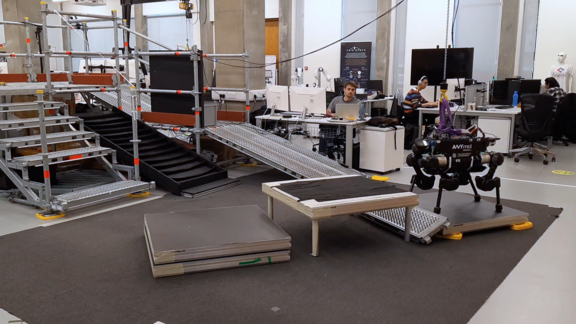}\\
\\[-2.5ex]
\cline{2-6}
\\[-2.2ex]
\rowname{d}&
\boxit{\videoLink{275}}{65pt}%
\subFigD{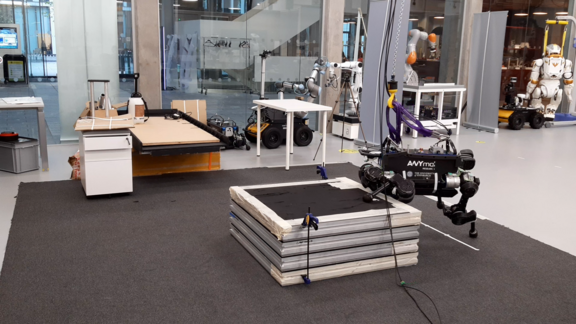} &
\subFigD{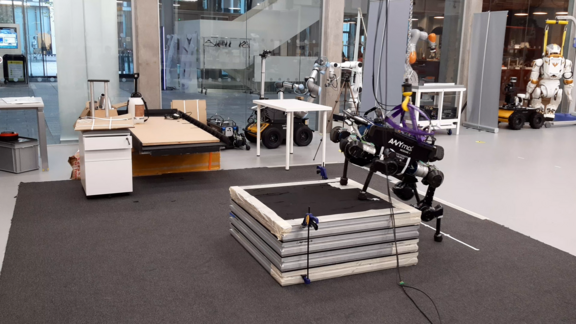} &
\subFigD{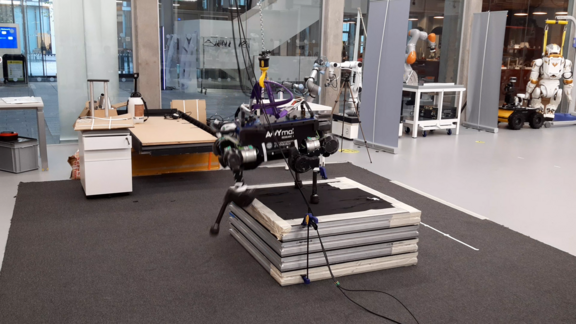} &
\subFigD{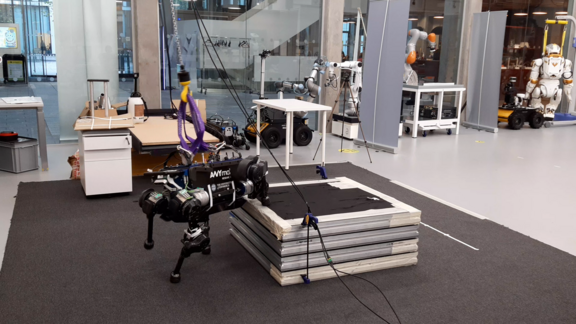} &
\subFigD{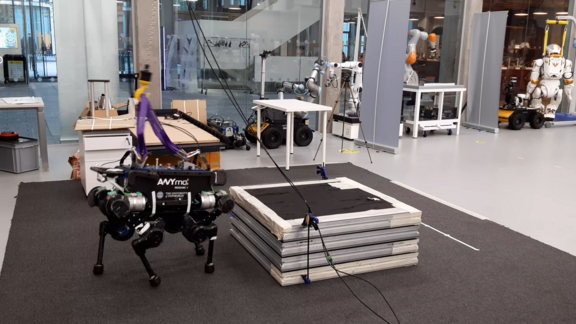}\\
\\[-2.5ex]
\cline{2-6}
\\[-2.2ex]
\rowname{e}&
\boxit{\videoLink{275}}{80pt}%
\subFigE{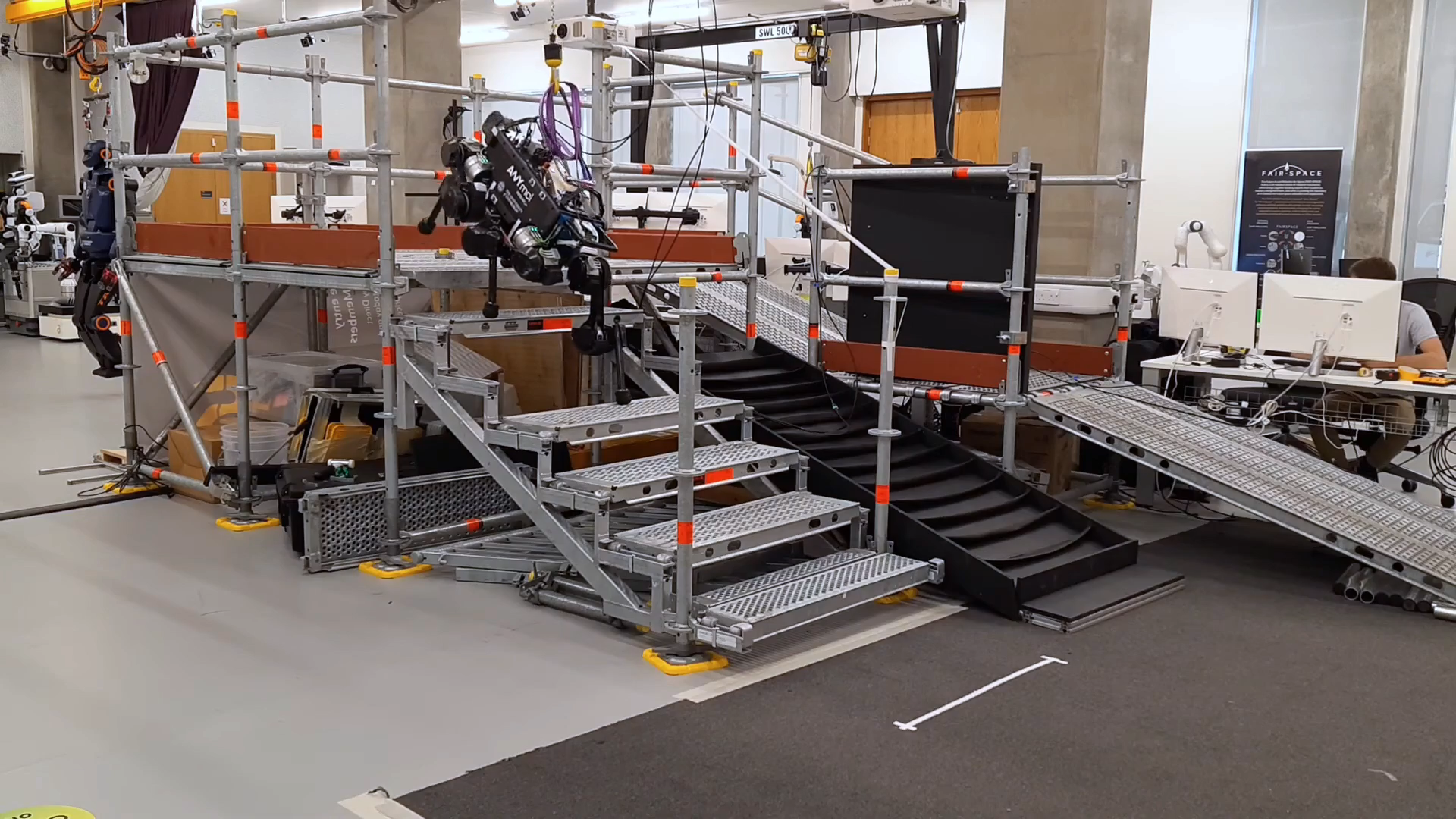} &
\subFigE{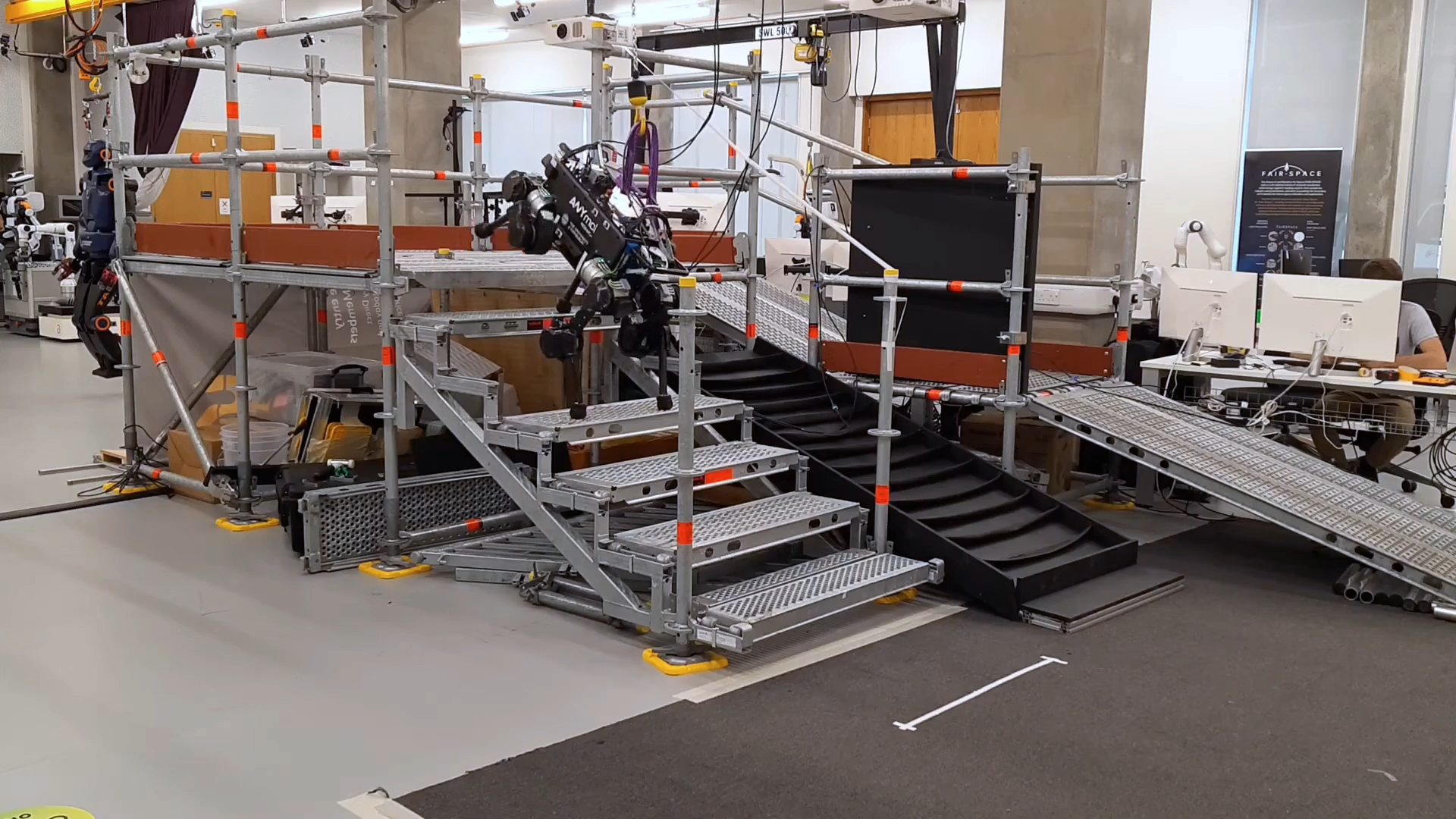} &
\subFigE{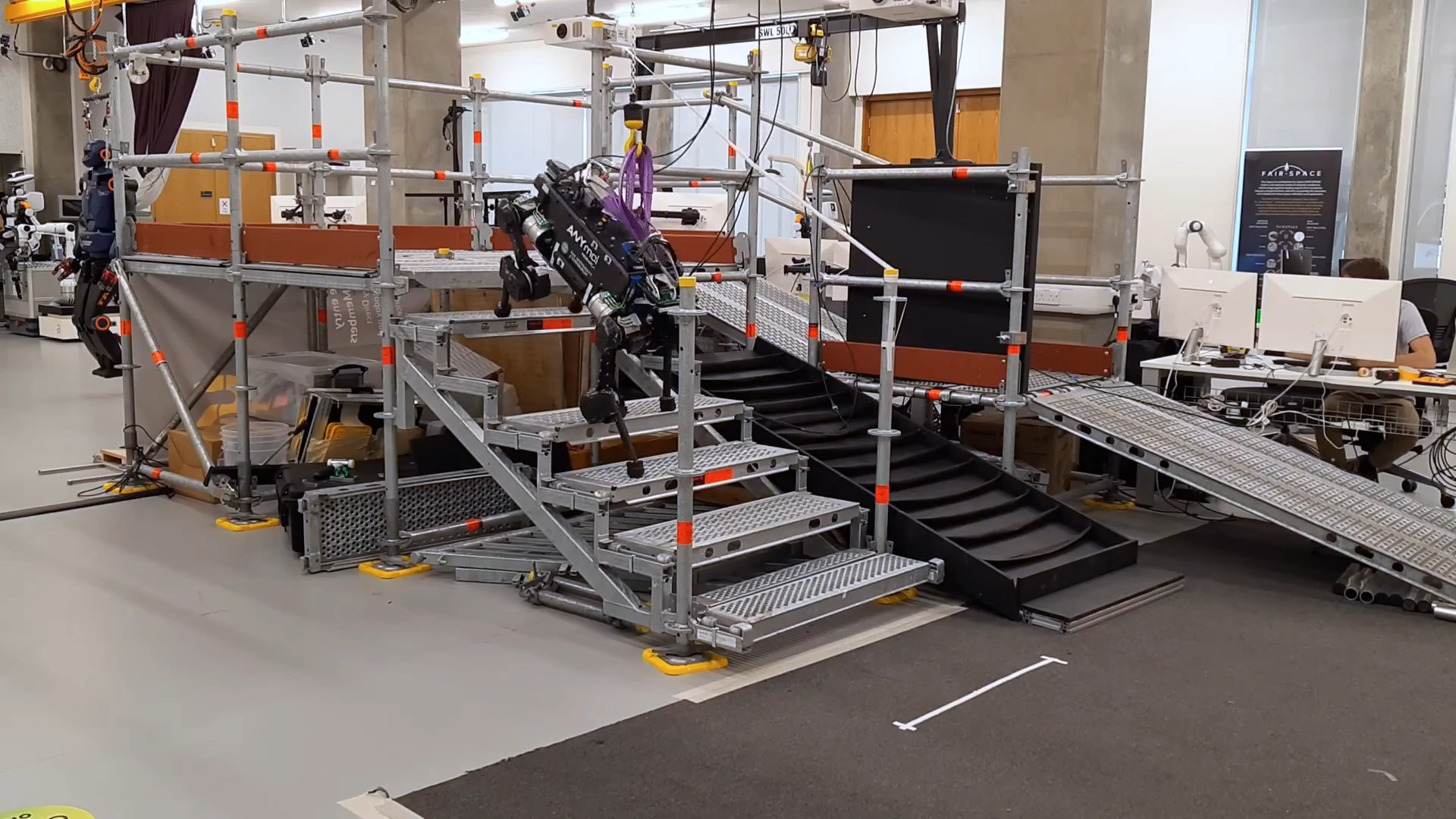} &
\subFigE{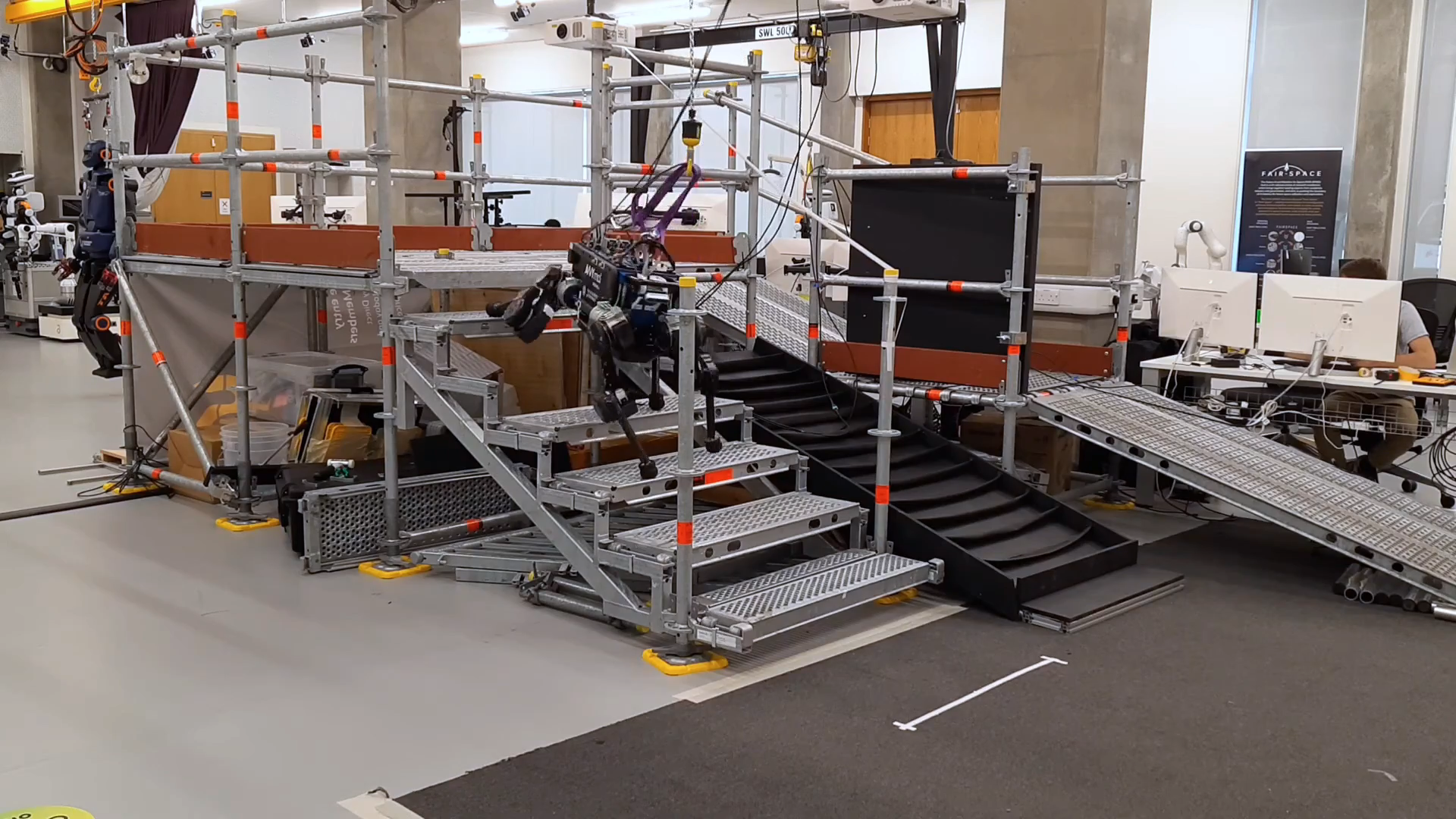} &
\subFigE{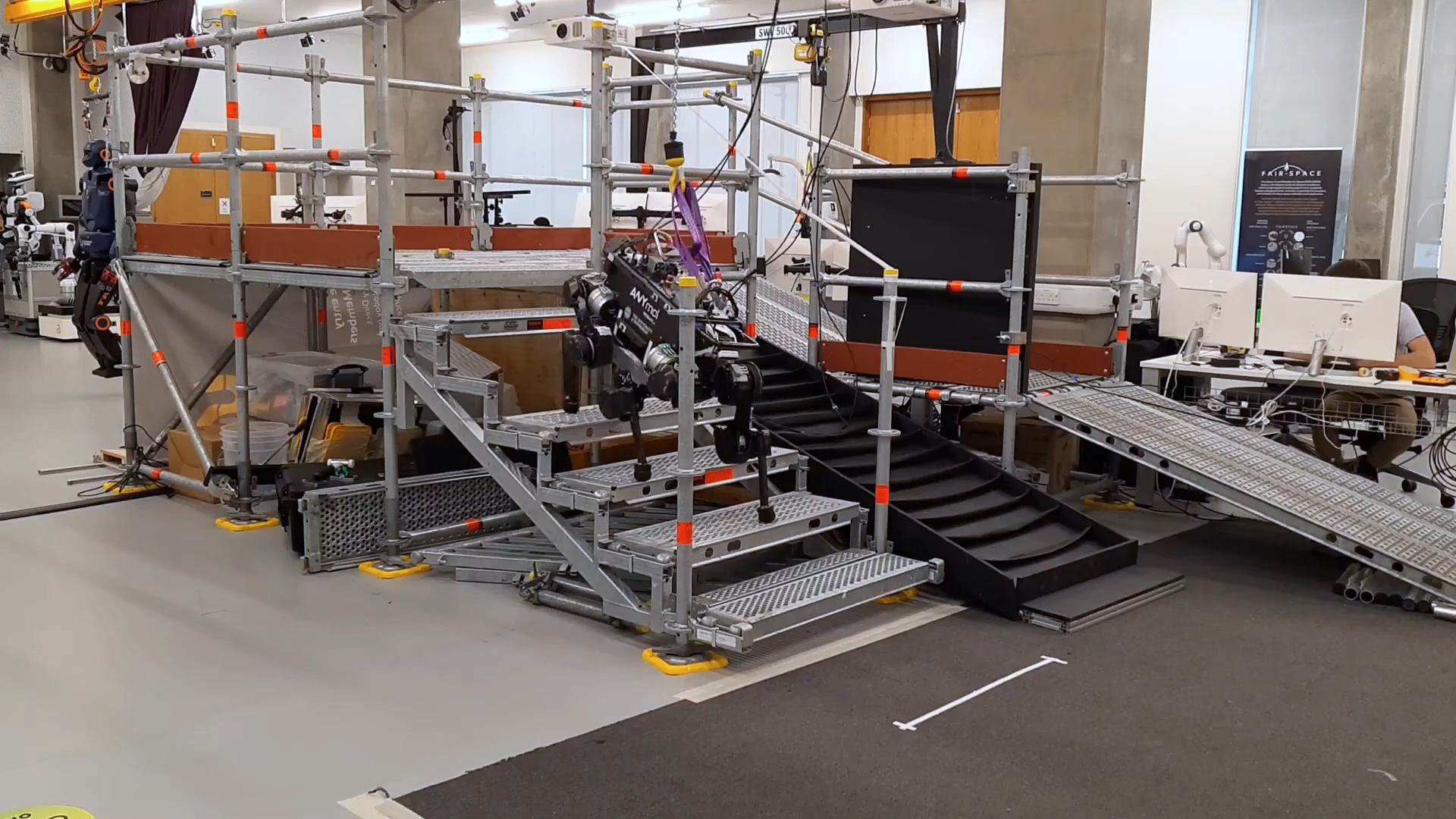}\\
\\[-2.5ex]
\cline{2-6}
\\[-2.2ex]
\rowname{f}&
\boxit{\videoLink{320}}{90pt}%
\subFigF{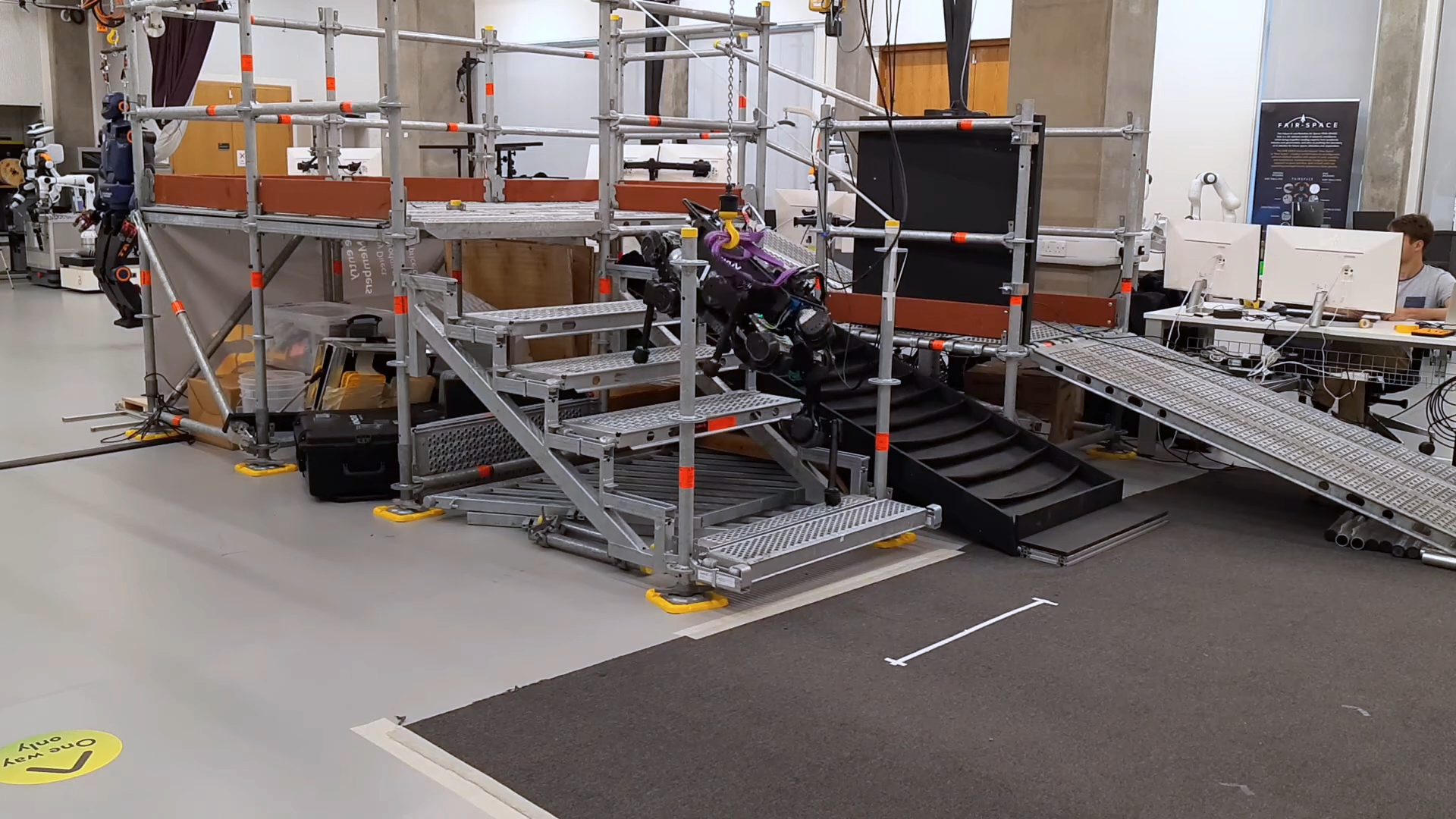} &
\subFigF{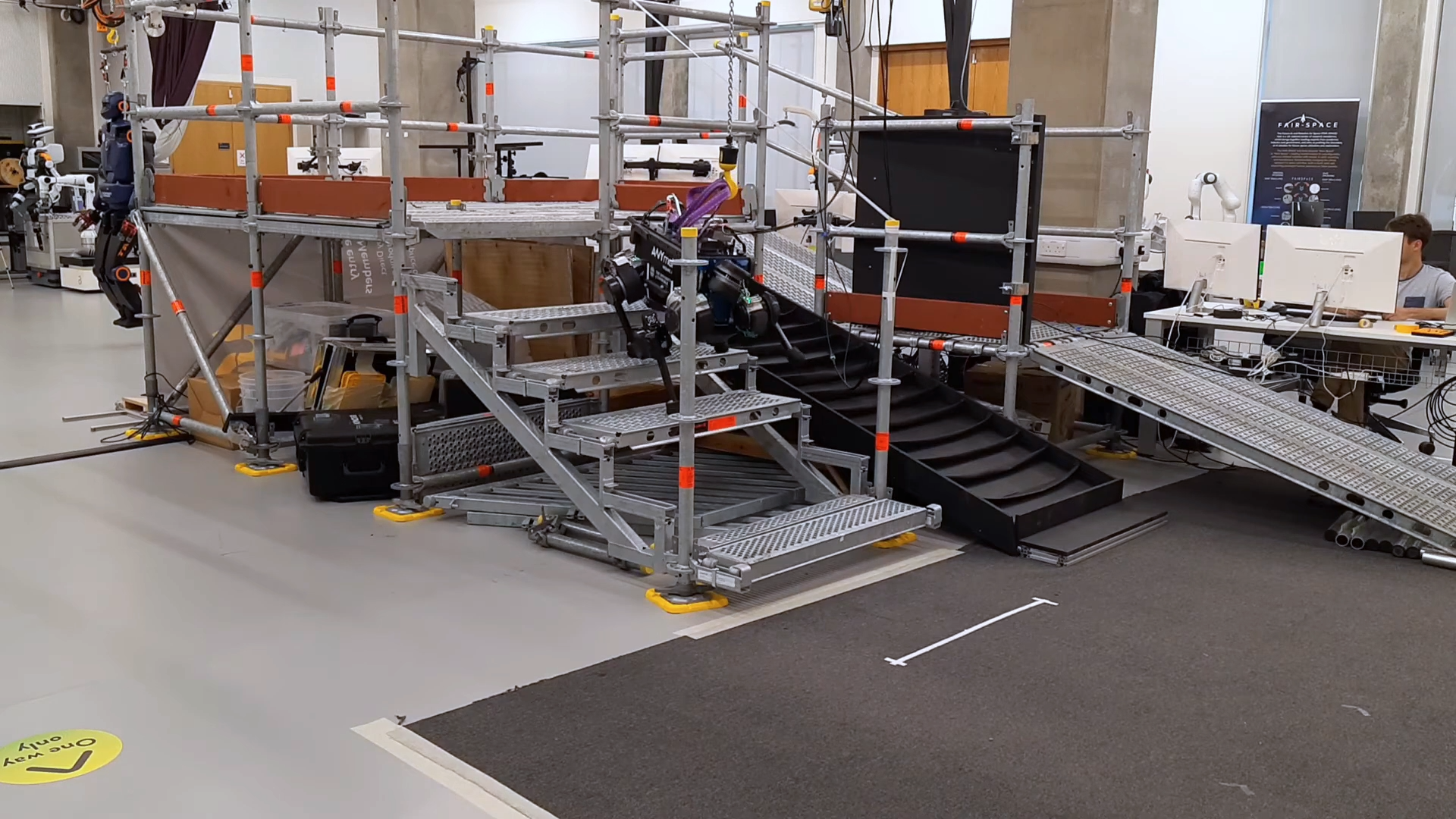} &
\subFigF{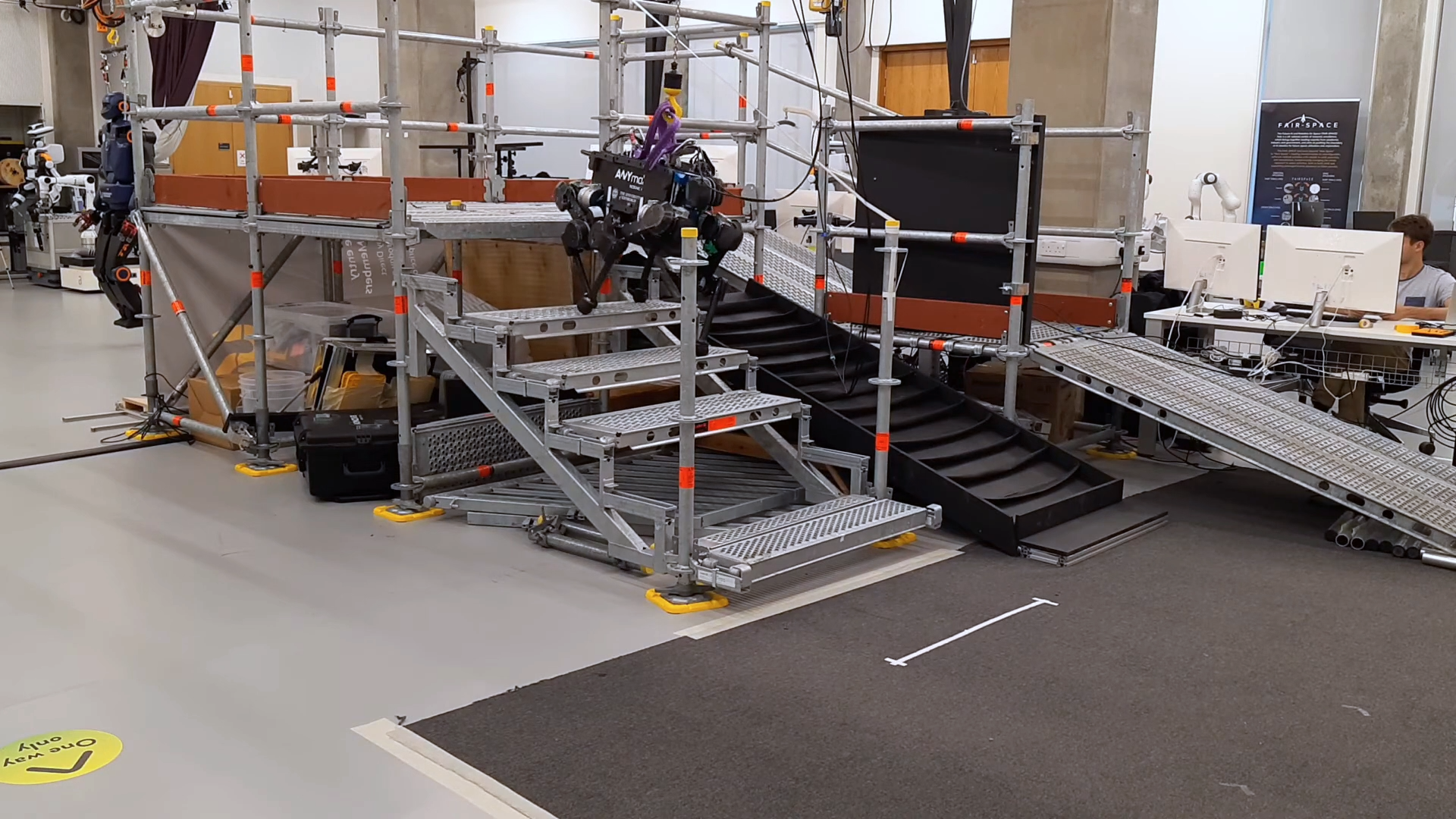} &
\subFigF{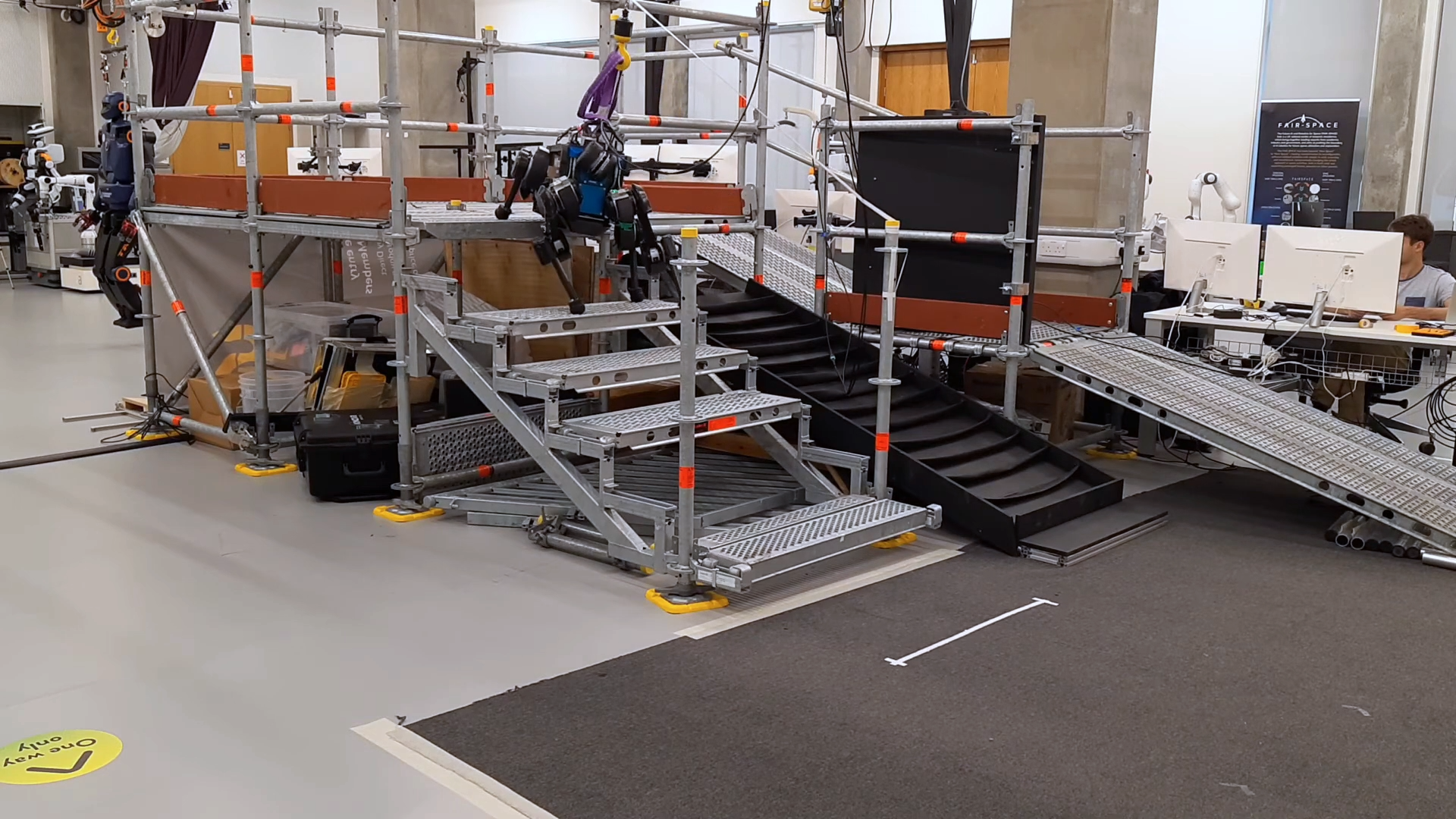} &
\subFigF{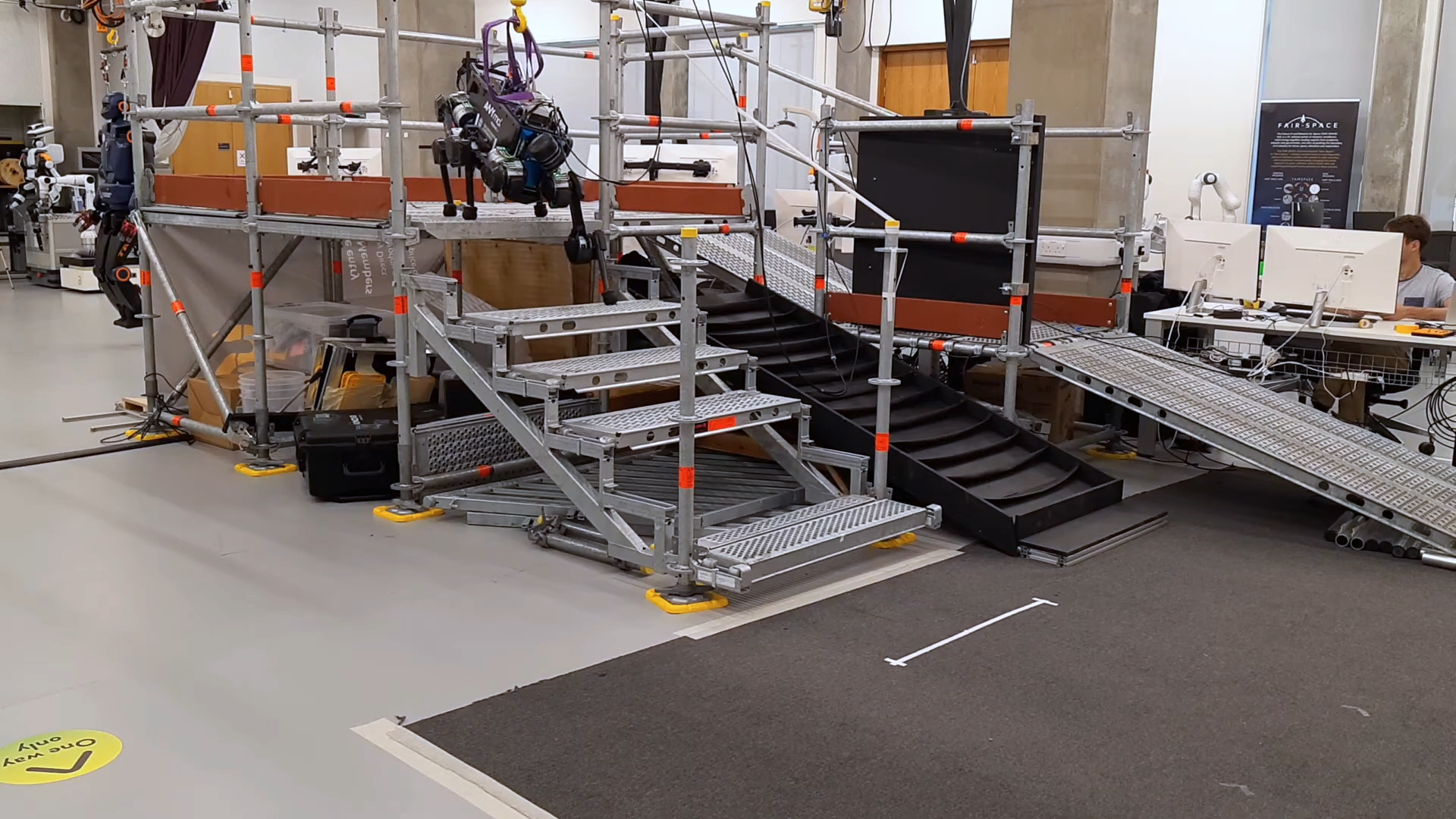}\\
\\[-2.5ex]
\cline{2-6}
\\[-2.2ex]
\rowname{g}&
\boxit{\videoLink{320}}{50pt}%
\subFigG{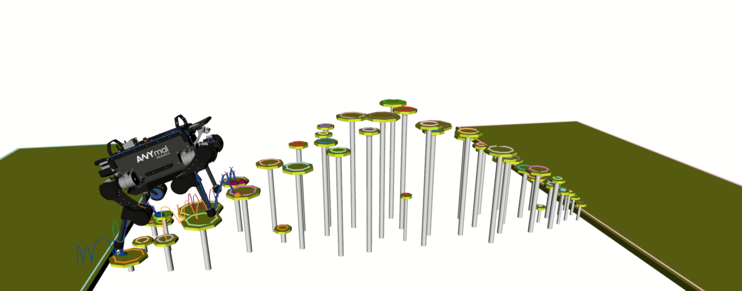}&
\subFigG{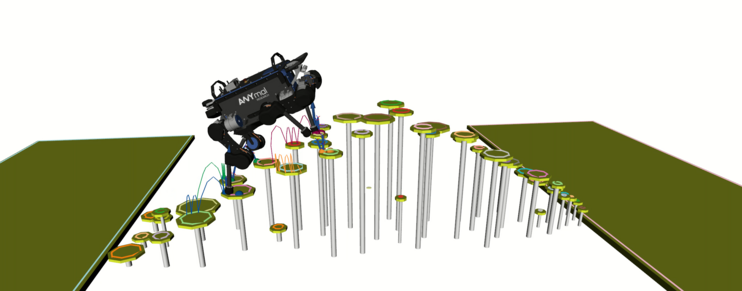}&
\subFigG{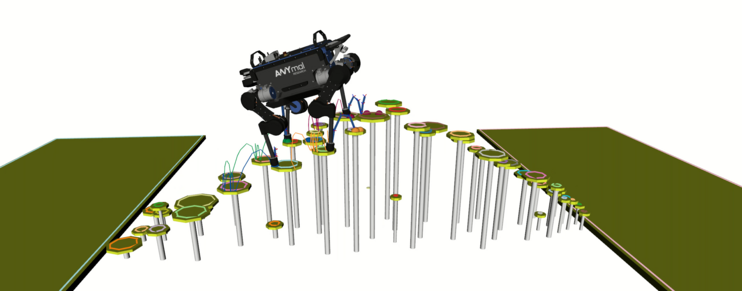}&
\subFigG{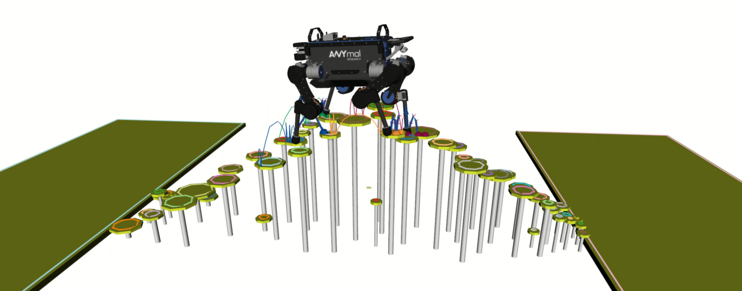}&
\subFigG{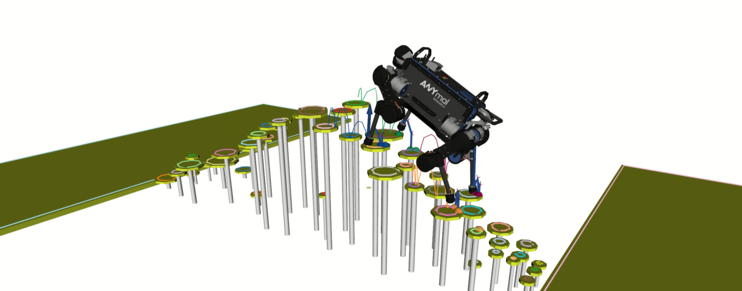}\\
\\[-2.5ex]
\cline{2-6}
\\[-2.2ex]
\rowname{h}&
\boxit{\videoLink{135}}{60pt}%
\subFigH{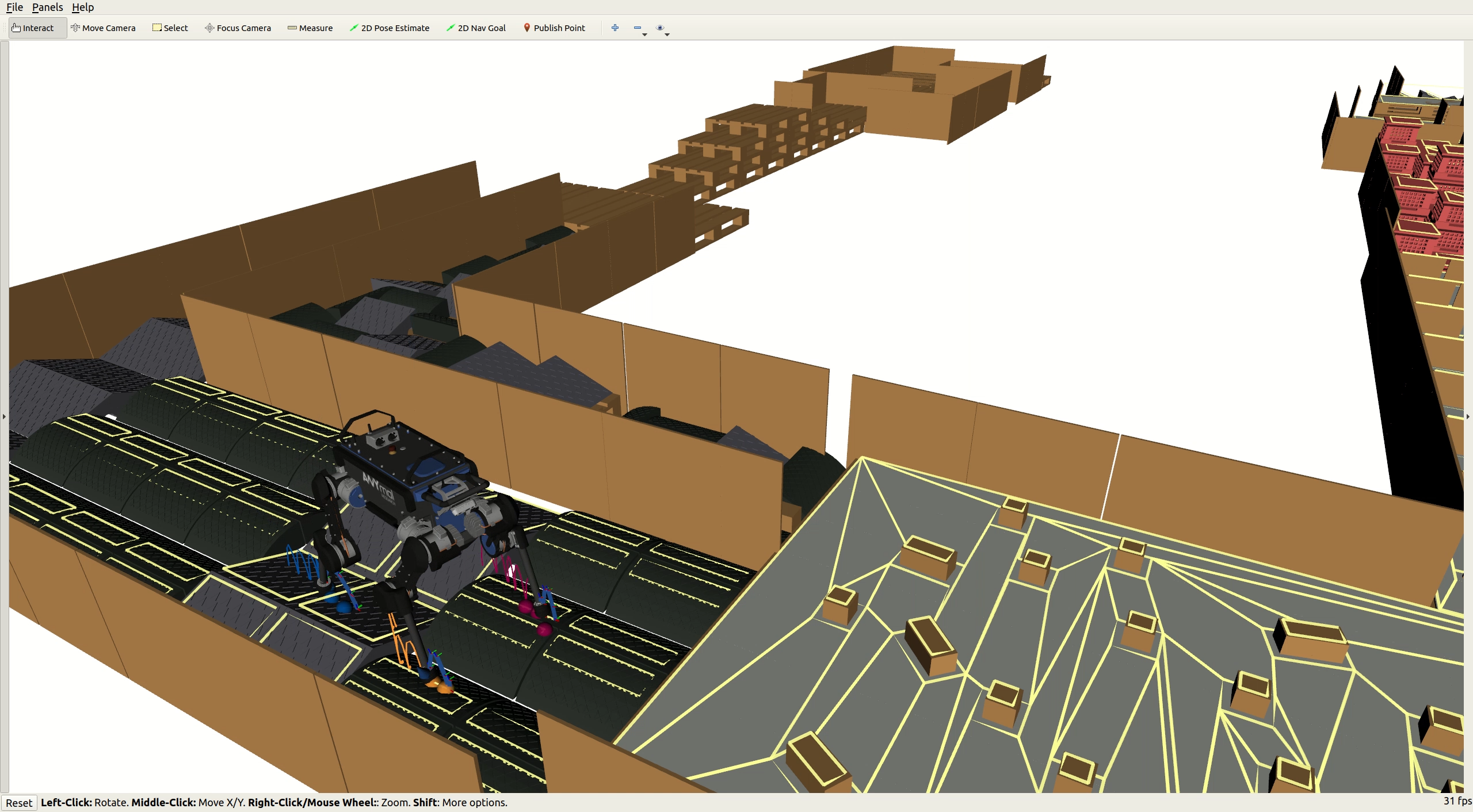} &
\subFigH{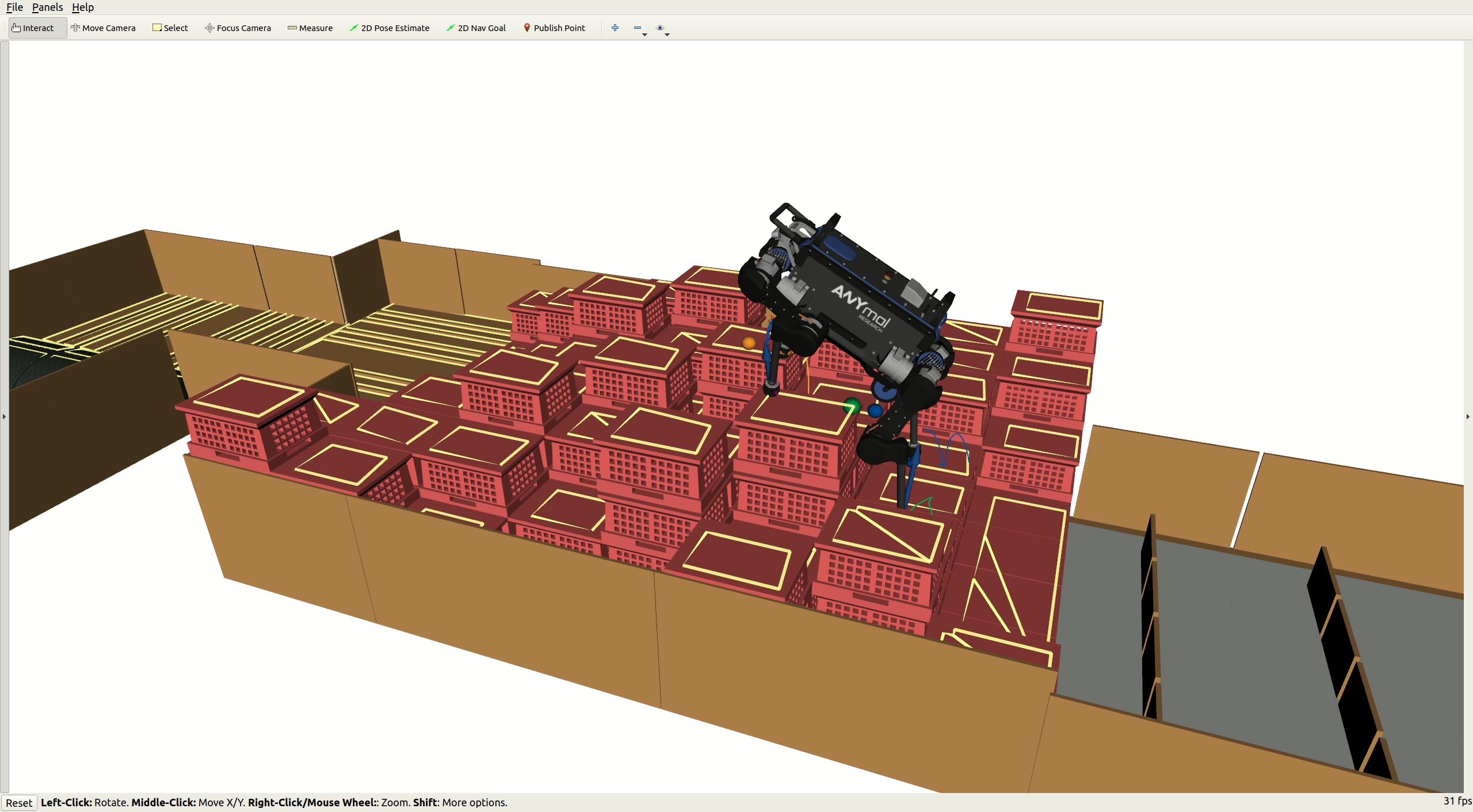} &
\subFigH{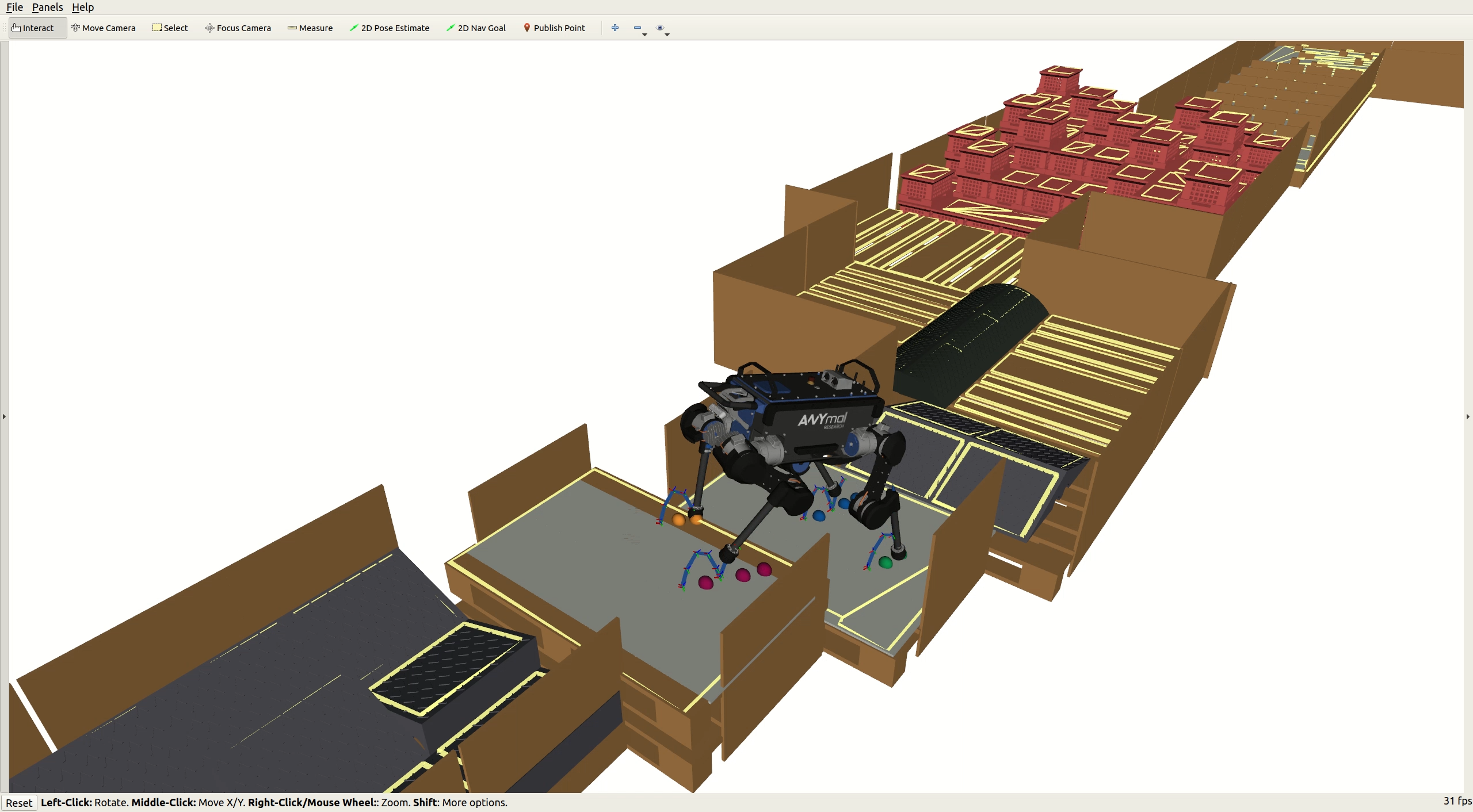} &
\subFigH{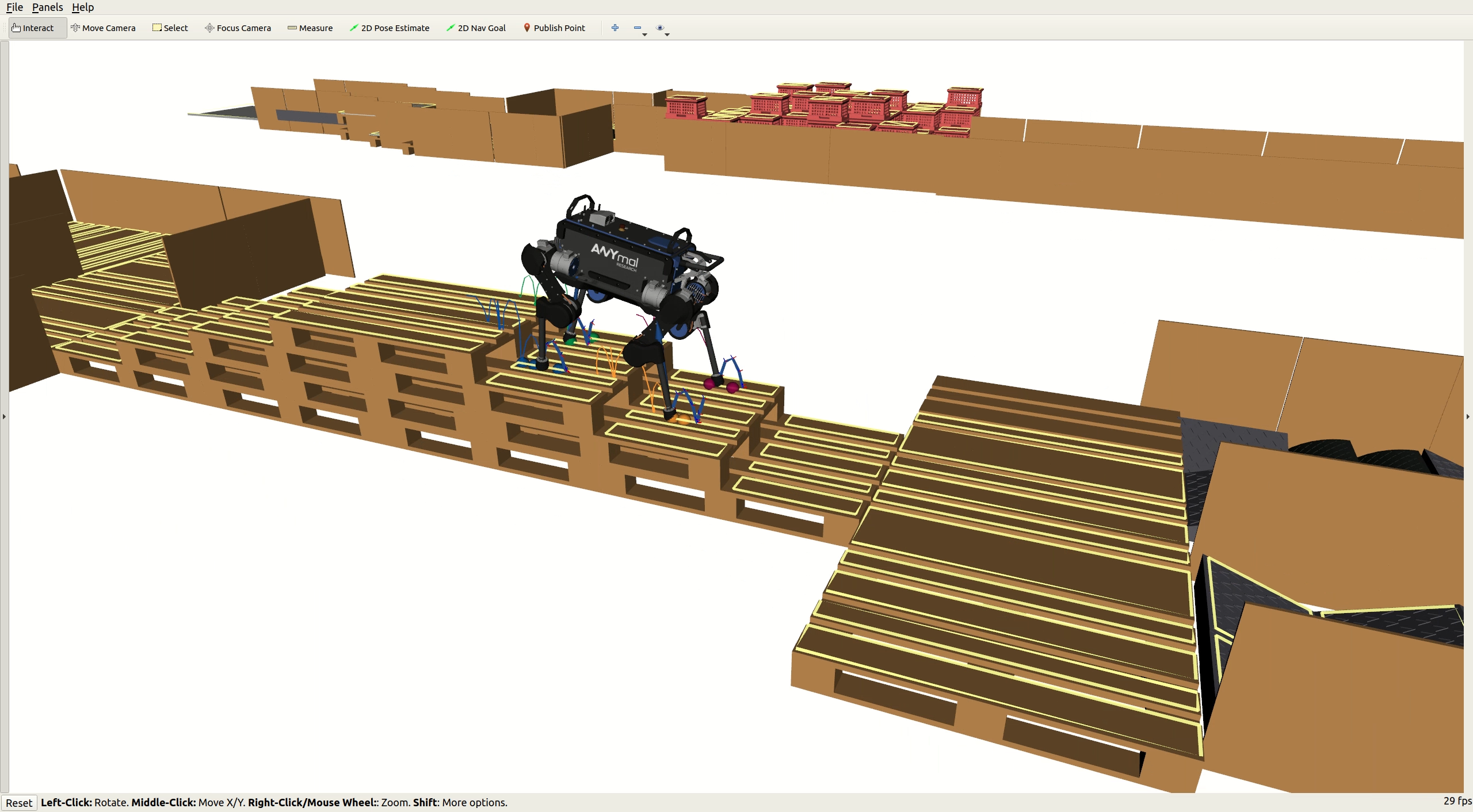} &
\subFigH{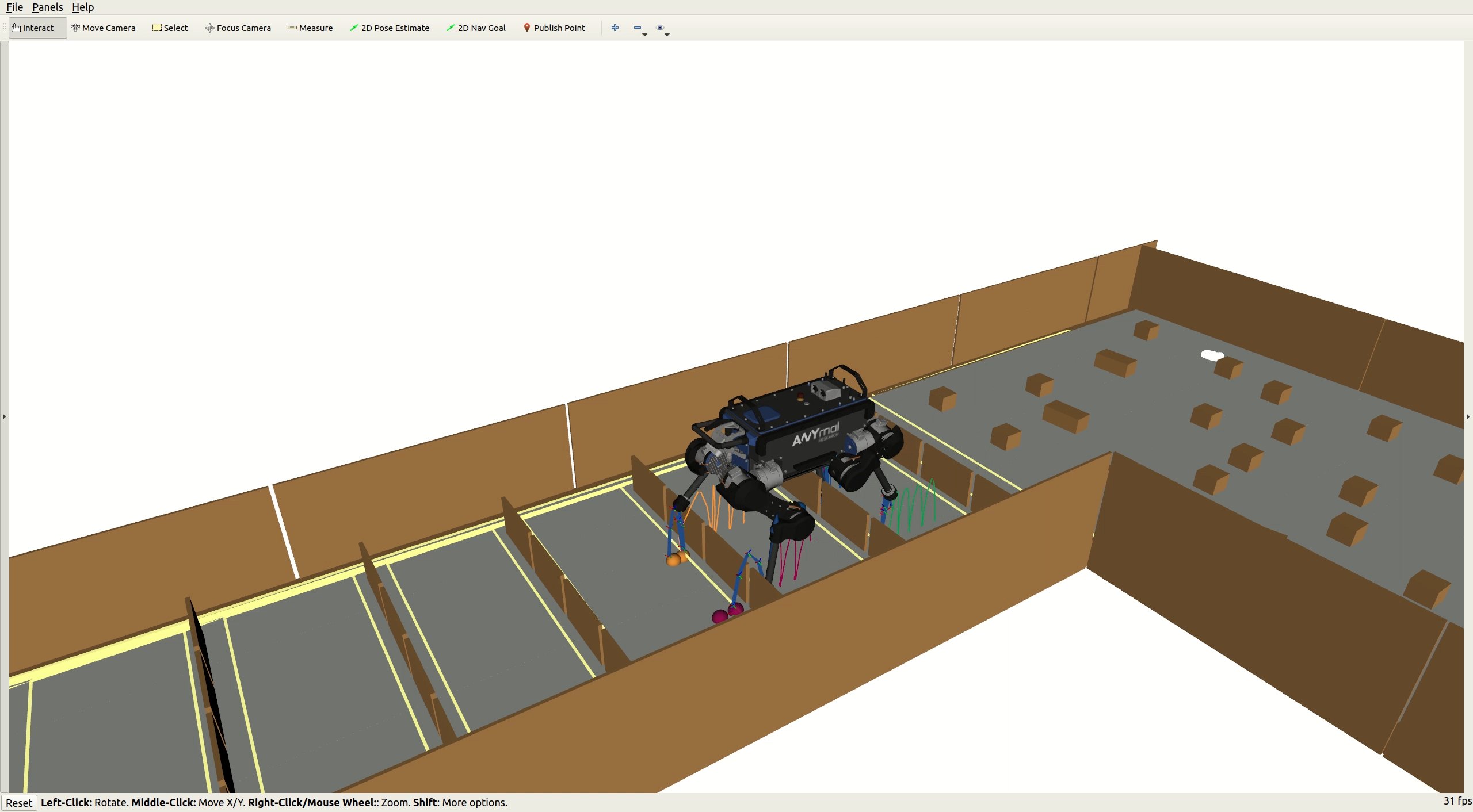}\\
\end{tabular*}
\caption{Screenshots of different experiments highlighting our architecture's capabilities. The first three rows of results were obtained using the onboard camera and a complete perception pipeline for active obstacle detection. Rows (d), (e), and (f) were performed without perception setup but instead used a pre-computed model of the environment to overcome the perception system limitations and test the controller's limits. The last two rows were obtained through simulation to further test our pipeline in challenging scenarios. Video accessible at \videoUrl.}%
\label{fig::screenshots}
\end{figure*}
\endgroup

\subsection{Implementation}
\label{sec::results::implementation}
Communication between each module is achieved  via a low-latency ROS communication layer (TCP, no-delay) \cite{ROS}. Three onboard computers (\texttt{Intel(R) Core(TM) i7-5600U CPU @ 2.6GHz}) share the main tasks of locomotion, perception and estimation. Point-cloud framework, heightmap generator, state estimation and Riccati-gain controller are performed onboard. We used an additional computer (\texttt{Intel(R) Core(TM) i5-8365U CPU @ 1.60GHz}) to extract the segmented planes from the heightmap, post-process the extracted surfaces and run the mixed-integer program in a different thread. A final computer (\texttt{Intel(R) Core(TM) i9-9900KF CPU @ 3.60GHz}) is used to run the MPC which sends the plan to the Riccati-gain lower controller at \SI{50}{\hertz}. 

\subsection{Experiments}
The pipeline has been tested in various scenarios, first with onboard perception and then with a model of the environment. This is to emphasize the motion generation part and break away from the perception constraints. During all experiments, the user commands the robot's velocity with a joystick.  
Some of the experiments (\textit{1.3}, \textit{1.4} and \textit{2.1}) were conducted using the inverse dynamic formulation and presented in the related paper \cite{Inv-Dyn-MPC-mastalli} as evidence of the formulation's effectiveness. We present these experiments again in this paper to specifically highlight the planning aspect. 
We release the environments tested for this paper in a public repository (\url{https://github.com/thomascbrs/walkgen-environments}). The source code for our planner is available as an open source package (\url{https://github.com/loco-3d/sl1m.git}) and we will release the rest of our code upon acceptance of the paper.

\makeatletter
\newcommand{\labelExpe}[2]{\hypertarget{#1}{\textit{#2}}\global\@namedef{labelExpe@#1}{\textit{#2}}}
\newcommand{\linktoExpe}[1]{%
\@ifundefined{labelExpe@#1}{\textbf{??}\@latex@warning{Reference `#1' on page \thepage \space undefined}}%
{\hyperlink{#1}{\@nameuse{labelExpe@#1}}}%
}
\makeatother

\subsubsection{Using onboard perception}
\label{itm::sec::perception}
We evaluated our complete pipeline on three major scenarios, listed below. 
\begin{itemize}
    \item \textit{Experiment}~\labelExpe{updown}{1.1} (Fig.~\ref{fig::intro_walk_down}): The first experiment is a 5-minute experiment, representative of the type of environment our perceptive locomotion pipeline enables. The robot starts at the bottom of an industrial staircase with 7 steps, each \SI{17}{\centi\meter} high and \SI{29}{\centi\meter} deep. The average terrain slope is 30 degrees. Once at the top, the robot makes a U-turn on the industrial platform  and performs its descent.
    \item \textit{Experiment}~\labelExpe{moving_obs}{1.2} (Fig.~\ref{fig::screenshots}-a): The second scenario corresponds to two platforms of 1 by 1 meters connected by a piece of wood placed diagonally to the right of the assembly. Once the robot is on the first platform, we manually remove the ground from the contact surface list to prevent the robot from moving forward. A 20x30\,cm block was then added to the left of the assembly, and after a few seconds, the platform is detected and the robot moved forward. This demonstrates the pipeline's reactive capabilities.
    \item \textit{Experiment}~\labelExpe{2wood}{1.3} 
    (Fig.~\ref{fig::screenshots}-b): It is the same configuration as the previous experiment with 2 pieces of wood connecting the two platforms. This experiment highlights the accurate execution of footstep plans.  
    \item \textit{Experiment}~\labelExpe{multi}{1.4} 
    (Fig.~\ref{fig::screenshots}-c): Using the robot onboard perception, the final experiment aims to mix various terrains types and heights. Since the metallic plate was not fixed, it slid at the end of the experiment, showing the stability of the controller.
\end{itemize}

\subsubsection{Using a model of the environment}
To further evaluate the locomotion capabilities of our pipeline, we conducted additional experiments in which the contact surfaces are known a priori, and not given by Plane-Seg. The robot's pose with respect to the world frame is still estimated using the onboard sensors (LIDAR and proprioceptive sensors) and in general, the rest of the framework remains the same.

\begin{itemize}
    \item \textit{Experiment}~\labelExpe{up-missing}{2.1} 
    (Fig.~\ref{fig::screenshots}-d) : The first experiment was to climb stairs with 2 missing steps. We removed steps 2 and 6 of the stairs. It corresponds to climbing a slope of 30 degrees with two gaps of 34 cm. This pushes the robot to its kinematic and the actuation limits. This, therefore, highlights the benefit of taking into account the entire robot model to plan the motion and adapt the posture. This justifies our approach, as discussed in Sec.~\ref{sec::results::whole-blody}.    
    \item \textit{Experiment}~\labelExpe{down-missing}{2.2} 
    (Fig.~\ref{fig::screenshots}-e): The second experiment is similar and corresponds to descenting stairs with 1 missing step (step 6). It corresponds to crossing a slope of 30 degrees with a gap of \SI{34}{\centi\meter}. 
    \item \textit{Experiment}~\labelExpe{step40cm}{2.3}
    (Fig.~\ref{fig::screenshots}-f) : The last experiment was conducted on a platform of size  \SI{1}{\meter} $\times$ \SI{1}{\meter} and \SI{38}{\centi\meter} in height. It is higher than the robot's height in its nominal position. It is the only experiment where we had to increase the safety margin around the obstacle (outer margin in Sec.~\ref{sec::perception}) to \SI{12}{\centi\meter} since the robot's shoulders are at the same height as the obstacle and collide with it. 
\end{itemize}

\subsubsection{Simulation experiments}
A final set of experiments were run in the \textsc{PyBullet} simulator \cite{coumans2021} to show more dynamic gaits such as trotting. This was done to illustrate the diversity of terrains the framework can target. The setup and communication protocol are identical to the one used on the hardware and described previously in Sec.~\ref{sec::results::implementation}. 

\begin{itemize}
    \item \textit{Experiment}~\labelExpe{pylons}{3.1}
    (Fig.~\ref{fig::screenshots}-g): The first experiment corresponds to a set of stepping stones of different sizes and heights crossed with a trotting gait.
    \item \textit{Experiment}~\labelExpe{icra}{3.2}
    (Fig.~\ref{fig::screenshots}-h): The second experiment involves the ICRA 2023 Quadruped Challenge, which comprises a parkour task featuring a range of obstacles, including pallets, inclined ramps, rounded rubber ramps, and a stack of plastic crates. This challenge serves as a valuable benchmark for evaluating our pipeline capabilities. For most parts, our pipeline navigates parkour using a dynamic trot. For some of them, we had to marginally modify some parameters. The wooden pallets with gaps between the wooden slats imply many potential surfaces for each contact, up to 12, even in the pre-selection section. Hence, we had to reduce the surface planner's horizon to optimise the next 4 contact sequences (and not 6 as used previously). Step height need to be increased by up to \SI{25}{\centi\meter} on flat terrain with wood panels of \SI{20}{\centi\meter}. It would be ideal to include height map directly in trajectory optimization. This is rather than solely relying on surfaces for obstacle avoidance. as discussed previously in Sec.~\ref{sec:foot_trajectory_generation}. Furthermore, when dealing with a stack of plastic crates, adapting the gait to a walking gait (in which only one foot moves at a time) is more stable and enables safe passage across the terrain, compared to a more dynamic trotting gait. 

\end{itemize}

\subsection{Evaluation of the perception pipeline}
 
\begingroup
\newcommand{\subFigg}[2]{\subfloat[]{\includegraphics[clip,trim=150px 35px 100px 100px,width={{#1}}]{#2}}}
\newcommand{\subFigB}[2]{\subfloat[]{\includegraphics[clip,trim=0px 0px 0px 0px,width={{#1}}]{#2}}}
\newcommand{\widthFig}{0.98\linewidth}

\setlength{\belowcaptionskip}{0.\baselineskip}
\begin{figure}[ht]
    \captionsetup[subfigure]{labelformat=empty}
     \centering
     \subFigg{\widthFig}{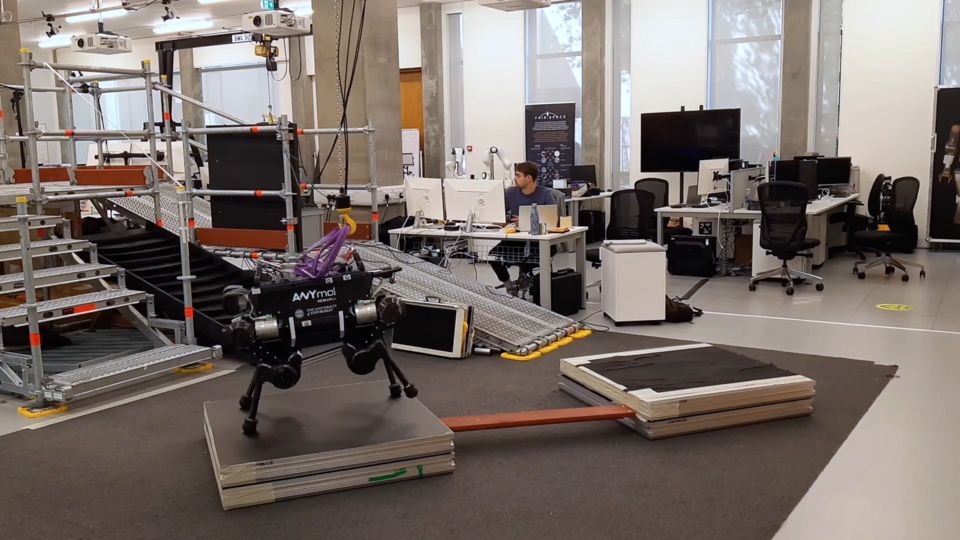}  
     
    \vspace{-4.2ex}
     \subFigB{\widthFig}{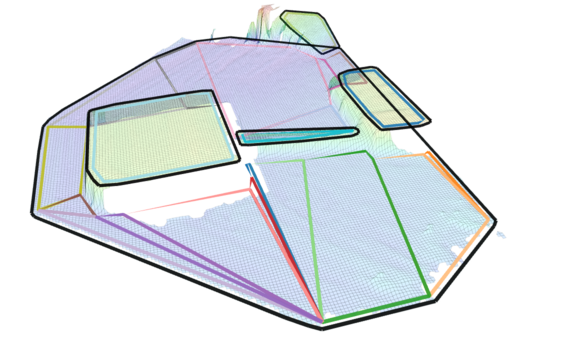}
     
    \caption{Perception output during experiment 1-(ii) before ground removal. Black surfaces: initial convex surfaces; coloured surfaces: filtered.}
    \label{fig::heightmap_post_process}
\end{figure}
\endgroup

\begingroup
\newcommand{\subFigg}[2]{\subfloat[]{\includegraphics[draft=false,clip,trim=100px 23px 66px 0px,width={{#1}}]{#2}}}
\newcommand{\subFigB}[2]{\subfloat[]{\includegraphics[draft=false,clip,trim=100px 33px 100px 33px,width={{#1}}]{#2}}}
\newcommand{\widthFig}{0.49\linewidth}
\def\boxit#1#2{%
  \smash{\llap{\rlap{\hspace{4pt}\href{#1}{\strut\raisebox{\height}{\phantom{\rule{0.9\linewidth}{#2}}} } }~}}\ignorespaces}
\hypersetup{pdfborder=0 0 0}

\setlength{\belowcaptionskip}{0.\baselineskip}
\begin{figure}[ht]
    \captionsetup[subfigure]{labelformat=empty}
     \centering
     \boxit{\videoLink{210}}{35pt}%
     \subFigg{\widthFig}{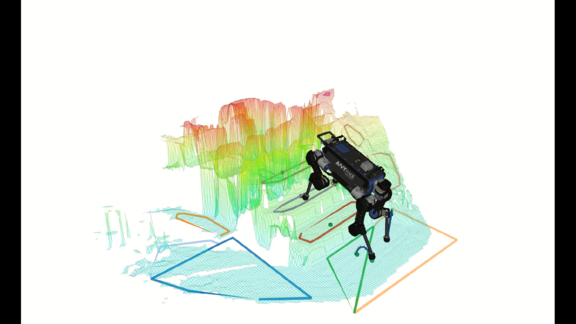}
     \subFigg{\widthFig}{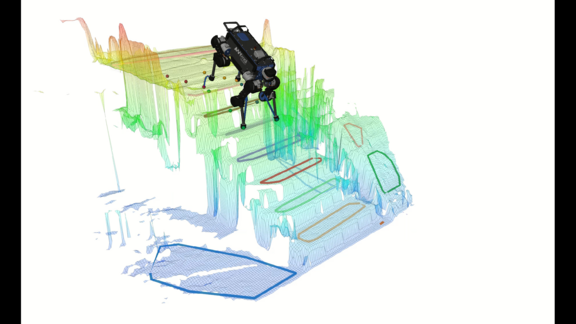}
     
    \vspace{-4.2ex}
     \boxit{\videoLink{210}}{35pt}%
     \subFigB{\widthFig}{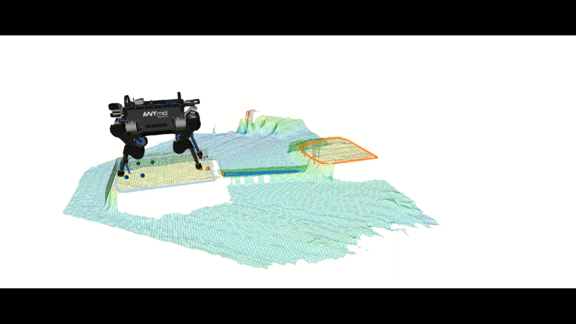}
     \subFigB{\widthFig}{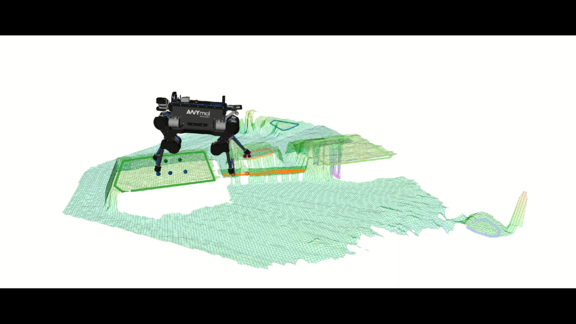}
    \caption{Height map and resulting surfaces during experiment 1-(i) (first row) and experiment 1-(ii) (second row). (\videoUrl).}
    \label{fig::heightmap_visu}
\end{figure}
\endgroup

In this section, we evaluate the perception pipeline, from the heightmap to the resulting surfaces filtered and reshaped for security margins. Fig.~\ref{fig::heightmap_post_process} shows the scene during experiment \linktoExpe{updown}, before the manual removal of the ground. The necessity of the filtering routine, described in Sec.~\ref{sec::perception} is highlighted in this example since the convex planes extracted from the height map overlap. Additionally, we note that the security margin is applied inside the obstacles to avoid walking on the edges. This is due in part to estimation errors, which are around evaluated to 3-4 cm. More height maps and resulting potential surfaces are shown in Fig \ref{fig::heightmap_visu}, taken from experiments \linktoExpe{updown} and \linktoExpe{moving_obs}. It is interesting to note the evolution of the terrain estimation around the robot during these experiments. When climbing or descending stairs, the robot's camera only identifies the next 1 or 2 stairs ahead. During the second experiment \linktoExpe{moving_obs}, where an obstacle was added in front of the robot, it took several iterations of the probabilistic algorithm to update the heightmap with the detected object and consequently 2 to 3 seconds to obtain a feasible surface to walk on.

\subsection{Computation performance}

Table~\ref{tab::info-surface-planning} presents the computation time statistics for each module of the pipeline during experiment \linktoExpe{updown}, i.e., climbing up and down stairs with active perception.

\begingroup
\renewcommand{\arraystretch}{1.5} 
\begin{table}[ht]
\centering
\caption{Computing time}
\label{tab::info-surface-planning}
\begin{tabular}{l c c c}
    \multicolumn{1}{c}{\textbf{Name}} & \textbf{Mean} & \textbf{Min} & \textbf{Max}  \\
    \hline\hline
    \quad \quad \textbf{\textit{Surface Processing}} \\
    \hline
    Number of surfaces processed & \textbf{23.71} & 8. & 45. \\ 
    Surface processing [ms] & \textbf{88.89} & 19.49 & 164.22 \\    
    \hline
    \quad \quad \textbf{\textit{Surface Selection}} \\
    \hline
    Number of potential surfaces & \textbf{3.08} & 1. & 9. \\ 
    Pre-selection [ms] & \textbf{17.16} & 5.47 & 47.46 \\
    MIP [ms] & \textbf{98.62} & 32.95 & 253.15 \\ 
    \hline
    \quad \quad \textbf{\textit{Foot Trajectory}} \\
    \hline
    Foot location [ms] & \textbf{0.14} & 0.08 & 0.74 \\ 
    Foot trajectory [ms] & \textbf{0.082} & 0.007 & 0.33 \\
    \hline
    \quad \quad \textbf{\textit{Motion Generation (MPC)}} \\
    \hline
    Solve time [ms] & \textbf{9.29} & 8.02 & 13.52 \\
    Total time [ms] & \textbf{13.27} & 8.76 & 18.50 
\end{tabular}
\end{table}
\endgroup

\subsubsection{Surface Processing}
It takes \SI{90}{\milli\second} on average to post-process the incoming surfaces from perception with Algorithm~\ref{alg::surface_decompo} presented in Sec.~\ref{sec::perception}. In comparison of the plane's update frequency, which is between \SI{0.5}{\hertz} and \SI{1}{\hertz} with an non-optimised code, it represents roughly an increase of 5-10\%.

\subsubsection{Surface Selection}
Surface selection, as described in Sec.~\ref{sec::surface_selection} encompasses the pre-selection step to reduce the number of potential surfaces considered by the mixed-integer program. It takes around \SI{120}{\milli\second} to find the next surface of contact in this experiment. Pre-selection reduces the potential surfaces from 23 to 3 for each foot. During this experiment, we optimise over 8 contact phases, which correspond to 8-foot positions optimised with a walking gait (1 contact is created/broken for each contact phase). The mixed-integer optimisation starts at the beginning of each upcoming foot trajectory and the next surface information needs to be received before the beginning of the upcoming cycle. In this experiment, the foot trajectory is \SI{600}{\milli\second}, which is enough for the maximum time taken. However, for a trotting gait, the foot trajectory is set to \SI{300}{\milli\second}. We must reduce the number of foot locations to 6, thus allowing 3 contact phases (2-foot locations optimised for each contact phase) to ensure a safe margin regarding the computing time. We note that the computing time of mixed-integer optimisation depends heavily on the number of potential surfaces for each contact and the number of contact locations optimised. 
For a detailed analysis of the MIP computation times, we refer the reader to \cite{tonneau:sl1m:9454381}.

\subsubsection{Foot trajectory}
The foot trajectory encompasses the optimisation of the footstep inside the attributed surface and the end-effector curve optimisation as described in Sec.~\ref{sec:foot_trajectory_generation}. It represents only 1\% of every MPC step.

\subsubsection{Motion generation (MPC)}
On average the MPC step takes around \SI{13}{\milli\second}, including the update of the optimal control problem and solving this latter with 1 iteration which takes around \SI{9}{\milli\second}. This is below the maximum \SI{20}{\milli\second} time required for an MPC running at \SI{50}{\hertz}.

\subsection{Analysis of the whole-body MPC}
\label{sec::results::whole-blody}
\begingroup
\newcommand{\subFigg}[2]{\subfloat[]{\includegraphics[clip,trim=50px 0px 33px 16px,width={{#1}}]{#2}}}
\newcommand{\widthFig}{0.14\linewidth}
\def\boxit#1#2{%
  \smash{\llap{\rlap{\hspace{4pt}\href{#1}{\strut\raisebox{\height}{\phantom{\rule{0.98\textwidth}{#2}}} } }~}}\ignorespaces}
\hypersetup{pdfborder=0 0 0}

\setlength{\belowcaptionskip}{0.\baselineskip}
\begin{figure*}
    \captionsetup[subfigure]{labelformat=empty}
     \centering
     \boxit{\videoLink{145}}{42pt}%
     \subFigg{\widthFig}{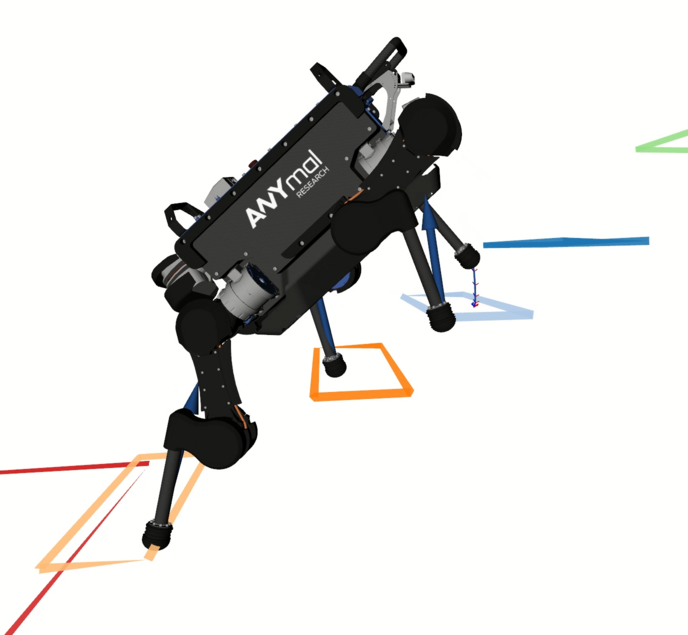}
     \subFigg{\widthFig}{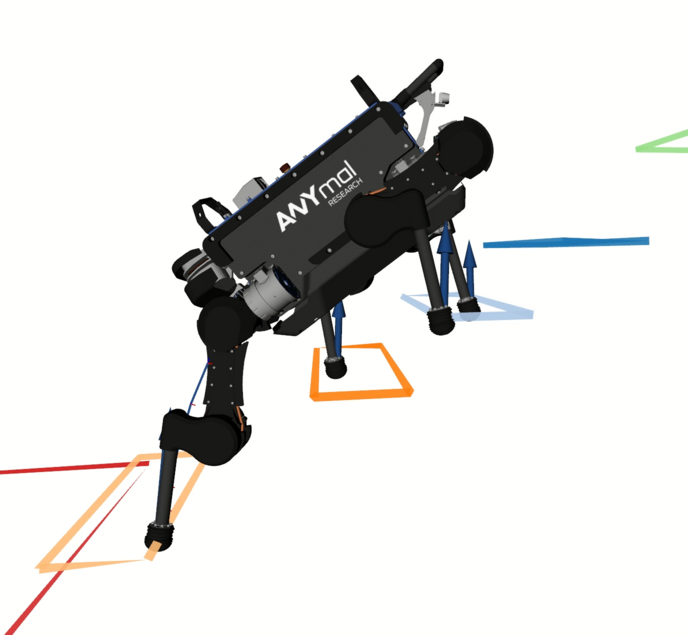}
    \vspace{-25px}%
     \subFigg{\widthFig}{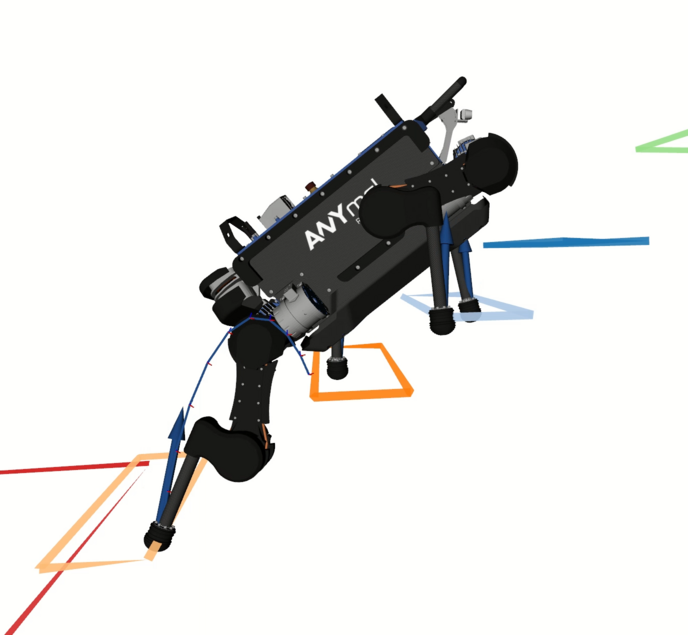}
     \subFigg{\widthFig}{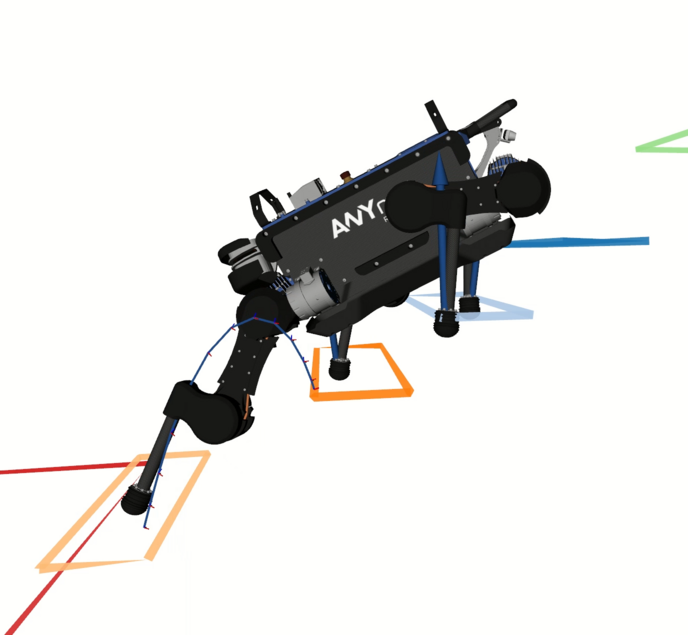}
    \vspace{-25px}%
     \subFigg{\widthFig}{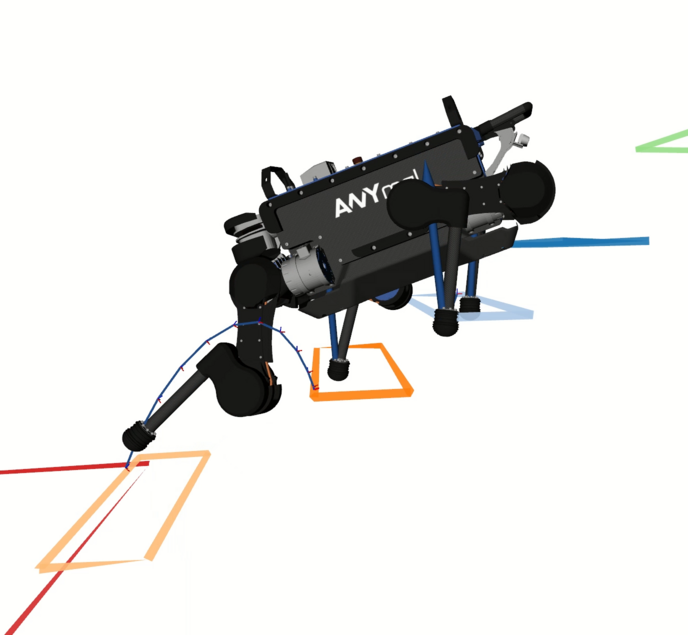}
     \subFigg{\widthFig}{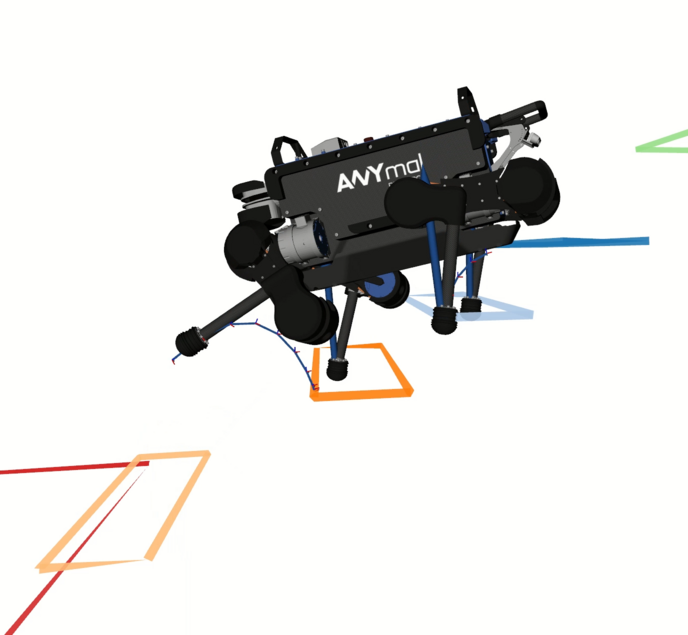}
     \subFigg{\widthFig}{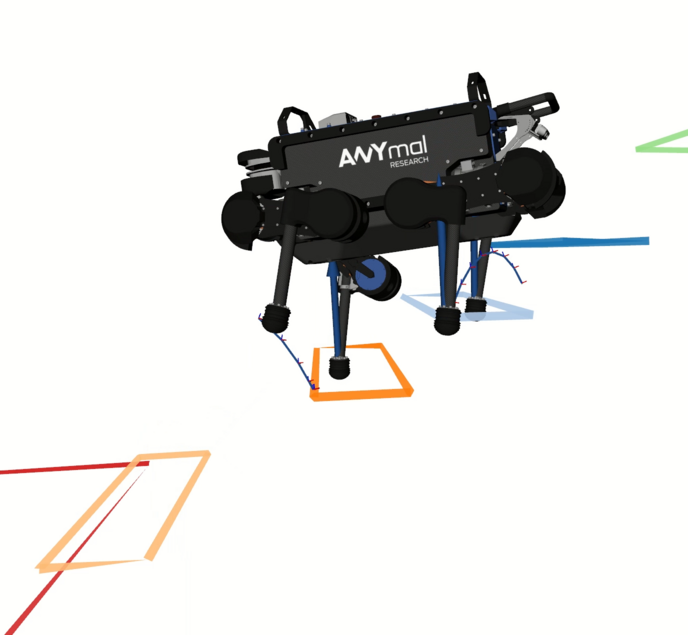}
     \vspace{35px}%
    
    \caption{Posture adjustment (experiment \protect\linktoExpe{up-missing}). The hind right leg reaches the torque limits on both HFE (hip flexion/extension) and KFE (knee flexion/extension) when crossing the gap. This corresponds to the peak torque at 52s. The whole-body MPC adjusts the body posture to compensate. The body leans forward and lowers as much as possible, reaching the kinematic limit of the hind leg to reduce torque on it. (\videoUrl).}
    \label{fig::body-posture-results}
\end{figure*}
\endgroup

In scenarios involving climbing steps, the torque limit is reached on at least one actuator in each of these experiments. It is in this case that the whole-body MPC becomes crucial as it adjusts body posture to reduce joint torques. Experiment \linktoExpe{up-missing} is a representative case. First, we observe that the COM motion leans forward and close to the feet, almost in contact with the stairs, at the moment of giving the last motion to cross the gap (Fig.~\ref{fig::body-posture-results}). The torque limit is reached on both joints of the hind right leg during this motion, as shown in Fig.~\ref{fig::torques-results}. We analyse the following quantities: torques, angular position and angular velocities at the HFE joint (hip flexion/extension) and the KFE joint (knee flexion/extension). The HAA joints (hip abduction/abduction) are less prone to reach torque limits. While crossing the gap, we can observe two torque peaks that correspond exactly to the robot configuration shown in Fig.~\ref{fig::torques-results}. In this instance, the overshoot of the torque command above the joint limits (a few Nm) can be attributed to two main reasons. First, the constraints on torque limits can be violated (Sec.~\ref{sec::motion_generation}) as the Riccati controller guarantees joint limits within a neighbourhood.

\begin{figure}[hbt!]
     \centering
     \subfloat[]{%
      \includegraphics[width=0.4\linewidth, keepaspectratio = true]{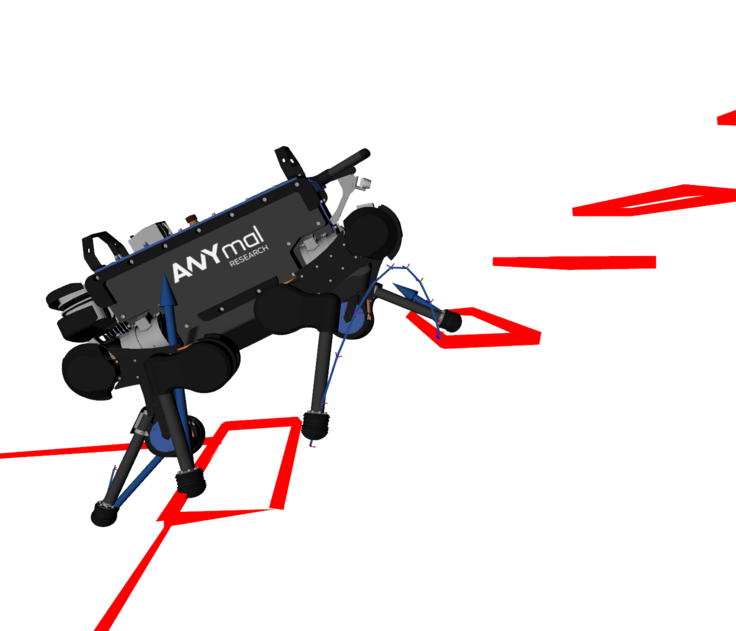}
      \label{fig::results::phase1}}
    \subfloat[]{%
      \includegraphics[width=0.4\linewidth, keepaspectratio = true]{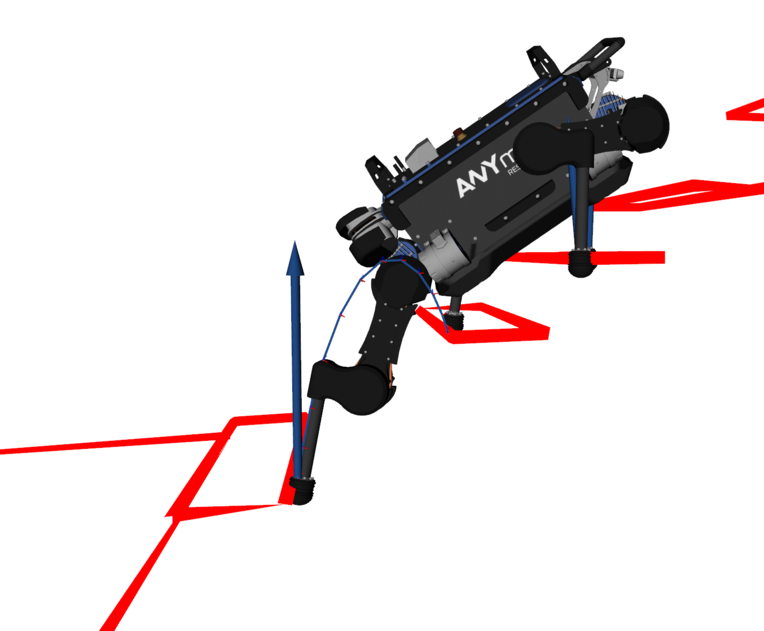}
      \label{fig::results::phase2}}

    \subfloat[]{%
      \includegraphics[width=0.9\linewidth,  keepaspectratio = true]{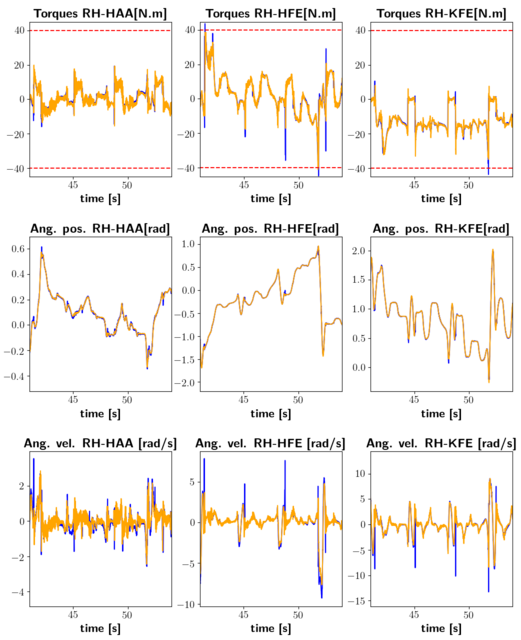}
      \label{fig::results::torques}}
    
    \caption{Torque, angular position and velocities of the joints HAA (hip abduction/abduction), HFE (hip flexion/extension) and KFE (knee flexion/extension) of the hind right leg during experiment \protect\linktoExpe{up-missing}. Blue quantities are computed by the state feedback controller and orange ones are measured on the robot. The maximum torque limit is approximately \SI{40}{\newton\meter} and is represented by red dashed curves. Each actuator can reach 12 $\textrm{m.s}^{\textrm{-1}}$. Two torque peaks reaching the boundaries can be observed around the \SI{40}{\second} and \SI{52}{\second} which correspond to the robot configurations shown above.}
\label{fig::torques-results}
\end{figure}

\subsection{Evaluation of the collision-free foot trajectory}


\begin{figure}[ht]
     \centering
    \subfloat[]{%
      \includegraphics[width=0.48\linewidth, keepaspectratio=false]{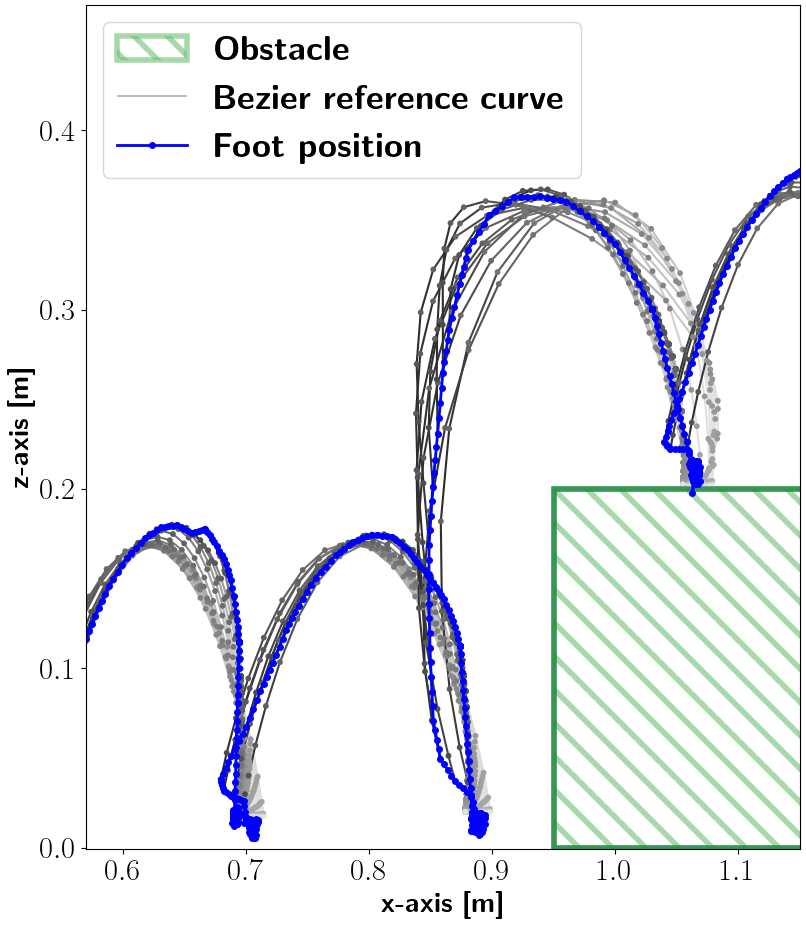}
      \label{fig::results-trajectory-obstacle}}
    \subfloat[]{%
      \includegraphics[width=0.48\linewidth]{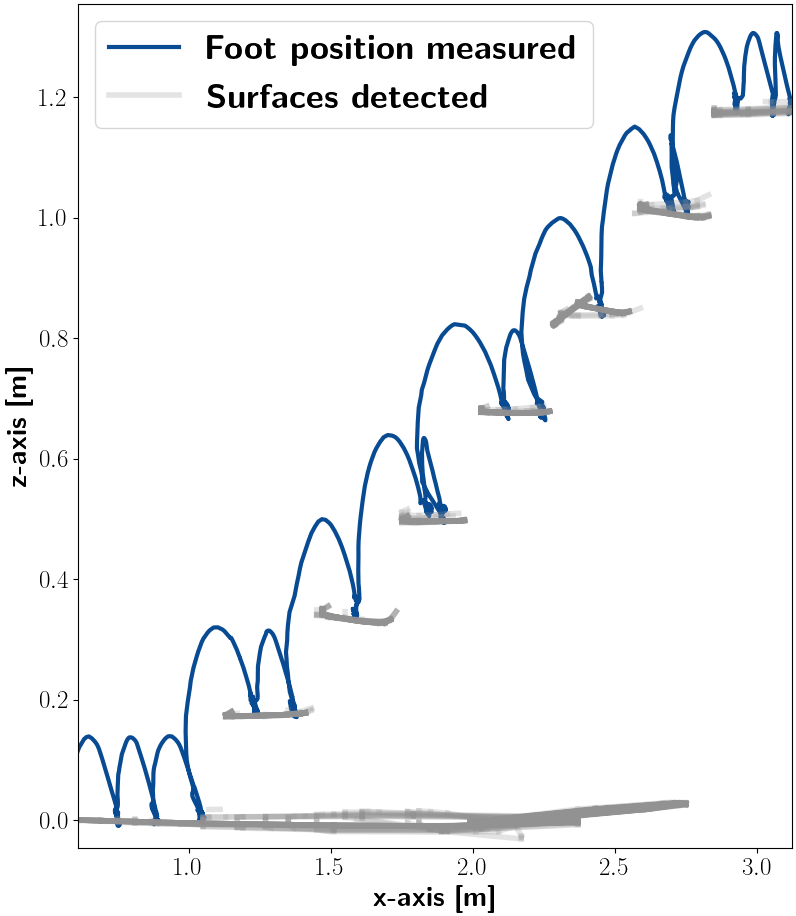}
      \label{fig::results-trajectory-stairs}}

    \caption{Side view of the front-left foot's trajectory while crossing a \SI{20}{\centi\meter} obstacle (left) and the staircase (right) using a walking gait pattern at a reference velocity of 0.1 $\textrm{m.s}^{-1}$. The grey curves on the left graph are reference Bezier curves computed during each control cycle. Their transparency indicates the prediction horizon. More transparent curves represent further predictions for the future. Grey areas on the right correspond to the cumulative position of surfaces detected by the camera.}
    \label{fig:results-trajectory}
\end{figure}

The end-effector trajectory is re-computed at \SI{50}{\hertz} in order to get robust tracking. Some minor modifications were necessary when transitioning from simulation to hardware. An offset of -\SI{1}{\centi\meter} has been added on the z-axis to accommodate for perception errors, reaching 2/3 cm at the end of our longest experiments (\linktoExpe{updown}), and ensuring contact creation occurs. A finer control of the feet's trajectory according to the contact detection was not necessary in our case, as the whole-body MPC is robust to state estimation errors (Sec. \ref{sec::results-robustness}). Additionally, the swing-foot trajectory and footstep are no longer updated once 70\% of the flight phase has passed in order to avoid a sliding contact after a sudden change in the target position. 
Finally, the end-effector velocity feedback is not taken into account in the optimisation due to large uncertainties about the end-effector velocities. The foot velocity is therefore assumed to be tracked properly and the initial velocity while re-computing the curve is taken from the previous control cycle.

Fig.~\ref{fig:results-trajectory} shows a swing-foot trajectory while the robot crosses an obstacle of \SI{20}{\centi\meter} and while climbs the stairs during experiment \linktoExpe{updown}. We observe the state estimation errors on the foot position when landing on the ground, resulting in the foot slightly bouncing.

\subsection{Evaluation of the velocity tracking and design choices}
\label{sec::base_position_ref_vel}

\begingroup
\newcommand{\widthFig}{0.48\linewidth}
\setlength{\tabcolsep}{0.25pt} 
\renewcommand{\arraystretch}{1.5} 
\setlength\arrayrulewidth{1.pt}
\newcommand{\subFig}[1]{\raisebox{-.5\height}{\includegraphics[clip,trim={100px 80px 90px 110px},width=\widthFig]{{#1}}}}

\begin{figure}[hbt!]
\centering
\begin{tabular*}{\linewidth}{c c }
\subFig{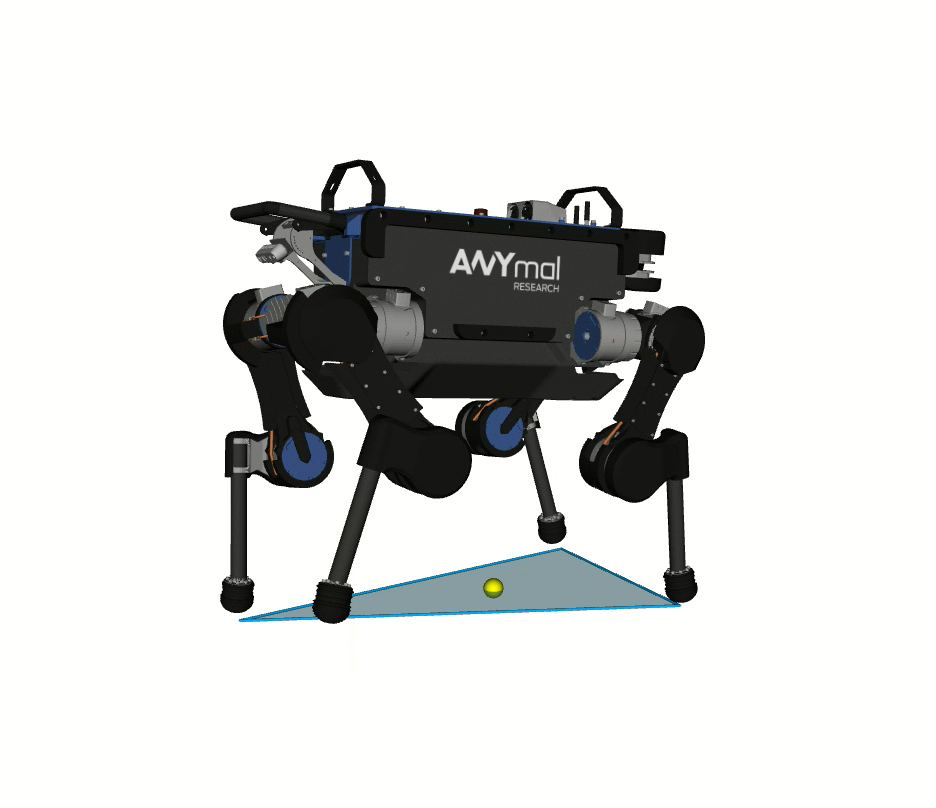}&
\subFig{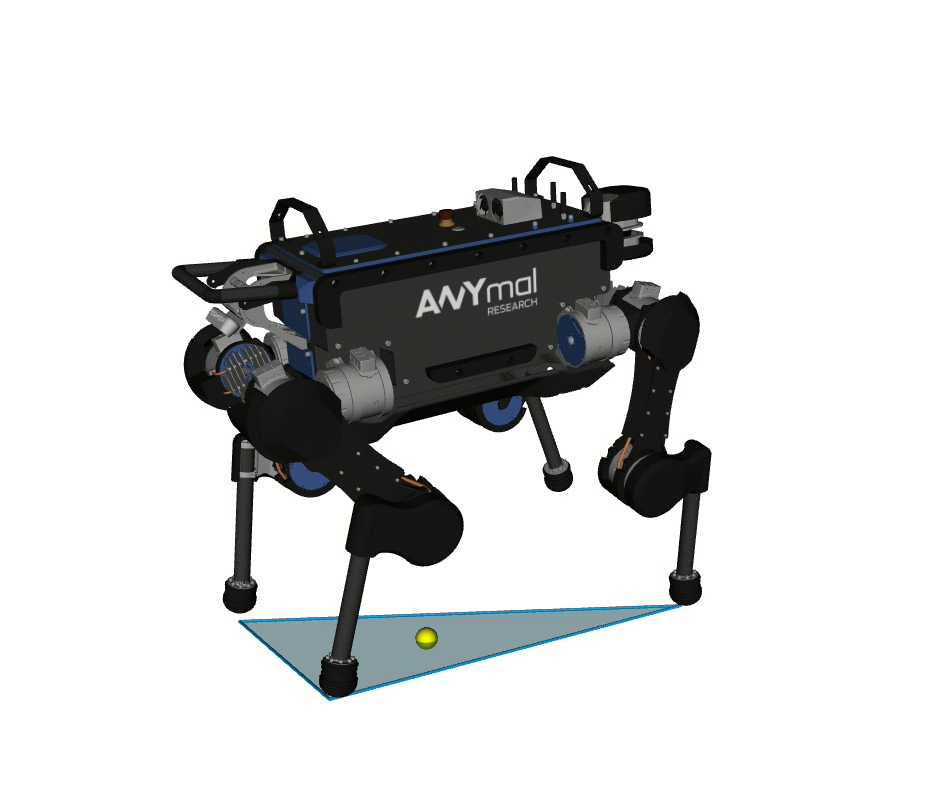}\\
\\[-3.6ex]
\subFig{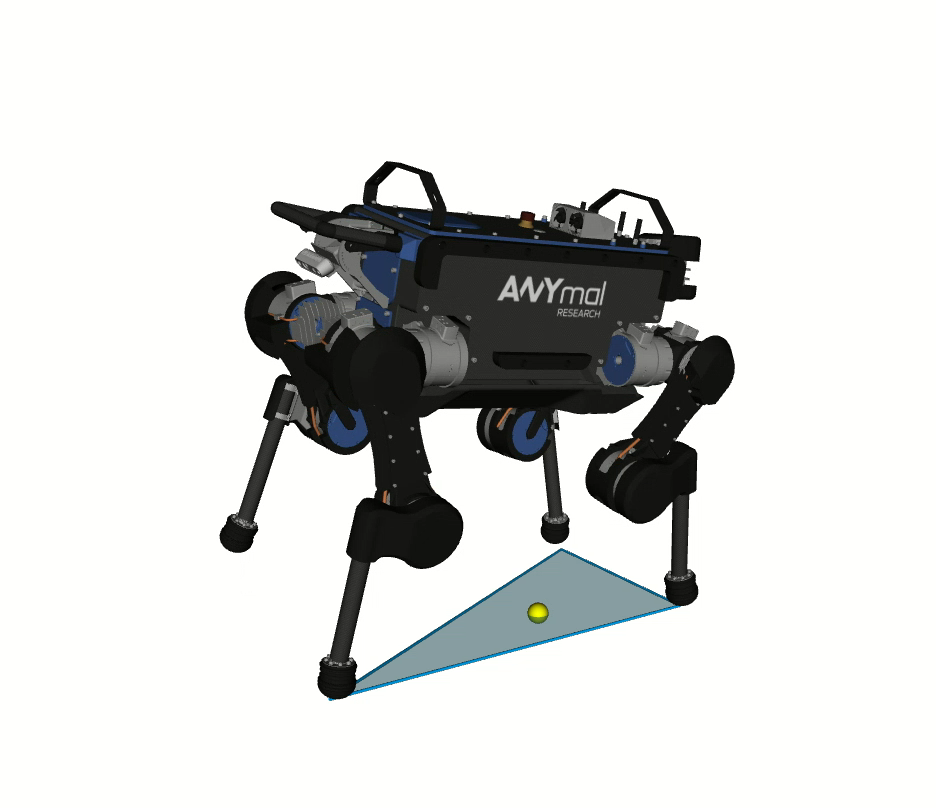}&
\subFig{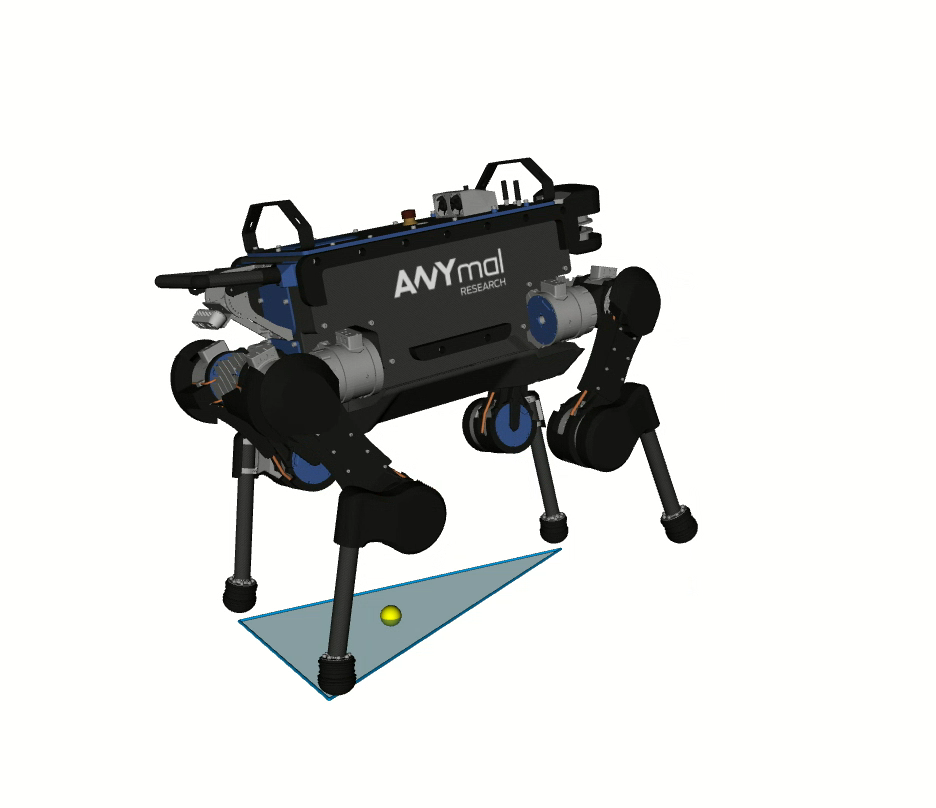}\\
\end{tabular*}
\caption{Screenshots of the motion resulting from our OCP regularization setup for a walking gait pattern. The blue area between represents the support polygon and the yellow circle is the Centre Of Pressure (COP) applied.}%
\label{fig::walking_pattern}
\end{figure}
\endgroup

\begingroup
\newcommand{\widthFig}{0.96\linewidth}
\setlength{\tabcolsep}{0.25pt} 
\renewcommand{\arraystretch}{1.5} 
\setlength\arrayrulewidth{1.pt}
\newcommand{\rowname}[1]{\rotatebox{0}{\makebox[\tempdima][c]{(#1)}}}

\newcommand{\subFig}[1]{\raisebox{-.5\height}{\includegraphics[clip,width=\widthFig]{{#1}}}}

\begin{figure}
\centering
\begin{tabular*}{\linewidth}{c@{\hskip 7pt} c}
\rowname{a}&
\subFig{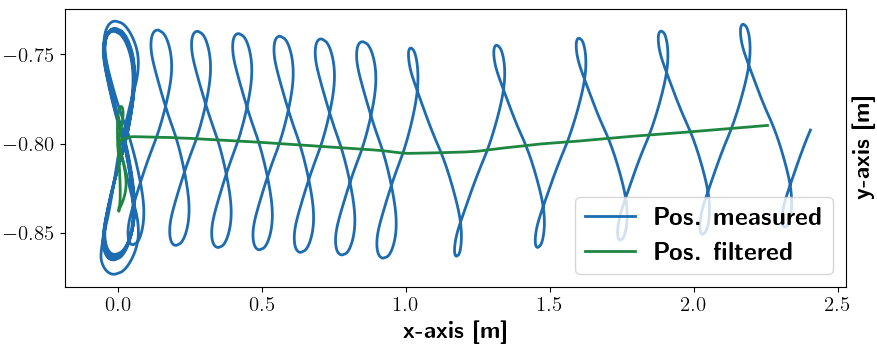}\\
\\[-3.ex]
\rowname{b}&
\subFig{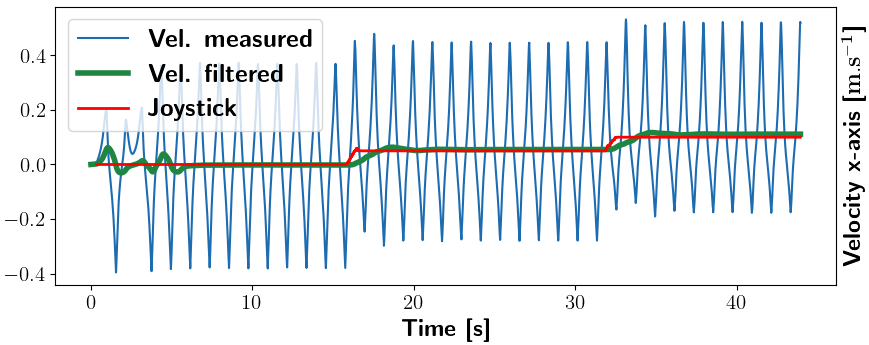}\\
\\[-3.ex]
\rowname{c}&
\subFig{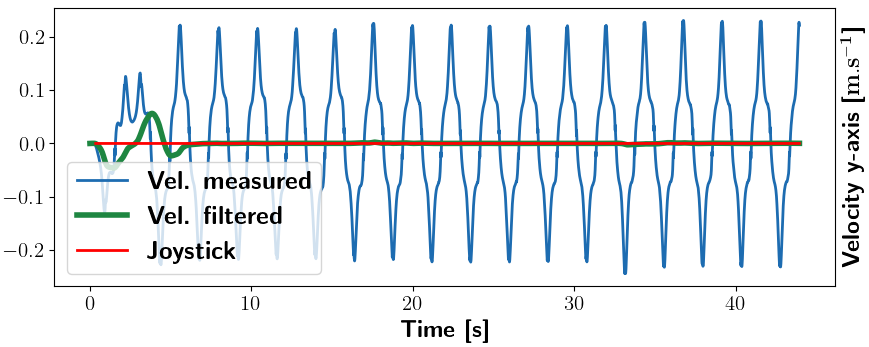}\\
\end{tabular*}
\caption{Base trajectory during a forward walking gait of period $2.4$s. The joystick command is in 2 separate steps of 0.05$m.s^{-1}$ and 0.1$m.s^{-1}$ along the x-axis. To reject disturbances, the filter is a moving average on the walk period. (a) base position on the ground floor plane. (b) and (c) base velocity along respectively the x and y axes. Small disturbances at first in the filtered quantities are due to the filter initialisation phase.}%
\label{fig::base_traj}
\end{figure}
\endgroup

Not constraining the robot's pose and velocity to follow a reference trajectory has a significant impact on the COM trajectory, especially in a walking gait pattern when only one foot is lifted at a time (Fig.~\ref{fig::walking_pattern}). Indeed, Fig.~\ref{fig::base_traj}-(a) shows that the base trajectory during a forward walk oscillates in a range of \SI{10}{\centi\meter} around the middle of the feet positions. However, constraining the base position and velocity would result in a fixed base orientation. This can be explained by the motion stability found in the optimization process. The centre of pressure is well located in the middle of the support polygon. The COM position is dragged inside this region resulting in this wave motion. Since it is a periodic motion, the state position has been filtered with a moving average for the walk period (Fig.~\ref{fig::base_traj}), rejecting all frequencies in synchronisation with the gait. This behaviour does not appear with a trotting gait since two opposite legs are left at the same time and the resulting oscillation is much smaller. 


\subsection{Robustness of the pipeline}
\label{sec::results-robustness}
A crucial point is to understand the repeatability of our experiments and how the locomotion controller adapts to unforeseen events.
We observed a significant difference in the reliability whether on-board perception was used or perfect knowledge of the environment was given.
The climbing stairs scenario was reliably repeated ten times with active perception, with a 80\% success rate, with errors mainly due to state estimation errors and drift. For climbing down the stairs the experiment was successfully repeated twice, but perception issues made the experiment difficult, as the camera position did not allow to clearly perceive all the stairs on the way down, resulting in unfeasible contact planning problems in some instances.
{In the absence of the perceptive part (experiments \linktoExpe{up-missing}, \linktoExpe{down-missing} and \linktoExpe{step40cm}), when the environment is perfectly known, the locomotion pipeline is stable and the experiments were successfully carried out on the first attempt and repeated twice.}

An interesting point attributed to the whole-body MPC is the ability to recover to unplanned situations, as in experiment \linktoExpe{down-missing} when ANYmal misses a step during the descent. Similarly, at the end of experiment \linktoExpe{multi}, a metal plate slips when walking on it. In both cases, the robot recovered properly (Fig.~\ref{fig::results:robustness}).

Finally, we refer the reader to our conference paper for an extensive ablation study on effect of the different elements of the pipeline on the success rate of the pipeline for the Solo robot~\cite{risbourg:hal-03594629}.

\begingroup
\newcommand{\widthFig}{0.48\linewidth}
\setlength{\tabcolsep}{0.25pt} 
\renewcommand{\arraystretch}{1.5} 
\setlength\arrayrulewidth{1.pt}
\newcommand{\subFigA}[1]{\raisebox{-.5\height}{\includegraphics[clip,trim={266px 66px 30px 66px},width=\widthFig]{{#1}}}}
\newcommand{\subFigB}[1]{\raisebox{-.5\height}{\includegraphics[clip,trim={166px 50px 133px 100px},width=\widthFig]{{#1}}}}

\def\boxit#1#2{%
  \smash{\llap{\rlap{\hspace{4pt}\href{#1}{\strut\raisebox{-.5\height}{\phantom{\rule{0.98\linewidth}{#2}}} } }~}}\ignorespaces}
\hypersetup{pdfborder=0 0 0}

\begin{figure}[hbt!]
\centering
\begin{tabular*}{\linewidth}{c c}
    \subFigA{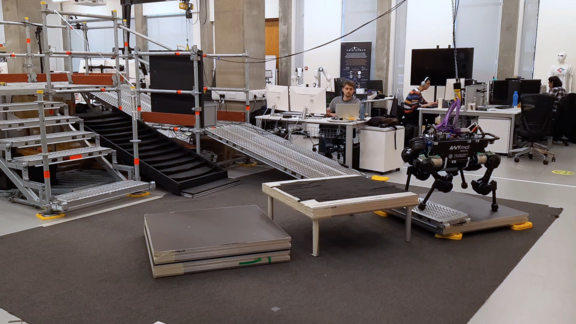}&
    \subFigA{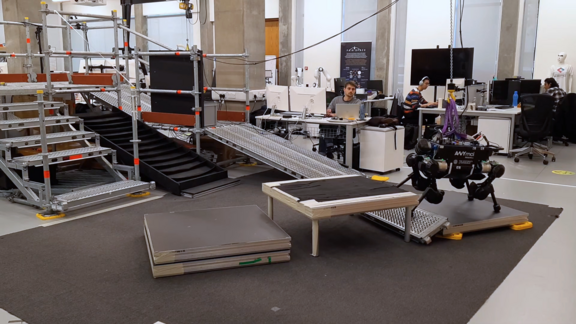}\\
    \\[-3.6ex]
    \subFigB{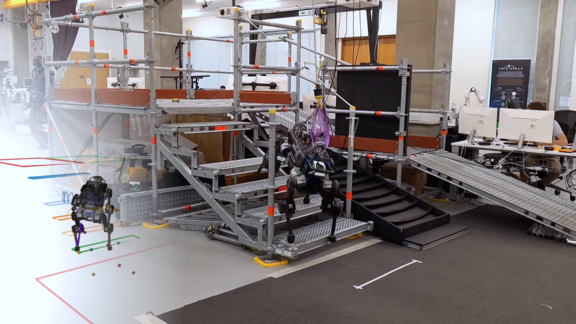}&
    \subFigB{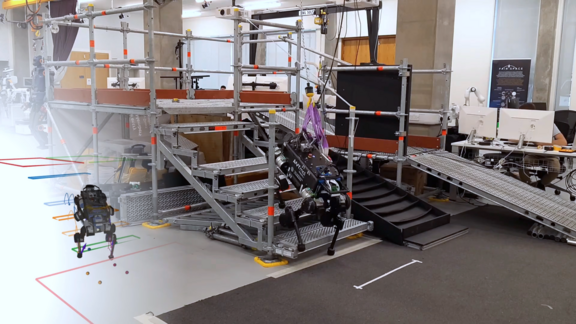}\\
\end{tabular*}
\caption{Screenshots of the robot recovering from an unplanned event. Top: the metal plate slips while the robot walks on it. Bottom: the two front feet slip and miss the intended step.}%
\label{fig::results:robustness}
\end{figure}
\endgroup

\subsection{Comparison with the state of the art}
\label{sec::results-sota}
The end-to-end implementation of frameworks similar to ours~\cite{Grandia-perceptive-through-NMPC, Jenelten_2022_TAMOLS} are not publicly available at the time of submission, neither the associated benchmarks and the quantity of work and resources to re-implement them is not reasonable, which prevents a quantitative comparison of the approaches.

\section{Discussion}
\label{sec::discussion}

In this paper, we propose a complete pipeline for perceptive quadrupedal locomotion; from onboard perception to locomotion generation and control. Our experimental results highlighted capabilities such as crossing challenging scenarios like climbing and descending industrial stairs or obstacle parkour with moving obstacles. Therefore, we have experimentally validated our architecture. This is based on the strong design assumption that a high-level planner, such as mixed-integer optimization, only selects the stepping contact surfaces. The utility of mixed-integer optimisation for design integration was partially validated in our previous work~\cite{risbourg:hal-03594629}. We used a different robot and integrated perception into our approach along with more complex experiments which reinforced interest in our decomposition.

\subsection{Perception pipeline} We have demonstrated the efficacy of our perceptive pipeline based on the probabilistic terrain mapping method \cite{Fankhauser2018ProbabilisticTerrainMapping} and the Plane-Seg algorithm \cite{Fallon2019PlaneSeg} which allow us to obtain satisfactory behaviour, in particular with our filtering work. Additionally, we demonstrated that our architecture can successfully navigate complex parkour terrains with moving objects or industrial staircases. Still, while our architecture is capable of navigating challenging scenarios when perception is removed, improving perception capabilities could enhance the overall robustness and reliability of the system. This would be particularly beneficial in scenarios such as the staircase where vision detection is restricted to only a few steps ahead due to our single depth camera with a 20\% downward angle. Additionally, the algorithm used to extract convex planes is not incremental and does not take into account previously computed planes, making it vulnerable to faulty scans in some cases. The implementation of a short term memory for the plane decomposition system would contribute to the robustness of the approach and will be considered in future work. 

\subsection{Collision avoidance of the body} The trade-off proposed in our approach is to strongly regularise the OCP around a reference end-effector trajectory while avoiding obstacles, which proved to work in a wide range of environments. This occurs even if a collision with the environment is not specifically considered in our approach. This could be considered within the MPC but would require a dedicated study since it is a challenging nonlinear optimisation problem. The safety margin around obstacles was sufficient in almost all scenarios to avoid collisions, except for the \SI{40}{\centi\meter} step in \linktoExpe{step40cm}, where we had to increase these margins to \SI{10}{\centi\meter}. As mentioned before, our approach ensured that there was no collision in the end-effector trajectory. It could have been done inside the MPC with additional terms in the objective. However, this would increase complexity and possibly affect convergence rates. Knee collision is implicitly considered by the margin around obstacles. However, this could be addressed explicitely by planning the foot trajectory including the whole leg, as is the case in \cite{Zucker-Optimisation-LittleDog}, although re-planning at \SI{50}{\hertz} might become unfeasible.

\subsection{Comparison with the state of the art} Two recent works~\cite{Grandia-perceptive-through-NMPC, Jenelten_2022_TAMOLS} present interesting similarities to our architecture. We can first observe the decomposed approach used in both cases with the foot location optimised outside the MPC. We have underlined the technical difficulties that would allow a objective comparison of the frameworks, although we would like to integrate the publicly available perception module proposed in \cite{Miki-ANYmal-wild-perception}. However, this approach does not return a selection of potential candidates surfaces and only one candidate, such that further integration will be required to make the approach compatible with our framework.

\subsection{Improvement points} As often occurs in model-based approaches to quadruped robots, the main limitation is the gait fixed beforehand. It would be interesting to replace the high-level planner with an acyclic planner to optimise the timing of contacts and the type of gait. This could lead to computing-time issues especially due to the increase in problem complexity. To take this further, it would be interesting to integrate this into a global planner to achieve autonomous behaviour.
Although we consider a stable walking gait necessary for the most challenging scenarios, we also acknowledge the potential benefits of implementing a trotting gait on the hardware. However, transitioning to a trotting gait, which works well in simulation, proved more challenging to implement on the robot. We believe that this limitation could be overcome with a faster controller and more integrative efforts.
Finally, our foot placement is based on Raibert's heuristic, which is optimal on flat ground when considering an inverse linear pendulum model for the robot. We extend it in 3D which produces satisfying foot placement as we demonstrated in our experiment. Nevertheless, this remains a heuristic and does not ensure the position of feet is feasible in terms of torque power limit. 

\section{Conclusion}

In this paper, we present  a complete methodology for crossing challenging terrains; from terrain perception to locomotion generation and control. We have demonstrated our pipeline on various terrains, such as an industrial staircase or a parkour-type environment with a moving object in the scene. These experiments have allowed us to further validate our approach based on a sub-division of the global problem. First, a high-level planner formulated as a mixed-integer optimisation selects only the next surfaces of contact with a horizon of a few contacts (6 to 8 in our experiments). To achieve this, one must adapt the perception to extract relevant potential surfaces as convex planes. For motion generation, we rely on an efficient whole-body MPC and a linear state feedback controller. Collision-free trajectory and footstep adaptation on the high-level planner are optimised separately. The OCP problem is then strongly regularized around this reference end-effector trajectory while robot's posture is adapted by the MPC. This represents a wise choice of design to leverage whole-body optimisation capabilities. To move further, we would like to use a more complex high-level planner to optimise contact timing to cross even more challenging terrains, for example, dynamic motions that include jumping over gaps or obstacles. 

\begin{acks}
This research was supported by (1) the European Commission under the Horizon 2020 project Memory of Motion (MEMMO, project ID: 780684) and (2) ROBOTEX 2.0 (Grants ROBOTEX ANR-10-EQPX-44-01 and TIRREX-ANR-21-ESRE-0015).
\end{acks}

\bibliographystyle{SageH}
\bibliography{biblio}

\end{document}